\documentclass[letterpaper]{article} % DO NOT CHANGE THIS
\PassOptionsToPackage{table}{xcolor}
\usepackage[preprint]{aaai2027}  % arxiv preprint: shows real authors + drops the AAAI copyright bar
\usepackage[hyphens]{url}  % DO NOT CHANGE THIS
\usepackage{graphicx} % DO NOT CHANGE THIS
\usepackage{natbib}  % DO NOT CHANGE THIS AND DO NOT ADD ANY OPTIONS TO IT
\usepackage{caption} % DO NOT CHANGE THIS AND DO NOT ADD ANY OPTIONS TO IT
\usepackage{algorithm}
\usepackage{algorithmic}

\usepackage{newfloat}
\usepackage{listings}
\DeclareCaptionStyle{ruled}{labelfont=normalfont,labelsep=colon,strut=off} % DO NOT CHANGE THIS
\floatstyle{ruled}
\newfloat{listing}{tb}{lst}{}
\floatname{listing}{Listing}

\usepackage{booktabs}

\usepackage{amsmath}
\usepackage{amssymb}
\usepackage{float}  % arxiv combined build: appendix uses [H] placement (AAAI's no-float rule does not apply to arxiv)
\usepackage{dblfloatfix}   % fixes double-column (figure*/table*) floats deferring to the document end

\usepackage[table]{xcolor}
\usepackage{array}
\usepackage{booktabs}

\definecolor{DWBlue}{HTML}{EAF3FA}
\definecolor{RefGray}{HTML}{F1F3F4}
\definecolor{PanelBlue}{HTML}{D9E8F3}
\definecolor{RowGray}{HTML}{F8F9FA}
\definecolor{RuleBlue}{HTML}{537895}

\usepackage{enumitem}

\usepackage[most]{tcolorbox}
\usepackage{enumitem}
\usepackage{xcolor}

\definecolor{AlgBackground}{HTML}{F5F6F7}
\definecolor{AlgPhaseBackground}{HTML}{E1E5E8}
\definecolor{AlgTitleBackground}{HTML}{D6DCE0}
\definecolor{AlgBorder}{HTML}{AAB4BB}
\definecolor{AlgAccent}{HTML}{2F4554}

\newcounter{factoralgorithm}[section]
\renewcommand{\thefactoralgorithm}{
    \thesection.\arabic{factoralgorithm}
}

\newlist{algsteps}{enumerate}{1}
\setlist[algsteps]{
    label=\textcolor{AlgAccent}{\bfseries\arabic*.},
    leftmargin=2.2em,
    labelsep=0.55em,
    itemsep=3pt,
    topsep=2pt,
    parsep=0pt,
    partopsep=0pt
}

\newcommand{\AlgPhase}[1]{%
    \item[]
    \vspace{3pt}
    \noindent
    \colorbox{AlgPhaseBackground}{%
        \parbox{\dimexpr\linewidth-2\fboxsep\relax}{%
            \vspace{2pt}
            \textbf{\textcolor{AlgAccent}{#1}}
            \vspace{2pt}
        }%
    }
    \vspace{1pt}
}

\definecolor{FTHeader}{HTML}{2F4554}
\definecolor{FTSubheader}{HTML}{607D8B}
\definecolor{FTBorder}{HTML}{87969F}
\definecolor{FTContract}{HTML}{E8ECEF}

\definecolor{LayoutCell}{HTML}{CFE3D4}
\definecolor{LayoutRow}{HTML}{F1F7F2}

\definecolor{AgentCell}{HTML}{CFE1F2}
\definecolor{AgentRow}{HTML}{F1F6FA}

\definecolor{VisibilityCell}{HTML}{F4DFC0}
\definecolor{VisibilityRow}{HTML}{FCF7EF}

\definecolor{InteractionCell}{HTML}{EBCFD3}
\definecolor{InteractionRow}{HTML}{FAF2F3}

\usepackage[table]{xcolor}
\usepackage{booktabs}
\usepackage{tabularx}
\usepackage{array}

\definecolor{CfgNavy}{HTML}{18364B}
\definecolor{CfgBlue}{HTML}{E4EEF7}
\definecolor{CfgCyan}{HTML}{E6F4F2}
\definecolor{CfgGold}{HTML}{FFF1CC}
\definecolor{CfgRose}{HTML}{F9E4E7}
\definecolor{CfgGreen}{HTML}{E1F1E6}
\definecolor{CfgGray}{HTML}{F2F4F5}
\definecolor{CfgInk}{HTML}{263A47}
\definecolor{CfgPending}{HTML}{9B3540}

\newcolumntype{Y}{>{\raggedright\arraybackslash}X}

\usepackage[table]{xcolor}
\usepackage{longtable}
\usepackage{booktabs}
\usepackage{array}
\usepackage{ragged2e}

\usepackage{booktabs,tabularx,makecell,array,xcolor,graphicx}
\definecolor{DecoderBlue}{HTML}{EAF3F8}
\definecolor{DecoderTeal}{HTML}{E6F4F1}
\definecolor{DecoderGold}{HTML}{FFF3D6}
\definecolor{DecoderGray}{HTML}{F2F3F4}

\usepackage{graphicx}

\newcommand{\scenepanel}[2]{%
    \begin{minipage}[t][2.95cm][t]{0.19\textwidth}
        \centering
        \begin{minipage}[c][2.35cm][c]{\linewidth}
            \centering
            \includegraphics[
                width=\linewidth,
                height=2.35cm,
                keepaspectratio
            ]{#1}
        \end{minipage}

        \vspace{0.4mm}
        {\small\strut #2}
    \end{minipage}%
}

\definecolor{TableBlue}{HTML}{315B7D}
\definecolor{LightBlue}{HTML}{EEF4F8}
\definecolor{TickGreen}{HTML}{16805C}
\definecolor{TickBG}{HTML}{E6F5EE}
\definecolor{CrossRed}{HTML}{B23A48}
\definecolor{CrossBG}{HTML}{FBEAEC}

\title{FactorJEPA: Factorizing Monolithic Futures into Layout–Agent–Interaction Channels for Crowded and Chaotic Global South Urban Worlds}

\author{
    Kapil Wanaskar\textsuperscript{\rm 1}\thanks{All authors conducted this work independently, outside their roles and employment at their respective companies.},
    Gaytri Jena\textsuperscript{\rm 2},
    Aman Chadha\textsuperscript{\rm 3},
    Vini{}ja Jain\textsuperscript{\rm 4},  % {} breaks the T1 "ij" ligature so it renders "Vinija", not "Vinĳa"
    Vasu Sharma\textsuperscript{\rm 5},
    Amitava Das\textsuperscript{\rm 6}
}
\affiliations{
    \textsuperscript{\rm 1}San Jose State University, USA\quad
    \textsuperscript{\rm 2}UC Berkeley, USA\quad
    \textsuperscript{\rm 3}Apple, USA\quad
    \textsuperscript{\rm 4}Meta, USA\quad
    \textsuperscript{\rm 5}PocketFM, USA\quad
    \textsuperscript{\rm 6}Pragya Lab, BITS Pilani Goa, India
}

\usepackage[colorlinks=true,urlcolor=blue,linkcolor=black,citecolor=black]{hyperref}
\makeatletter
\long\def\PackageError#1#2#3{}% swallow aaai2027's \AtBeginDocument hyperref guard
\makeatother

\begin{document}

\maketitle

\begin{abstract}

%\textbf{World models} have recently attracted significant attention as a framework for learning \textbf{latent representations} that preserve the \textbf{structure and dynamics of the physical world}. Among recent approaches, \textbf{Joint Embedding Predictive Architectures (JEPA)} are especially compelling because they predict \textbf{masked or future content directly in embedding space}, avoiding costly pixel-level generation while promoting semantic abstraction.

\textbf{World models} have attracted significant attention for their ability to capture and predict the \textbf{structure and dynamics of the physical world}. In this emerging landscape, \textbf{Joint Embedding Predictive Architectures (JEPA)} offer a particularly compelling direction. 

In this paper, we study a largely unexplored regime: \textbf{populous, crowded, and chaotic Global South urban environments}, which we call \textbf{DENSEWORLD}. Unlike the lower-density, lane-structured settings that dominate existing evaluations, these scenes exhibit \textbf{soft spatial boundaries}, \textbf{extreme agent heterogeneity}, \textbf{persistent occlusion}, and rapid \textbf{social negotiation} under mixed traffic. We introduce the first large-scale dataset for this regime, comprising \textbf{1,000 hours of drive-through, walk-through, and aerial video collected across 22 cities}. The dataset covers a wide variety of \textbf{scene types, times of day, weather conditions, crowd and traffic densities, traffic mixes, pedestrian--vehicle separation patterns, road layouts and surfaces, infrastructure quality, encroachment levels, locally distinctive objects, vegetation, lighting conditions, and video quality}. Our evaluation reveals that existing JEPA formulations struggle to preserve dense interaction dynamics under heterogeneity and partial observability.

\textbf{\emph{We introduce FactorJEPA}}, which makes \textbf{\emph{world structure a first-class predictive primitive}}. Rather than encoding the future in a monolithic latent, it composes \textbf{\emph{layout}}, \textbf{\emph{entities}}, and \textbf{\emph{interactions}}, using a visibility gate and separated subspaces to preserve partially observed agents and discourage cross-factor shortcuts. FactorJEPA improves \textbf{\emph{(i) future-latent accuracy}} (Future-frame L1), \textbf{\emph{(ii) intervention-sensitive prediction}} (Causal L1), and \textbf{\emph{(iii) robustness to reduced visual evidence}} (Mask-ratio slope), while exposing \textbf{\emph{(iv) a reproducible motion-information trade-off}} (Motion cosine). Method rankings replicate across 2B and 1B V-JEPA~2.1 backbones, with $\rho=0.895$--$0.978$ across these four diagnostics.

\noindent\textbf{Data and models.} We publicly release the \href{https://huggingface.co/datasets/anonymousML123/denseworld-115k}{DENSEWORLD-115k dataset} and the surgery-trained \href{https://huggingface.co/datasets/anonymousML123/factorjepa-outputs/tree/main/outputs/full/vjepa_2_1_vitg_1B/train/m09c_surgery_3stage_DI_diheavy_encoder}{FactorJEPA checkpoints}.

%As \textbf{world models} emerge as foundations for agents, robotics, and physical AI, JEPA-based models remain largely \textbf{\emph{silent simulators}}: they forecast futures without revealing what those futures look like. We introduce a \textbf{Cosmos-initialized latent-to-RGB decoder} that renders FactorJEPA's predicted future as a visually inspectable RGB frame.  To foster future research, we make the code and evaluation artifacts \href{https://anonymous.4open.science/r/NEPHOS-72D4/README.md}{\textbf{publicly available}}.

\end{abstract}

% Page-1 teaser (demo): declared AFTER the abstract so the abstract leads; single-column [t!] lands at the top of the right column on page 1.
\begin{figure}[t!]
    \centering
    \includegraphics[width=\columnwidth]{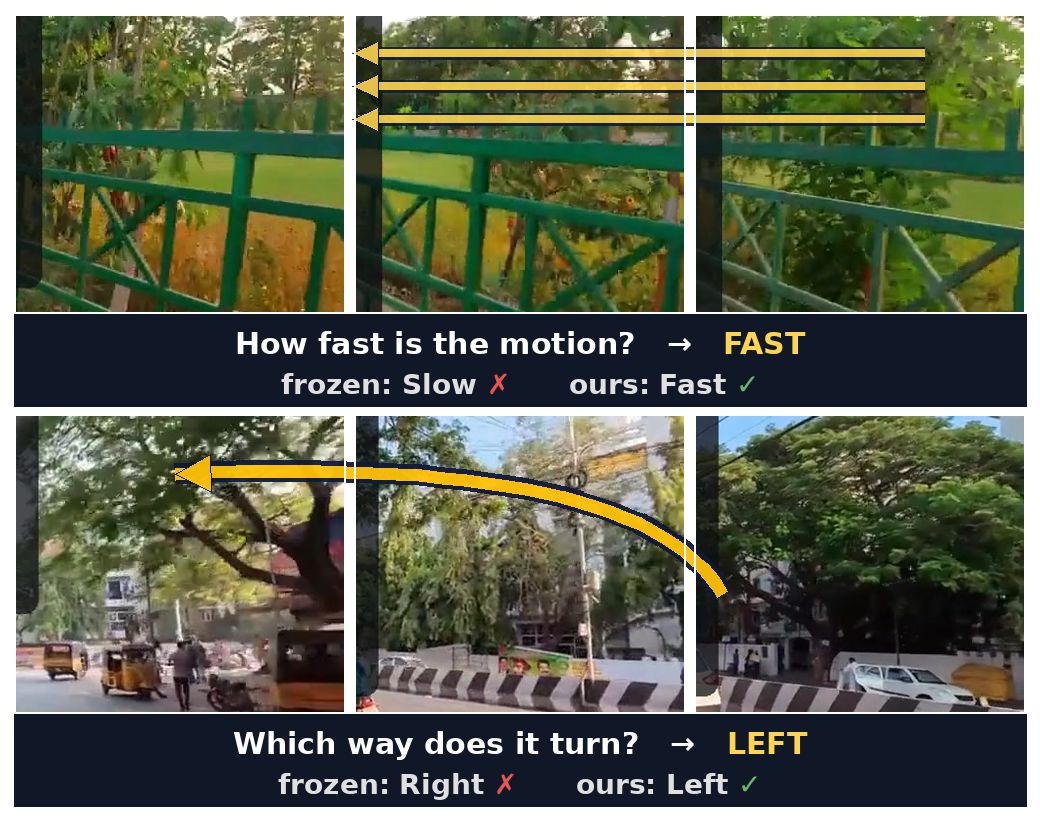}
    \caption{\textbf{Same question, two encoders.} Each row is a three-frame filmstrip of one held-out clip posed as a multiple-choice question, answered from an identical probe head over the frozen V-JEPA~2.1 encoder vs.\ ours (factor-view predictor surgery); only the backbone differs. \emph{Top:} motion speed (ours 69.8\% vs frozen 60.9\%). \emph{Bottom:} turn direction.}
    \label{fig:demo_cards_full}
\end{figure}

\begin{comment}
    
\begin{figure*}[t]
    \centering
    \includegraphics[width=\textwidth]{figures/denseworld_scene_types_overview.png}
    \caption{\textbf{DENSEWORLD 1.0 scene-type coverage.}
    Representative examples from the dataset illustrate the breadth of urban environments covered by \textbf{DENSEWORLD 1.0}, including \textbf{market}, \textbf{residential}, \textbf{commercial}, \textbf{promenade}, \textbf{transit}, \textbf{highway}, \textbf{heritage}, \textbf{junction}, \textbf{flyover}, \textbf{beach}, \textbf{ghat}, \textbf{bazaar}, and \textbf{skyline} scenes. This diversity reflects the spatial, social, and infrastructural heterogeneity of populous, crowded, and chaotic Global South urban environments.}
    \label{fig:denseworld_scene_types}
    \vspace{-1.5em}
\end{figure*}
\end{comment}

\begin{figure*}[t]
    \centering

    %-------------------- Row 1 --------------------%
    \scenepanel{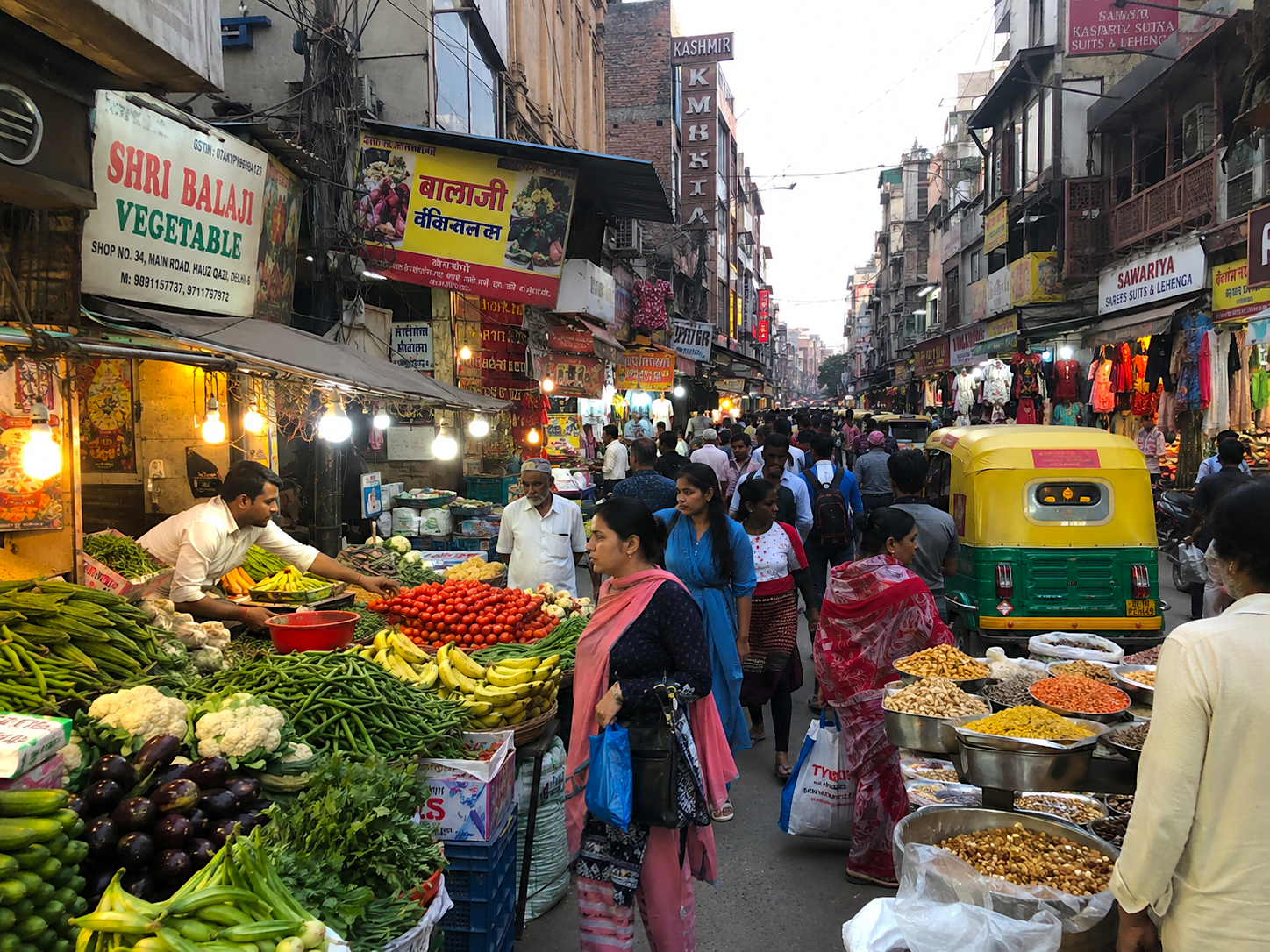}{market}
    \hfill
    \scenepanel{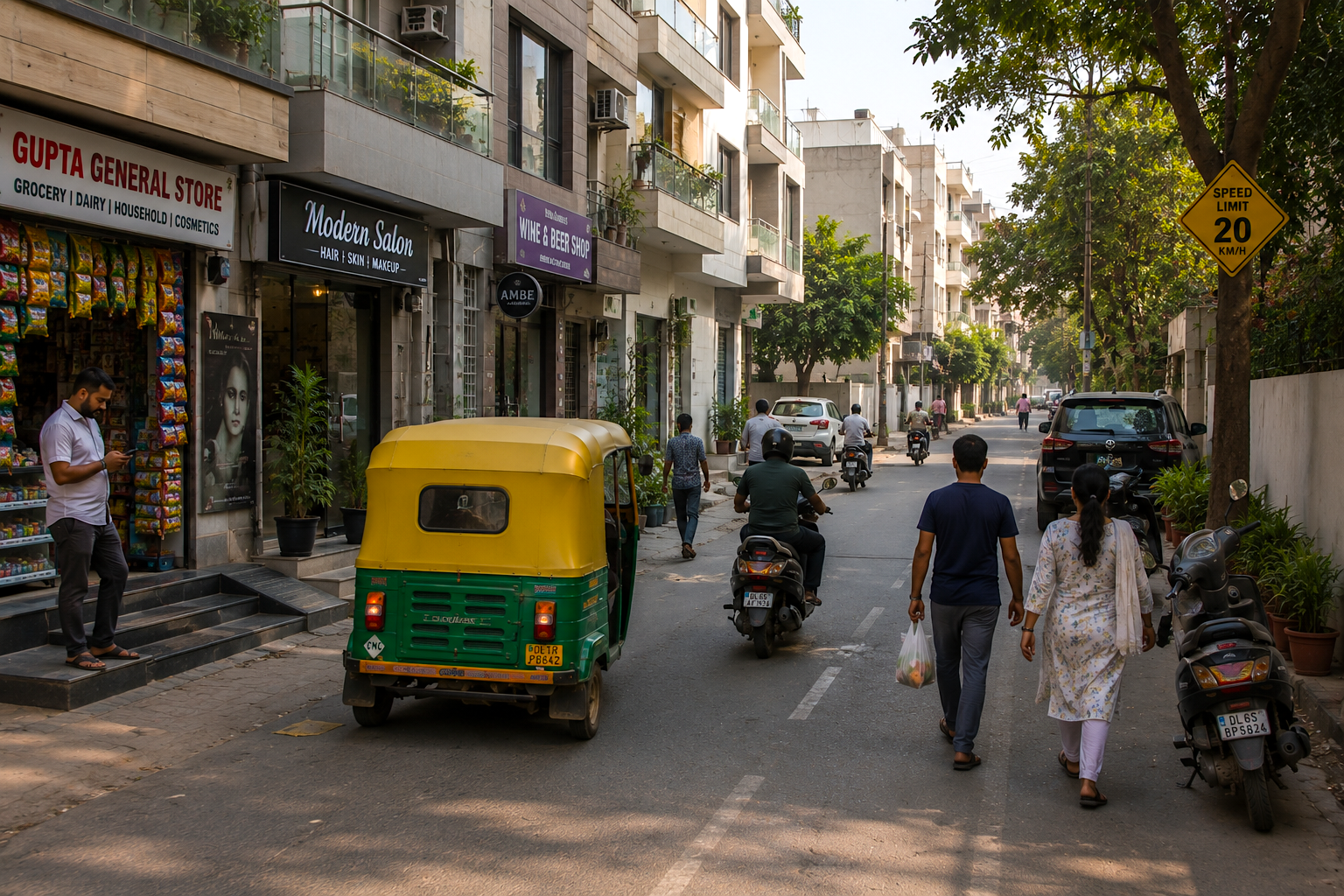}{residential}
    \hfill
    \scenepanel{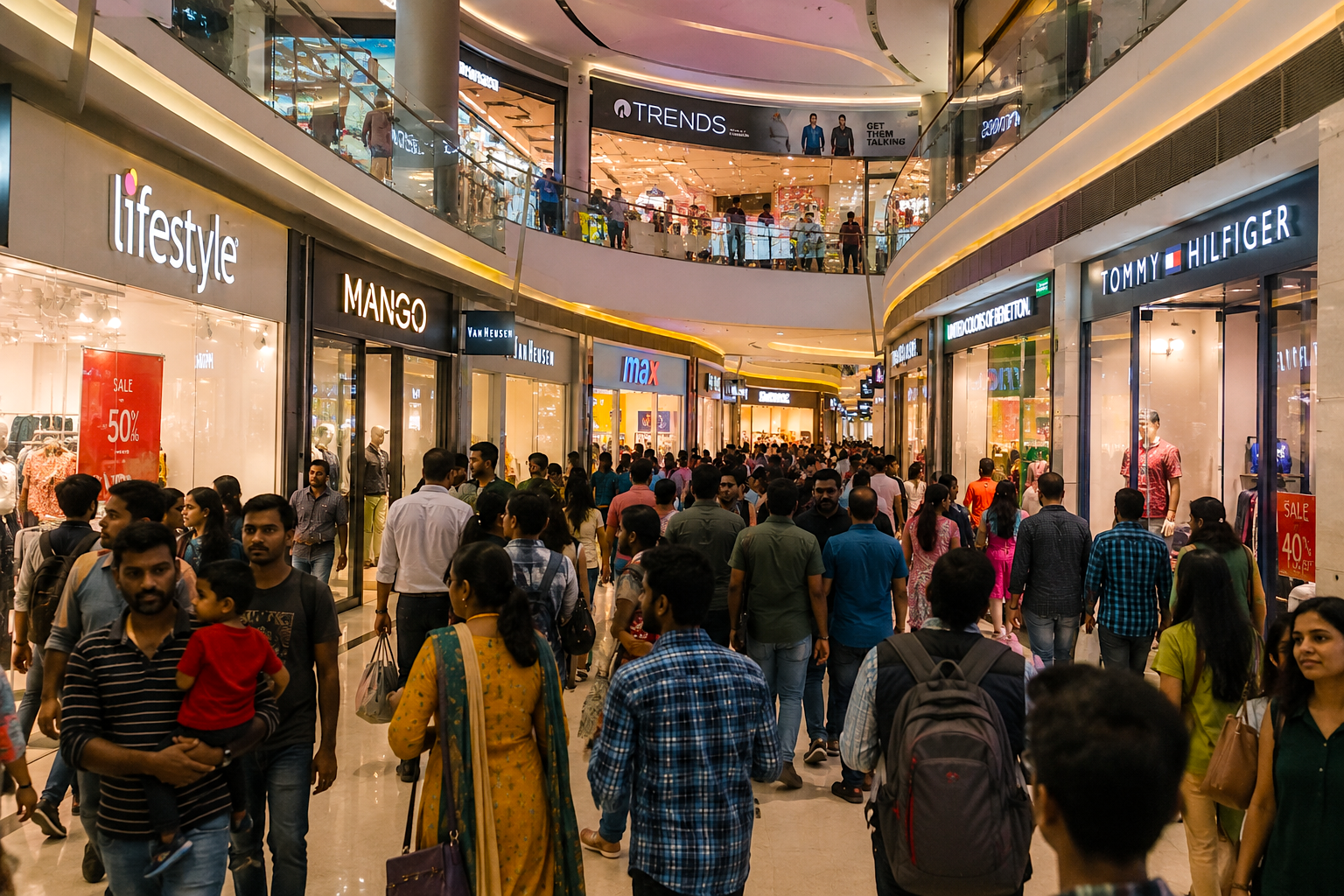}{commercial}
    \hfill
    \scenepanel{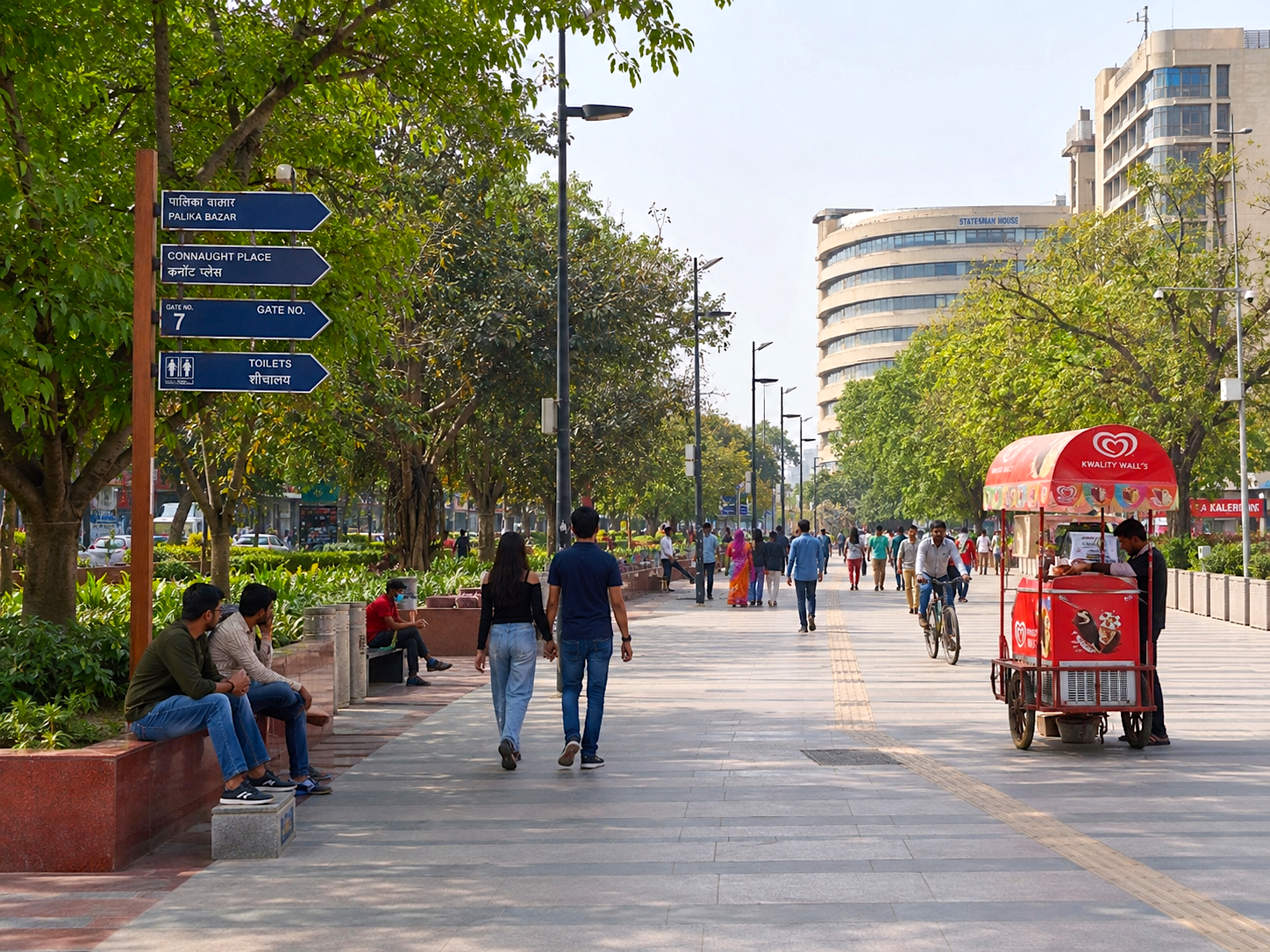}{promenade}
    \hfill
    \scenepanel{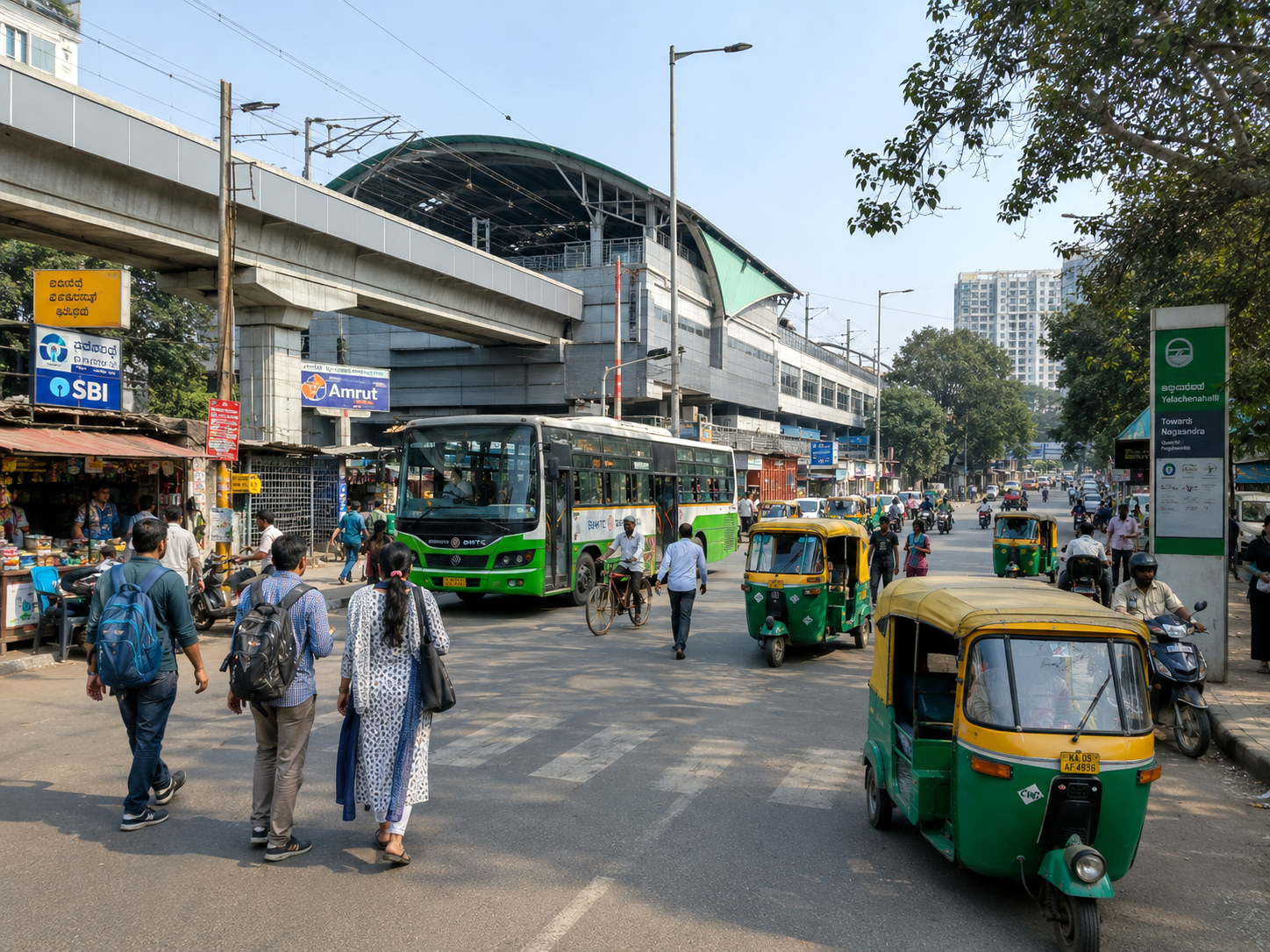}{transit}

    %\par\vspace{1.5mm}
    \vspace{-0.5em}

    %-------------------- Row 2 --------------------%
    \scenepanel{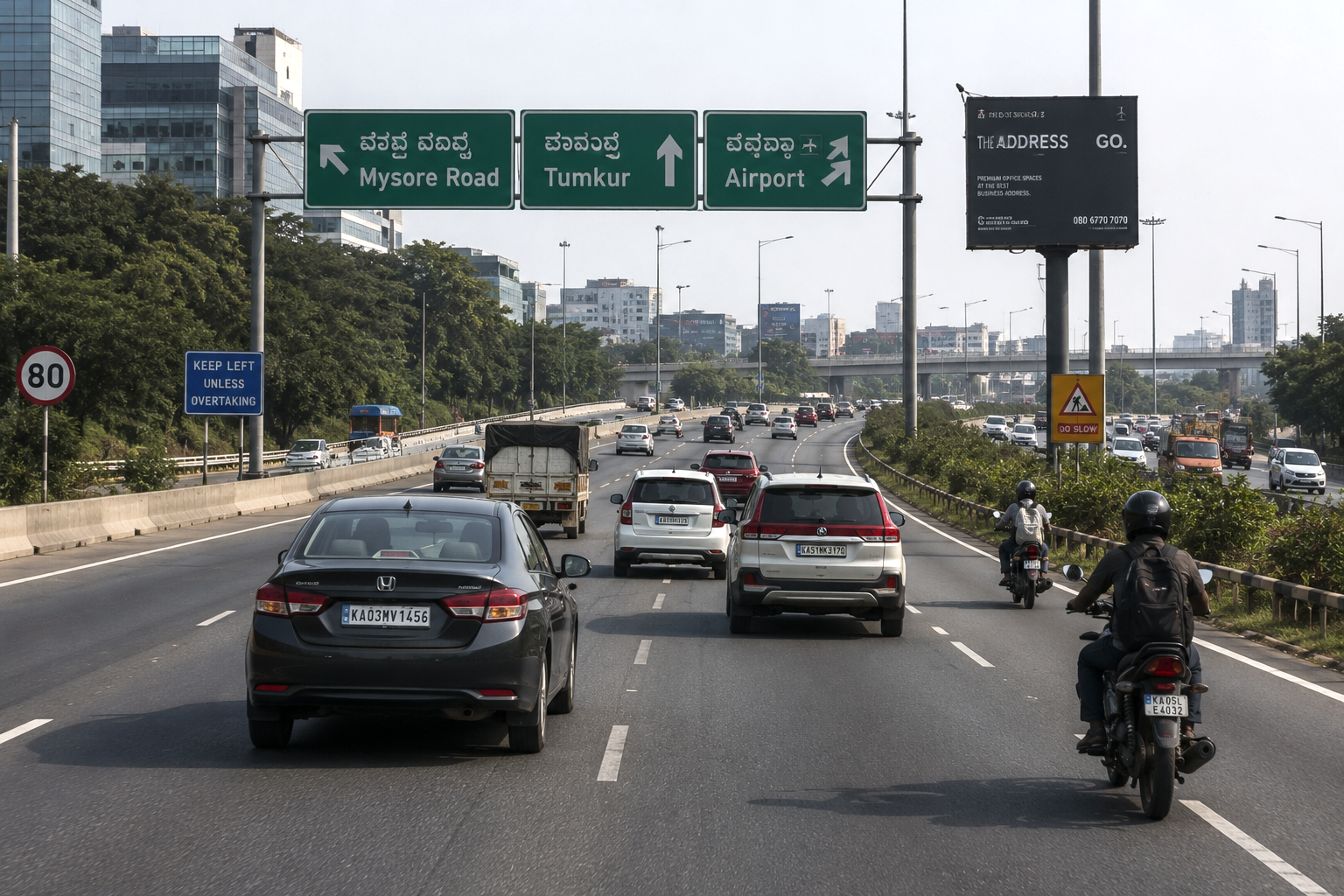}{highway}
    \hfill
    \scenepanel{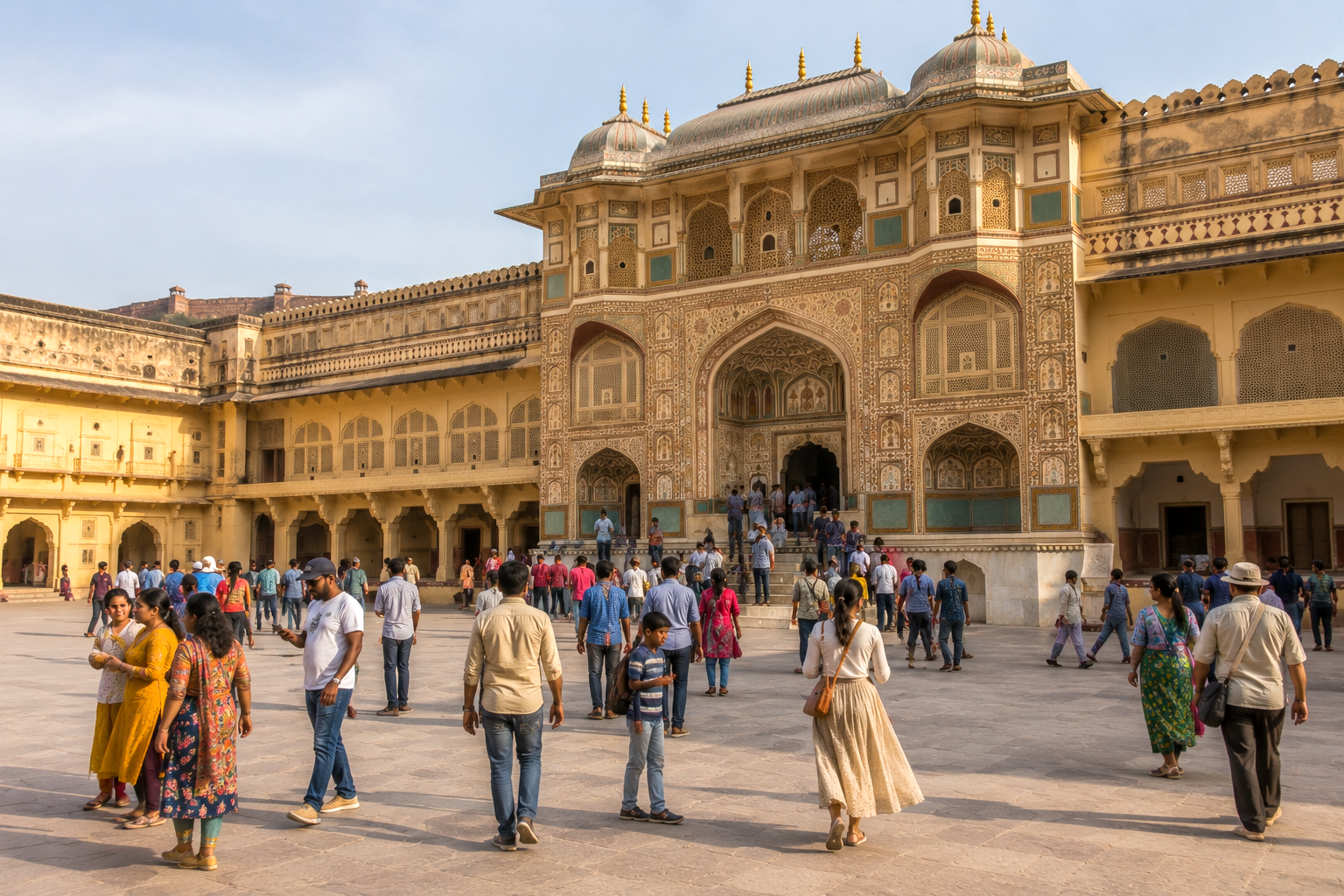}{heritage}
    \hfill
    \scenepanel{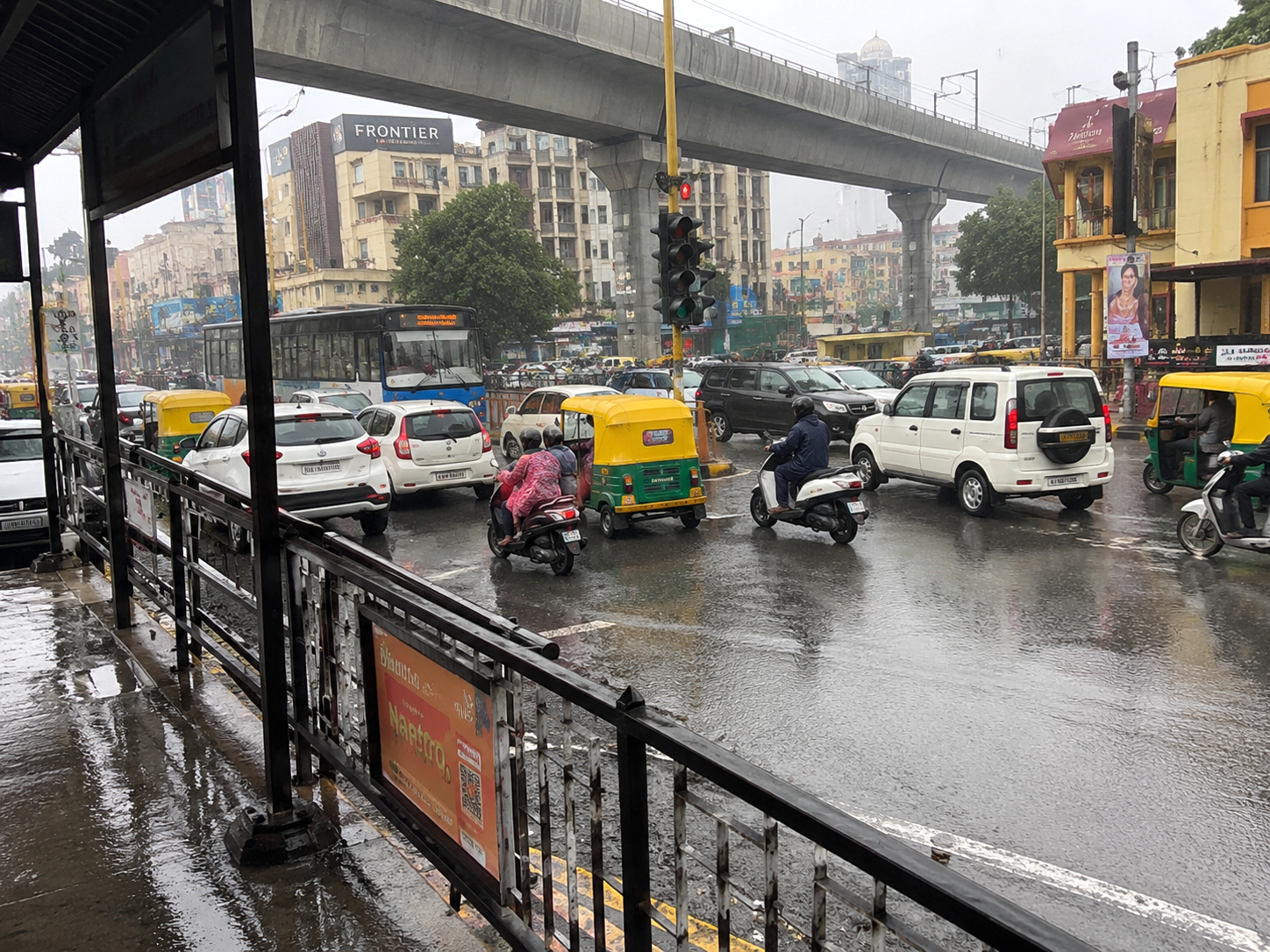}{junction}
    \hfill
    \scenepanel{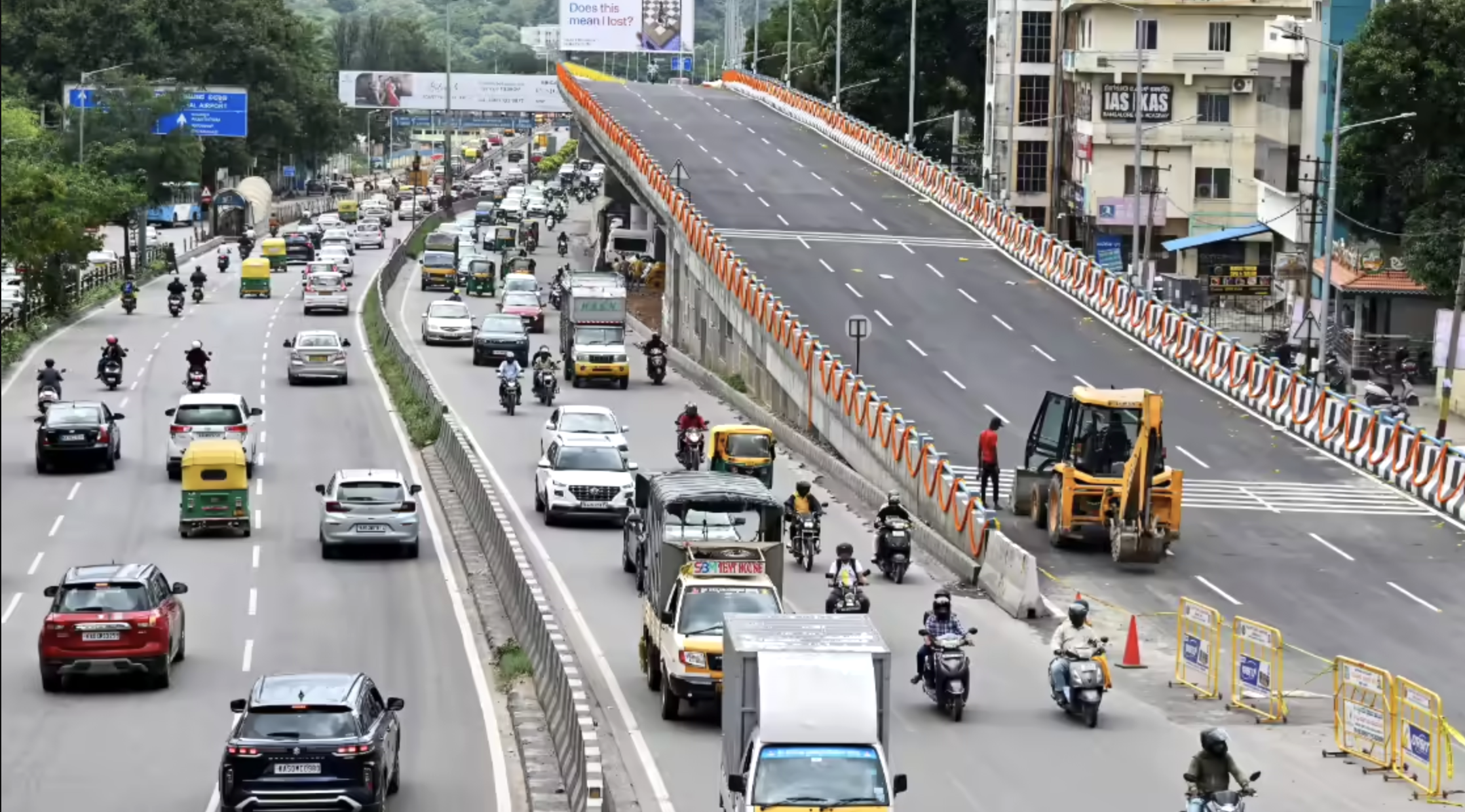}{flyover}
    \hfill
    \scenepanel{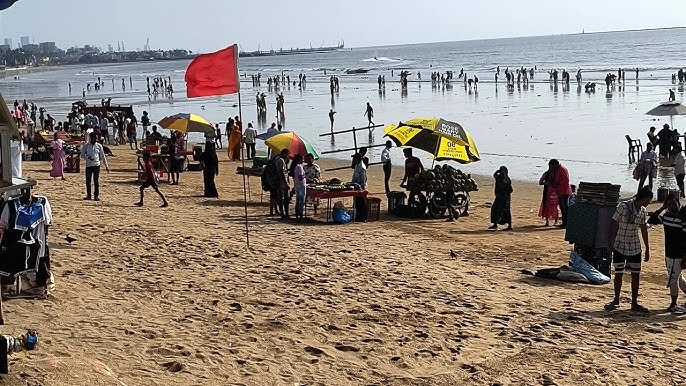}{beach}

    %\vspace{0.5mm}
    \vspace{-1em}

    \caption{
        \textbf{DENSEWORLD 1.0 scene-type coverage.}
        Representative examples from the dataset illustrate the breadth of
        urban environments covered by DENSEWORLD 1.0, including
        \textbf{market, residential, commercial, promenade, transit,
        highway, heritage, junction, flyover, and beach} scenes.
        This diversity reflects the spatial, social, and infrastructural
        heterogeneity of populous, crowded, and chaotic urban environments.
    }
    \label{fig:denseworld_scene_coverage}
    \vspace{-1em}
\end{figure*}

\section{\textbf{DENSEWORLD}: A Benchmark for \textbf{Populous, Crowded, and Chaotic} Global South}

We use \textbf{DENSEWORLD} to denote a \textbf{critical but underrepresented world-modeling regime}: \textbf{populous, crowded, and chaotic Global South urban scenes}. It is defined not only by geography, but by measurable properties: \textbf{high agent count density}, \textbf{high agent occupancy}, \textbf{persistent occlusion}, \textbf{agent heterogeneity}, and \textbf{interaction pressure}. This regime matters because many emerging urban environments combine \textbf{high population density}, \textbf{mixed mobility systems}, \textbf{diverse infrastructure patterns}, and \textbf{rapid spatial change}, making them scientifically important yet underrepresented in world-model benchmarks \cite{mahendra2021seven,ellis2016leveraging}. Thus, world models must handle uncertainty, density, and relational complexity that is less prominent in lower-density, lane-structured benchmarks. See Figure~\ref{fig:denseworld_scene_coverage} for representative DENSEWORLD scenes.

We characterize DENSEWORLD through four coupled properties. \textbf{First}, it exhibits \textbf{soft spatial boundaries}: road edges, drivable space, and pedestrian regions are often negotiated rather than crisply marked \cite{varma2019idd, dokania2023idd3d}. \textbf{Second}, it contains \textbf{extreme agent heterogeneity}, with pedestrians, cars, two-wheelers, carts, animals, delivery vehicles, and rickshaws coexist locally \cite{khan1999heterogeneous,asaithambi2012mixedtraffic}. \textbf{Third}, it has \textbf{persistent occlusion} from density, clutter, and near-field encounters, producing severe \textbf{partial observability} \cite{varma2019idd, dokania2023idd3d}; prediction must rely on \emph{memory}, \emph{object permanence}, and \emph{relational inference}, not only visible appearance. \textbf{Fourth}, motion is governed less by rigid rule compliance than by rapid \textbf{social negotiation} under mixed traffic and weak lane discipline \cite{papathanasopoulou2018weaklane,asaithambi2012mixedtraffic}.

Together, these properties make DENSEWORLD a \textbf{natural stress test} for world models: success requires representations that preserve \textbf{layout}, \textbf{agent state}, and \textbf{interaction dynamics} under uncertainty, rather than collapsing into one \textbf{entangled predictive latent}. As shown in Section~\ref{sec:denseworld_dataset}, we measure this regime through agent count density, agent occupancy, occlusion pressure, interaction pressure, and agent heterogeneity. This motivates a predictor that separates \textbf{layout}, \textbf{entities}, and \textbf{interactions}.

\subsection{\textbf{DENSEWORLD 1.0}: the Benchmark}
\label{sec:denseworld_dataset}

We partnered with two professional video-collection companies to record approximately
\textbf{1,000 hours} of urban footage across \textbf{22 Tier-1} and \textbf{Tier-2 cities in India}. The data spans \textbf{drive-through}, \textbf{walk-through}, and \textbf{aerial} viewpoints, and long-form recordings are segmented using scene-detection and shot-boundary methods into clips of \textbf{6--13 seconds}. DENSEWORLD 1.0 covers commercial, residential, transit, heritage, coastal, and high-density street settings, including \textbf{markets}, \textbf{commercial streets}, \textbf{transit corridors}, \textbf{junctions}, \textbf{flyovers}, \textbf{bazaars}, \textbf{ghats}, \textbf{beaches}, and \textbf{skylines}. It further spans variations in the \textbf{time of day}, \textbf{weather}, \textbf{crowd density}, \textbf{traffic mix}, \textbf{pedestrian--vehicle separation}, \textbf{road geometry}, \textbf{surface quality}, \textbf{encroachment}, and \textbf{lighting}. Beyond standard urban actors, the dataset includes locally frequent agents and objects such as \textbf{auto-rickshaws}, \textbf{cycle rickshaws}, \textbf{street vendors}, and \textbf{animals}. Before release, all footage is processed through privacy-preserving filters, including face and license-plate blurring and removal of sensitive segments.

A defining feature of DENSEWORLD is \textbf{heterogeneous mixed traffic}: diverse agents share the same right of way under \textbf{weak lane discipline}, creating strong lateral and longitudinal coupling \cite{khan1999heterogeneous,papathanasopoulou2018weaklane,asaithambi2012mixedtraffic}. Unlike lane-structured settings, these scenes exhibit \textbf{fluid spatial support}, unstable visibility, and dense \textbf{local negotiation} among cars, buses, trucks, auto-rickshaws, two-wheelers, bicycles, carts, and pedestrians.

\subsection{\textbf{How Dense is DENSEWORLD?}}

\begin{figure}[ht]
%\vspace{-1em}
    \centering
    \includegraphics[width=\columnwidth]{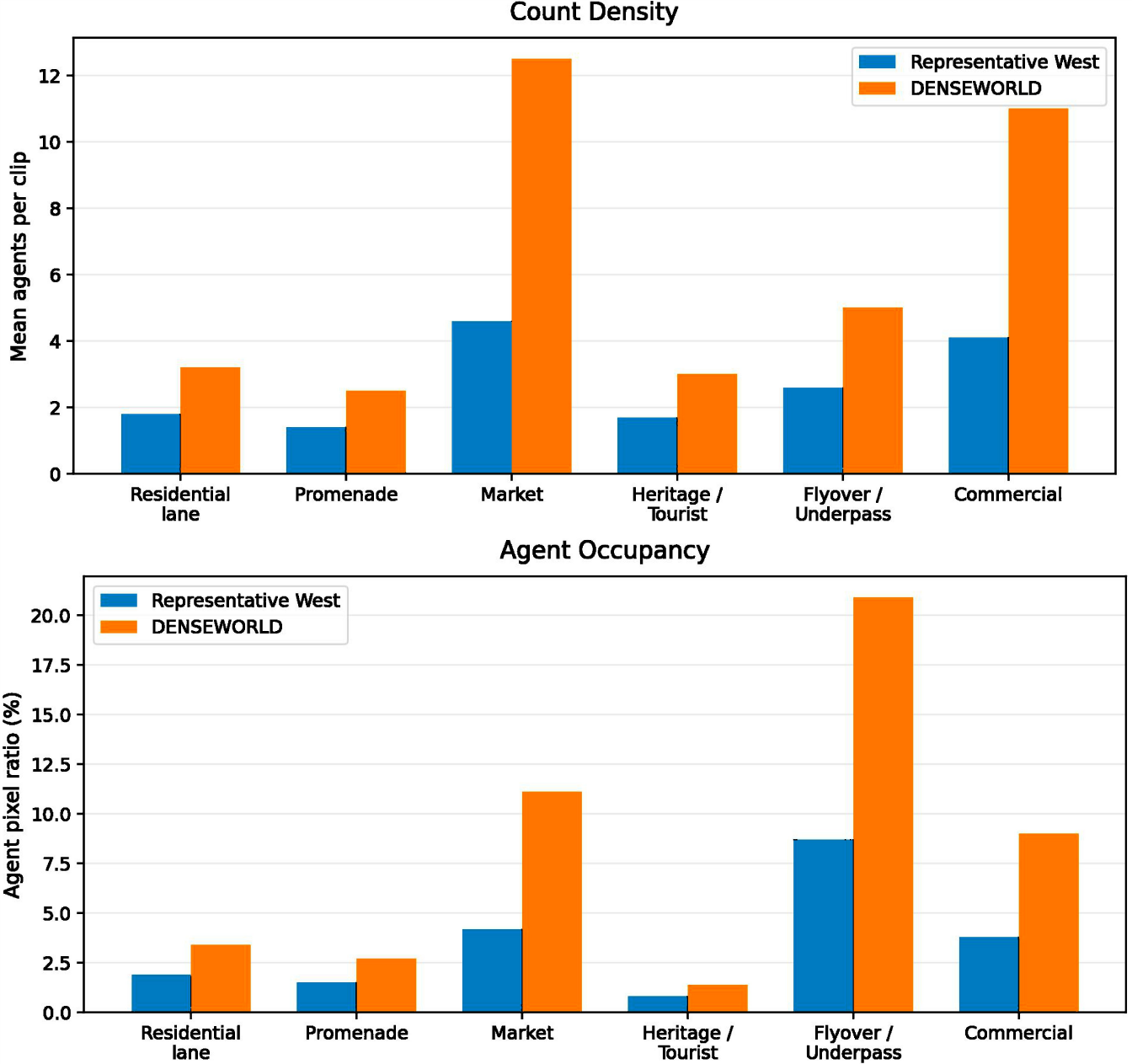}
    \vspace{-0.5em}
    \caption{
\textbf{DENSEWORLD exhibits higher multi-agent density than standard driving benchmarks.}
We compare \textbf{agent count density} and \textbf{agent occupancy} across matched scene types. DENSEWORLD shows large gaps in \textbf{market}, \textbf{commercial}, and \textbf{flyover/underpass} scenes, indicating stronger interaction pressure and heavier visual competition.
}
\label{fig:denseworld_vs_west_density}
\vspace{-1.2em}
\end{figure}

To make this regime measurable, we compare DENSEWORLD against \textbf{BDD100K} and \textbf{nuScenes}, two widely used benchmarks for autonomous-driving perception and scene understanding~\cite{yu2020bdd100k,caesar2020nuscenes}. The goal is not to claim that DENSEWORLD scenes merely look different, but to test whether they occupy a quantitatively distinct operating regime for world-model evaluation.

We measure this regime along five axes: \textbf{agent count density}, \textbf{agent occupancy}, \textbf{occlusion pressure}, \textbf{interaction pressure}, and \textbf{agent heterogeneity}. These metrics capture, respectively, the number of dynamic agents, the fraction of image area they occupy, the frequency of partial or heavy occlusion, the density of spatially proximate agent pairs, and the diversity of co-occurring actor categories. Together, they quantify the multi-agent load, visual congestion, visibility degradation, local interaction structure, and traffic diversity of each scene.

For an auditable comparison, all statistics are computed after mapping labels to a \textbf{shared dynamic-agent taxonomy} and matching comparable scene categories across datasets. 
%The full taxonomy mapping and scene-matching protocol are provided in Appendix~\ref{app:taxonomy_mapping}.
Figure~\ref{fig:denseworld_vs_west_density} shows that DENSEWORLD has consistently higher density and occupancy across matched categories, with the largest gaps in \textbf{market}, \textbf{commercial}, and \textbf{flyover/underpass} scenes. These results support the central premise of the benchmark: DENSEWORLD is not merely a geographic extension of existing driving data; it defines a higher-density, higher-occlusion, and higher-interaction regime for evaluating predictive world models.

\begin{figure*}[t]
    \centering
    \includegraphics[width=0.82\textwidth]{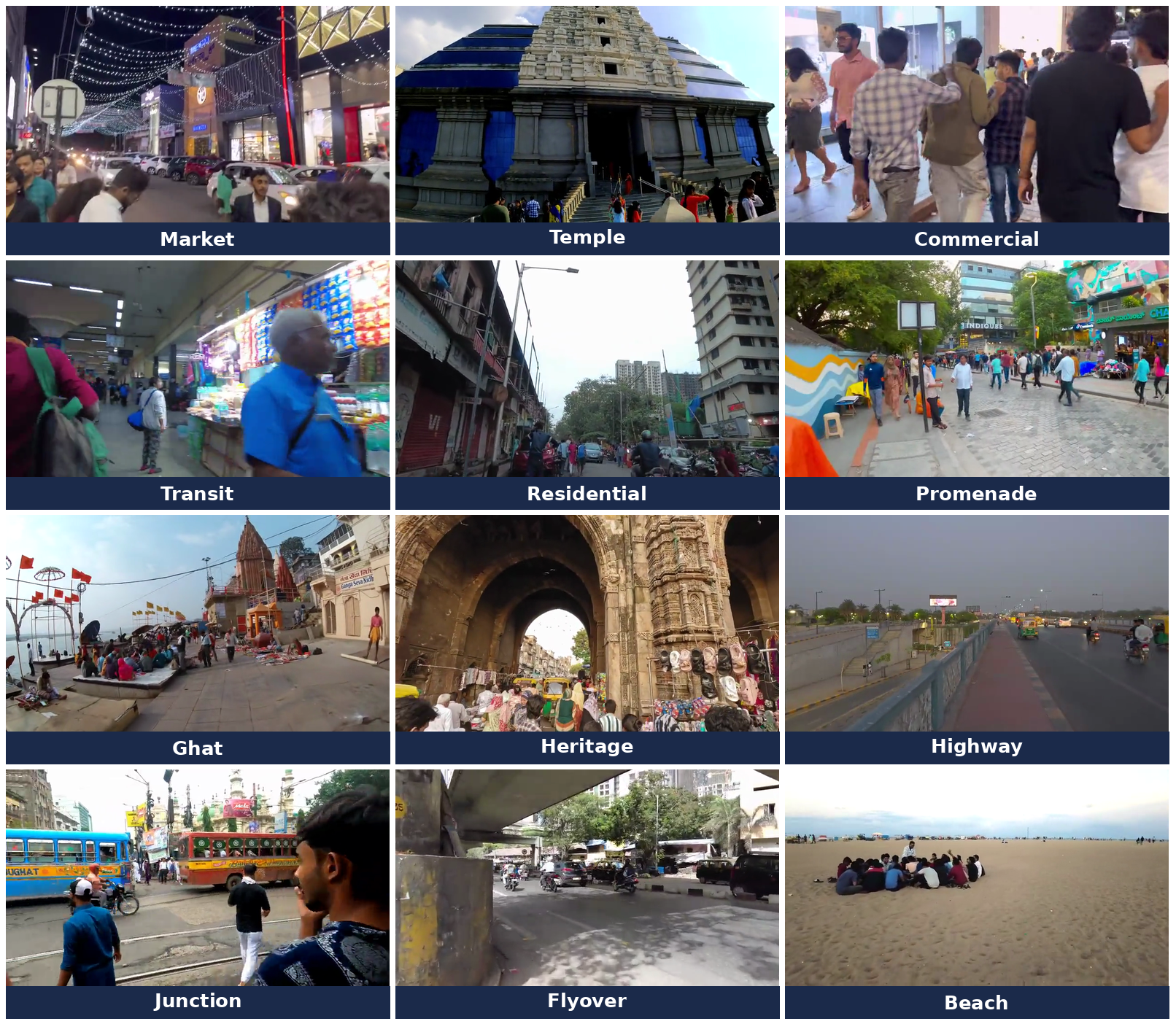}
    \caption{\textbf{DENSEWORLD 1.0 scene-type coverage.} Twelve representative scene types (market, temple, commercial, transit, residential, promenade, ghat, heritage, highway, junction, flyover, beach) across 22 Indian cities, shown from ground-level pedestrian viewpoints.}
    \label{fig:denseworld_scene_types_full}
\end{figure*} %Full automatic-annotation details, including thresholds, annotation taxonomy, export format, and loss mapping, are provided in Appendix~\ref{app:annotation_protocol}.

\section{Can Fine-Tuning Close the \textbf{DENSEWORLD} Gap?}
\label{sec:vjepa_adaptation_baselines}

Before introducing FactorJEPA, we ask whether conventional adaptation can recover the predictive structure required by DENSEWORLD without reorganizing the JEPA predictor. We evaluate three complementary strategies: \textbf{LoRA}~\cite{hu2022lora}, \textbf{DoRA}~\cite{liu2024dora}, and \textbf{Auto-RGN}~\cite{lee2023surgical}. All use the same raw clips, optimization steps, optimizer, masking policy, and evaluation protocol; only the parameter-update mechanism changes.

% =====================================================================
% Frozen-encoder scorecard — Table "Frozen encoders collapse".
% SINGLE SOURCE for this table: \input{tab_frozen_scorecard} wherever it belongs.
% NEVER copy-paste the numbers — moving the block by re-typing risks transcription drift.
% Values VERIFIED (2026-08-01) against the source CSV — all 30 cells match (rounded):
%   iter/iter17_ablations_model/result_outputs/v17a_frozen_eval/poc/probe_plot/eval/m13_frozen_scorecard.csv
%   (action_top1 -> A rounded to 0.1; motion_cos -> M and taxonomy_f1 -> T to 3 dp)
% Shading: per column, best = green!35, worst = red!35 (A higher=better, M higher=better, T higher=better).
% =====================================================================
\begin{table*}[t]
\centering
\caption{\textbf{\emph{Frozen encoders collapse.}} Ten frozen encoders on the DENSEWORLD motion probe ($n_{\mathrm{test}}=1{,}825$, $\pm$95\% BCa CI). Action top-1 (A) stays in a $37.5$--$44.4\%$ band (19.5\% majority), while adapted encoders reach $50.3$--$53.2\%$. M~=~motion-cos, T~=~taxonomy F1 ($\uparrow$ better). In each column the \textbf{best} value is shaded green and the \textbf{worst} red.}
\label{tab:frozen_scorecard_full}
\footnotesize\bfseries
\setlength{\tabcolsep}{16pt}
\renewcommand{\arraystretch}{1.18}
\begin{tabular}{@{}lccc@{}}
\toprule
Frozen encoder & A\,(\%) & M & T \\
\midrule
V-JEPA 2.1 (2B)   & \cellcolor{green!35}44.4 & 0.009 & 0.793 \\
V-JEPA 2.1 ViT-L  & 44.2 & \cellcolor{red!35}0.004 & 0.788 \\
V-JEPA 1 ViT-H    & 40.5 & 0.007 & 0.702 \\
LeJEPA ViT-L      & 40.1 & 0.014 & 0.740 \\
V-JEPA 1 ViT-L    & 39.9 & 0.008 & \cellcolor{red!35}0.660 \\
I-JEPA ViT-H      & 39.1 & 0.016 & 0.781 \\
V-JEPA 2.0 (SSv2) & 38.8 & 0.007 & 0.776 \\
DINOv2            & 38.5 & 0.016 & \cellcolor{green!35}0.816 \\
V-JEPA 2 ViT-L    & 37.9 & 0.013 & 0.778 \\
I-JEPA ViT-G/16   & \cellcolor{red!35}37.5 & \cellcolor{green!35}0.019 & 0.787 \\
\bottomrule
\end{tabular}
\end{table*}
   % single-source table (numbers verified vs m13_frozen_scorecard.csv)

\noindent Off-the-shelf frozen encoders collapse into a narrow $37.5$--$44.4\%$ band on the motion probe (Table~\ref{tab:frozen_scorecard_full}), and no single model wins every column, so frozen reuse is not enough.

Let $x$ denote a video clip; $M_c$ and $M_t$ the context and target masks; $f_\theta$ the online encoder; $\bar f_{\bar\theta}$ the momentum target encoder; and $g_\phi$ the predictor. All baselines retain the executed V-JEPA objective:
\[
\mathcal{L}_{\mathrm{JEPA}}
=
\mathbb{E}_{x}
\left[
\left\|
g_\phi\!\left(f_\theta(M_c\odot x),M_t\right)
-
\operatorname{sg}\!\left(
\bar f_{\bar\theta}(M_t\odot x)
\right)
\right\|_2^2
\right],
\]
where $\operatorname{sg}$ denotes stop-gradient. This data- and step-matched protocol isolates adaptation capacity without introducing a different prediction target or training signal.

\begin{table}[ht!]
\centering
\caption{
\textbf{Protocol-matched V-JEPA adaptation baselines.}
All methods retain the JEPA objective and train on the same raw clips; only the adaptation mechanism changes.
}
\label{tab:vjepa_adaptation_baselines}
\scriptsize
\setlength{\tabcolsep}{3pt}
\renewcommand{\arraystretch}{1.08}
\begin{tabular}{
    p{0.18\columnwidth}
    p{0.40\columnwidth}
    p{0.33\columnwidth}
}
\toprule
\textbf{Method} & \textbf{Adaptation mechanism} & \textbf{Question tested} \\
\midrule

\textbf{LoRA}~\cite{hu2022lora}
&
Adds low-rank updates,
$\Delta W=(\alpha/r)BA$, to selected frozen projections.
&
Is parameter-efficient low-rank adaptation sufficient?
\\
\midrule

\textbf{DoRA}~\cite{liu2024dora}
&
Separates weight magnitude from direction and applies low-rank updates to the directional component.
&
Does weight decomposition recover structure missed by LoRA?
\\
\midrule

\textbf{Auto-RGN}~\cite{lee2023surgical}
&
Selects transformer blocks using their relative gradient norms and updates only the selected subset.
&
Can gradient-guided selection localize the required adaptation?
\\

\bottomrule
\end{tabular}
\end{table}

For LoRA, an adapted weight matrix is
\[
W' = W_0 + \frac{\alpha}{r}BA,
\qquad
B\in\mathbb{R}^{d_{\mathrm{out}}\times r},
\quad
A\in\mathbb{R}^{r\times d_{\mathrm{in}}},
\]
where $r\ll\min(d_{\mathrm{in}},d_{\mathrm{out}})$. DoRA further separates the magnitude and direction of each weight vector:
\[
W'
=
m\odot
\frac{W_0+\Delta W}
{\left\|W_0+\Delta W\right\|},
\qquad
\Delta W=\frac{\alpha}{r}BA,
\]
allowing directional adaptation without coupling it to weight magnitude.

Following the relative-gradient-norm criterion of \citet{lee2023surgical}, our blockwise Auto-RGN implementation scores transformer block $\ell$ as
\[
s_\ell
=
\frac{
\left\|
\nabla_{\theta_\ell}\mathcal{L}_{\mathrm{JEPA}}
\right\|_2
}{
\left\|\theta_\ell\right\|_2+\epsilon
},
\qquad
\mathcal{S}_K
=
\operatorname{TopK}_{\ell}(s_\ell),
\]
and updates only $\{\theta_\ell:\ell\in\mathcal{S}_K\}$. Normalization by parameter magnitude prevents larger blocks from being favored solely because of scale. LoRA and DoRA target the same projection families, while Auto-RGN receives a matched trainable-parameter budget. 
\begin{figure*}[t]
    \centering

    % -------------------- Agents -------------------- %
    \begin{minipage}[t]{0.325\textwidth}
        \centering
        \includegraphics[
            width=\linewidth,
            height=0.225\textheight,
            keepaspectratio
        ]{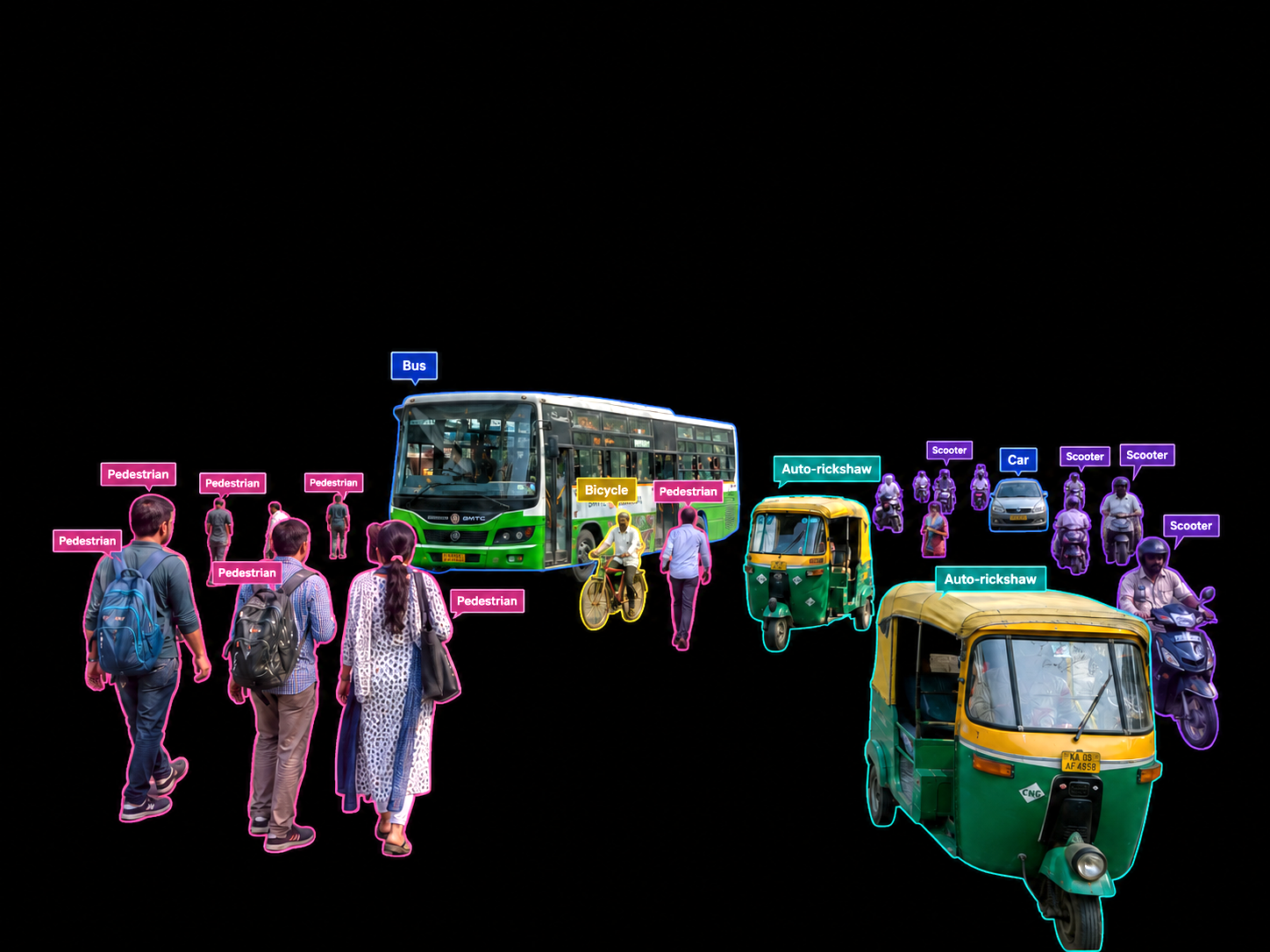}
        %\vspace{0.5mm}

        {\small\textbf{(a) Agents}}
    \end{minipage}
    \hfill
    % -------------------- Layout -------------------- %
    \begin{minipage}[t]{0.325\textwidth}
        \centering
        \includegraphics[
            width=\linewidth,
            height=0.225\textheight,
            keepaspectratio
        ]{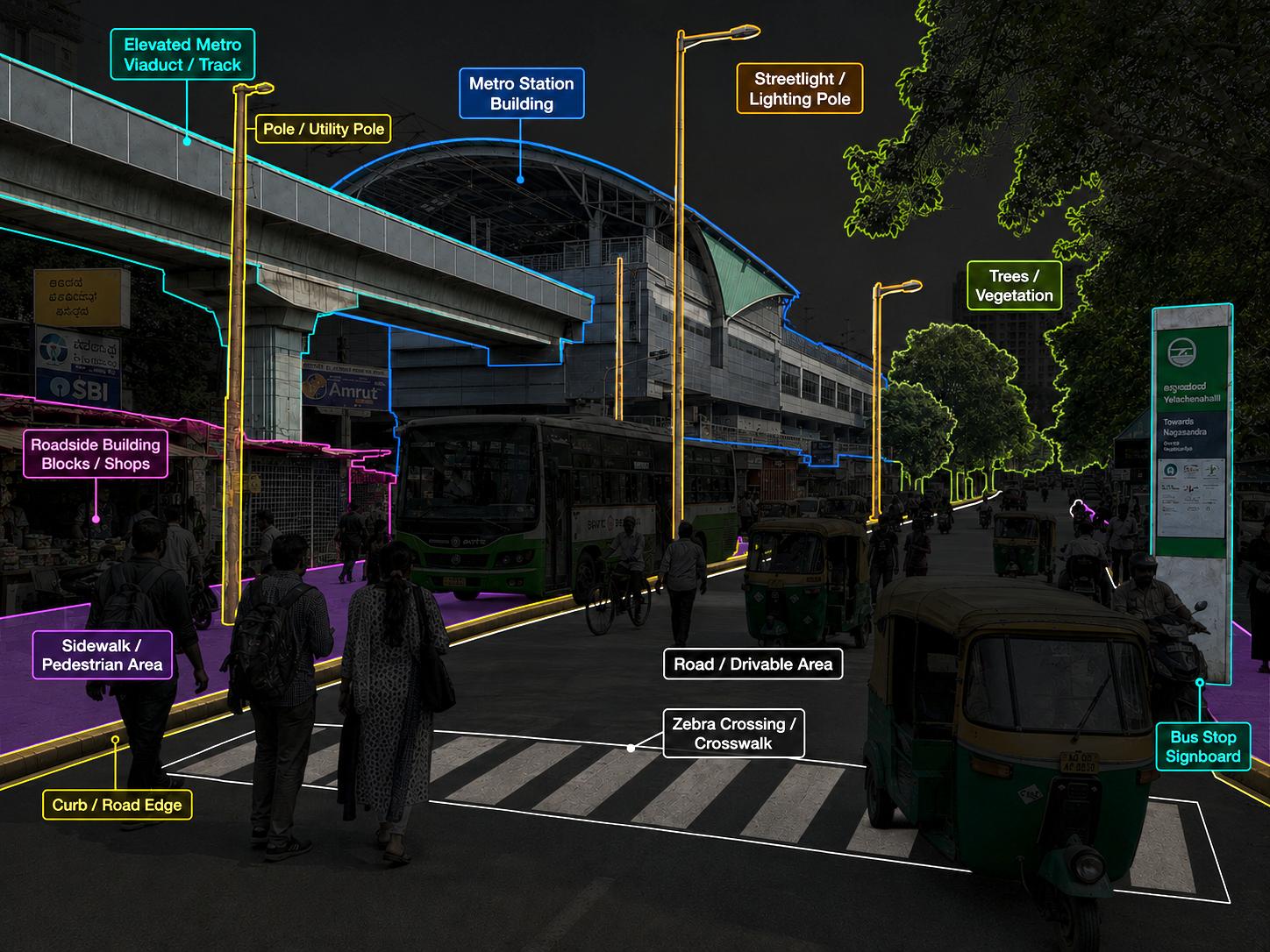}
        %\vspace{0.5mm}

        {\small\textbf{(b) Layout}}
    \end{minipage}
    \hfill
    % -------------------- Interactions -------------------- %
    \begin{minipage}[t]{0.325\textwidth}
        \centering
        \includegraphics[
            width=\linewidth,
            height=0.225\textheight,
            keepaspectratio
        ]{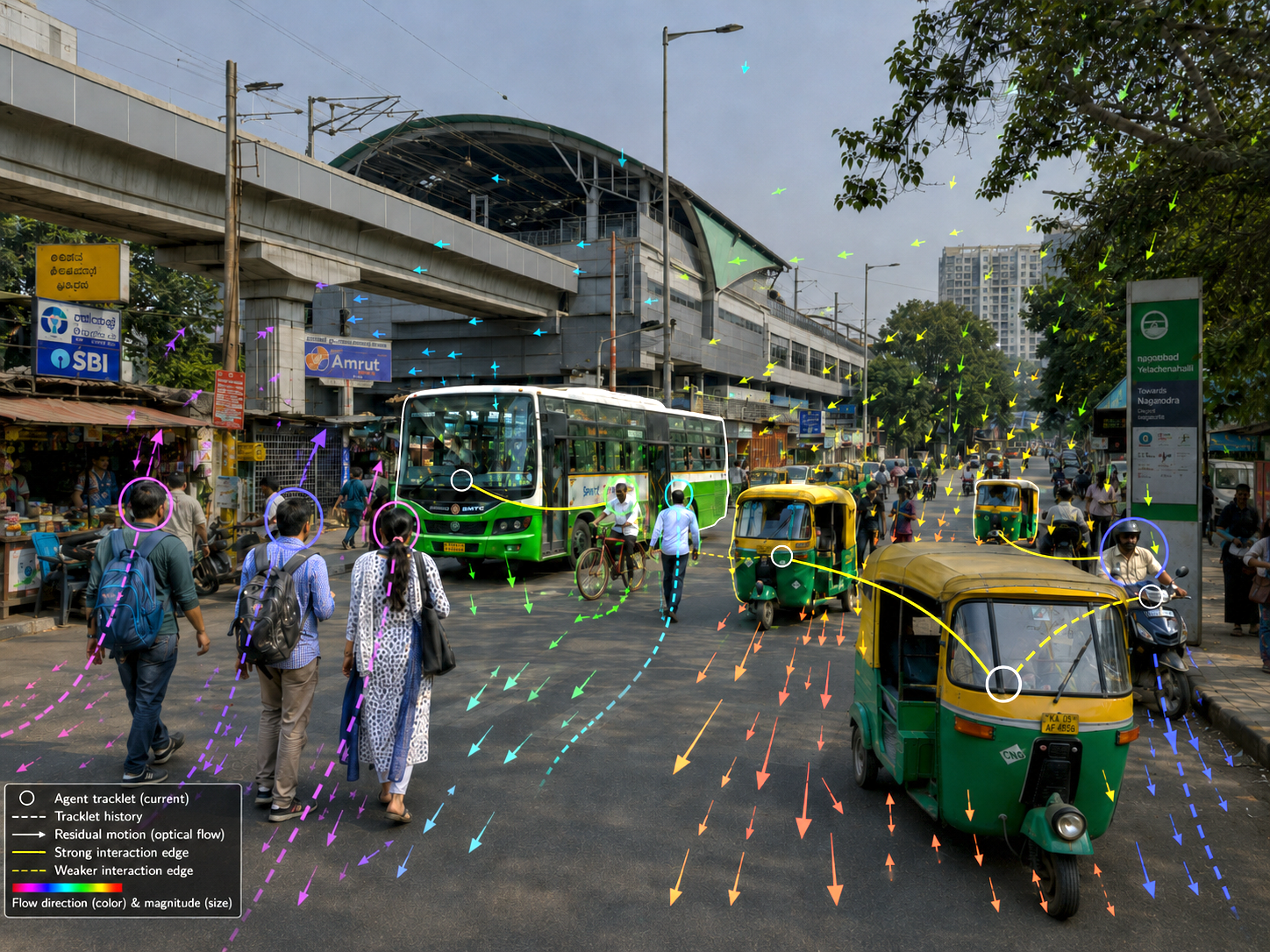}
        %\vspace{0.5mm}

        {\small\textbf{(c) Interactions}}
    \end{minipage}

    \vspace{-1em}

    \caption{
\textbf{Factorized decomposition of the same urban scene.}
\textbf{(a)} Agents are isolated from scene context.
\textbf{(b)} Layout highlights persistent spatial structure while retaining
suppressed agent silhouettes.
\textbf{(c)} Interactions show tracklets, residual motion, and sparse pairwise
coupling; arrow direction and length encode motion direction and magnitude,
while solid and dashed edges denote stronger and weaker interactions.
}
    \label{fig:factor_scene_decomposition}
\end{figure*}

\section{\textbf{\emph{FactorJEPA}: Explicitly Factorized Predictive Channels}}
\label{sec:factorjepa}

\paragraph{\textbf{\emph{Motivation.}}}
Let $x$ be a video clip, $M_c$ and $M_t$ its context and target masks,
$f_\theta$ the online encoder, and $\bar f_{\bar\theta}$ the momentum target
encoder. The context representation and stop-gradient future target are

\begin{equation*}
\resizebox{0.98\columnwidth}{!}{$\displaystyle
h_t=f_\theta(M_c\odot x),
\qquad
Y_{t+\Delta}^{\star}
=
\operatorname{sg}\!\left(
\Pi_{M_t}[\bar f_{\bar\theta}(x)]
\right)
\in\mathbb{R}^{m\times d}
$}
\end{equation*}

where $\Pi_{M_t}$ denotes the executed target-selection operator, $m$ the
number of target representations, and $d$ their embedding dimension. A
conventional JEPA predictor learns

\begin{equation*}
g_\phi:(h_t,M_t)
\longmapsto
\widehat{Y}_{t+\Delta}
\in\mathbb{R}^{m\times d}
\end{equation*}

by matching $\widehat{Y}_{t+\Delta}$ to $Y_{t+\Delta}^{\star}$. The objective
specifies \emph{what} to predict, but leaves \emph{how predictive information
is internally organized} unconstrained.

This ambiguity is consequential in \textbf{DENSEWORLD}, where layout, agent
density, visibility, and interaction pressure are strongly correlated. A
high-capacity predictor can exploit \textbf{\emph{shortcut mixtures}}---crowd
texture as a proxy for interaction, visible appearance for object persistence,
or road geometry for motion---without recovering the structure governing scene
evolution.

\textbf{\emph{FactorJEPA}} resolves this ambiguity by replacing the monolithic
predictor with explicit \textbf{\emph{layout}}, \textbf{\emph{agent}}, and
\textbf{\emph{interaction}} channels. A \textbf{\emph{soft visibility gate}}
attenuates uncertain observations, while block-structured regularization limits
\textbf{\emph{cross-factor leakage}}. Together, these components factorize the
future JEPA embedding into semantically anchored coordinates and
factor-specific predictive subspaces.

\subsection{\textbf{\emph{DINOv2-Based Segmentations Agent-Layout-Interactions}}}
\label{sec:factor_target_construction}

For each privacy-filtered clip $x_n$, a frozen DINOv2 pipeline
\cite{oquab2024dinov2} extracts region masks, boxes, descriptors, and
confidences, which are temporally associated into tracklets:
\[
\mathcal P_n
=
\mathcal D_{\mathrm{pre}}
\!\left(\mathcal F_{\mathrm{DINOv2}}(x_n)\right),
\qquad
\mathcal T_n=\mathcal A(\mathcal P_n).
\]
A deterministic structural map converts region geometry, visibility, temporal
continuity, and relative motion into layout, agent, visibility, and interaction
targets with reliability weights:
\[
\Psi_{\mathrm{struct}}(\mathcal P_n,\mathcal T_n)
\longmapsto
\left\{(T_{n,k},q_{n,k})\right\}_{k\in\{L,A,V,I\}} .
\]
Factor-specific heads $\widehat T_{n,k}=P_k(Z_{n,k})$ are trained using
\[
\mathcal L_{\mathrm{factor}}
=
\sum_{k\in\{L,A,I\}}
\frac{
\sum_{n=1}^{N} q_{n,k}\,
\ell_k(\widehat T_{n,k},T_{n,k})
}{
\epsilon+\sum_{n=1}^{N}q_{n,k}
},
\]
while visibility is optimized separately through $\mathcal L_v$.

DINOv2-derived targets supervise training only and are never reused as
evaluation labels. Checkpoints, association and interaction rules, thresholds,
target coverage, and reliability audits are provided in
Appendix~\ref{app:factor_targets}.

\begin{figure*}[t]
    \centering
    \includegraphics[width=\textwidth]{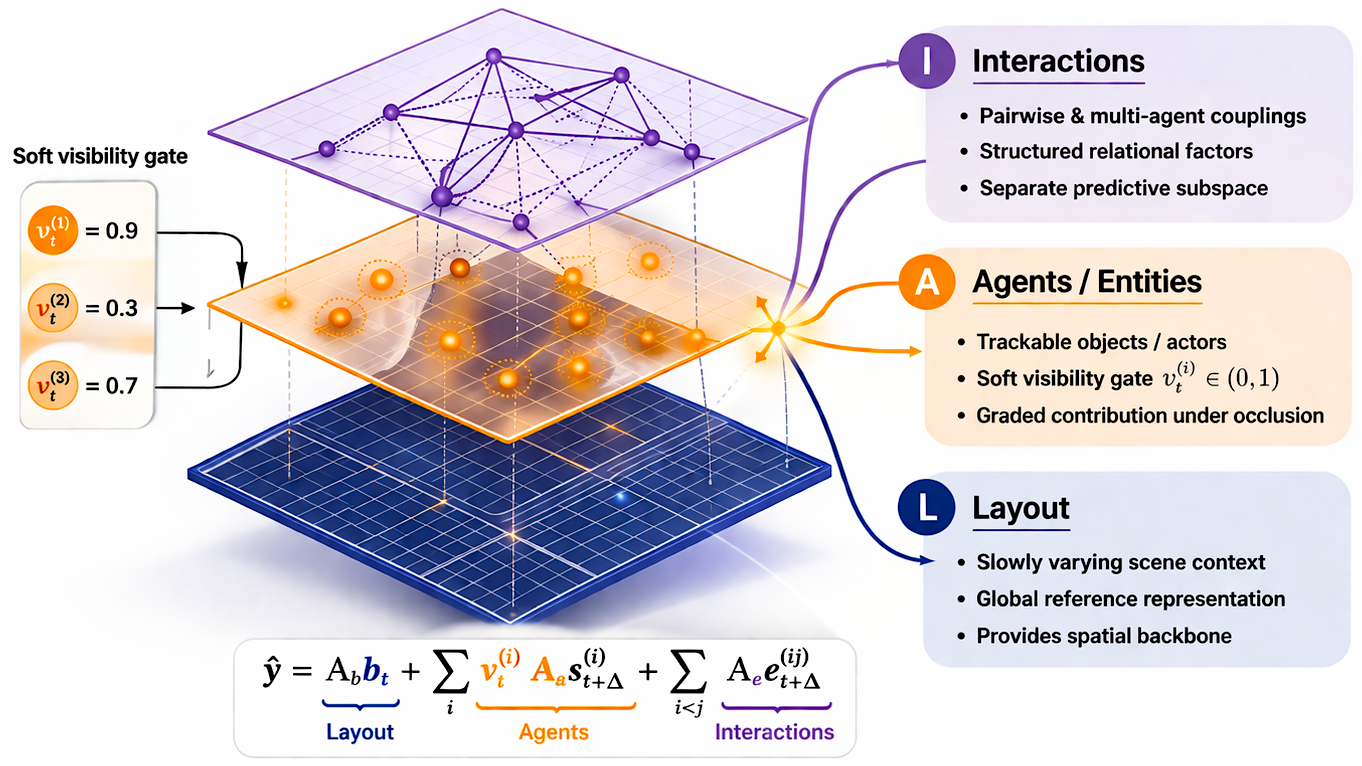}
    \caption{
    \textbf{\emph{Factorized prediction.}}
    \textbf{\emph{FactorJEPA}} composes the target embedding from
    \textbf{\emph{layout}}, \textbf{\emph{agents}}, and
    \textbf{\emph{interactions}}, with a soft visibility gate applied only to
    entity terms.
    }
    \label{fig:factorjepa_factors}
\end{figure*}

\begin{figure*}[t]
    \centering
    \includegraphics[width=\textwidth]{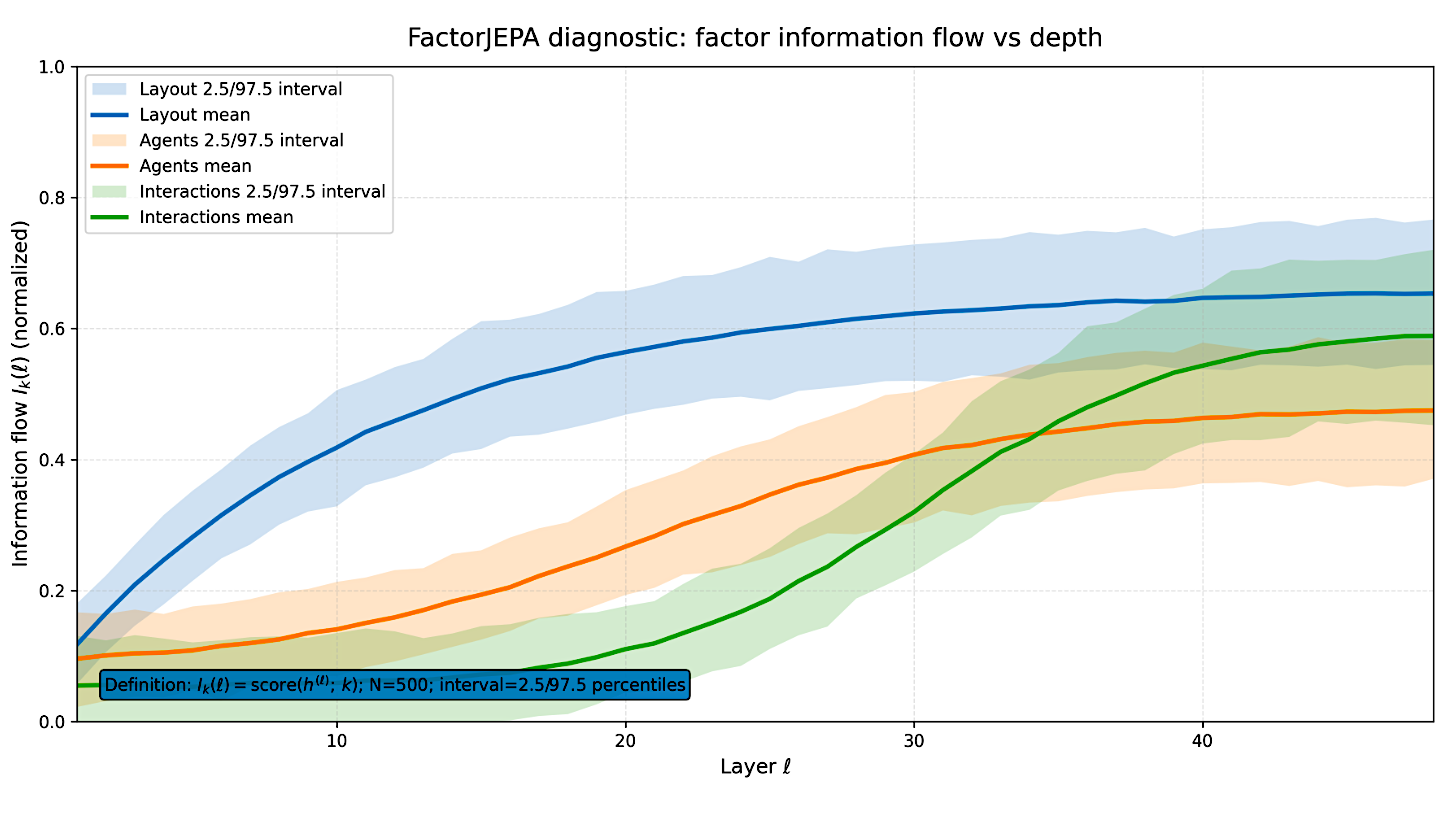}
    \caption{
    \textbf{\emph{Depth-wise factor realization.}}
    Lightweight probes show a staged profile: \textbf{\emph{layout}}
    emerges early, \textbf{\emph{agents}} grow gradually, and
    \textbf{\emph{interactions}} peak in deeper layers.
    }
    \label{fig:factorjepa_channels}
\end{figure*}

\noindent This factorization is not only architectural: lightweight probes show the three channels emerge at distinct network depths (Figure~\ref{fig:factorjepa_channels}).

\subsection{\textbf{\emph{Structured Factor Coordinates}}}

For target token $p$, FactorJEPA forms the conditioned query
\[
q_{n,p}=Q\!\left(f_\theta(M_c\odot x_n),\pi_p,M_t\right)
\]
and predicts
\[
c_{n,p}
=
\operatorname{col}
\!\left(c_{n,L,p},c_{n,A,p},c_{n,I,p}\right)
\in\mathbb R^{r_L+r_A+r_I}.
\]
We suppress $p$ below and write $h_n=q_{n,p}$.

\paragraph{\textbf{\emph{Layout.}}}
Slowly varying spatial support is encoded as
\[
c_{n,L}=g_L(h_n),
\qquad
\widehat T_{n,L}=P_L(c_{n,L}).
\]

\paragraph{\textbf{\emph{Visibility-gated agents.}}}
For agent representation $o_n^{(i)}$,
\[
s_n^{(i)}=g_A(h_n,o_n^{(i)}),
\qquad
v_n^{(i)}=\sigma\!\left(g_V(h_n,o_n^{(i)})\right),
\]
and
\[
c_{n,A}
=
\frac{\sum_i v_n^{(i)}s_n^{(i)}}
{\epsilon+\sum_i v_n^{(i)}}.
\]
The gate softly suppresses uncertain or occluded agents without removing them.
Given visibility targets $T_{n,V}^{(i)}$ and reliabilities $q_{n,V}^{(i)}$,
\[
\mathcal L_v
=
-
\frac{
\sum_{n,i}q_{n,V}^{(i)}
\left[
T_{n,V}^{(i)}\log v_n^{(i)}
+
(1-T_{n,V}^{(i)})\log(1-v_n^{(i)})
\right]
}{
\epsilon+\sum_{n,i}q_{n,V}^{(i)}
}.
\]

\paragraph{\textbf{\emph{Sparse interactions.}}}
For each candidate pair $(i,j)\in\mathcal E_n$,
\[
e_n^{(ij)}
=
g_I(h_n,o_n^{(i)},o_n^{(j)}),
\qquad
\bar e_n^{(ij)}
=
\frac{e_n^{(ij)}}{\epsilon+\|e_n^{(ij)}\|_2},
\]
with soft interaction strength
\[
w_n^{(ij)}
=
\sigma\!\left(g_W(h_n,o_n^{(i)},o_n^{(j)})\right).
\]
Let $M_n=\max\{1,|\mathcal E_n|\}$. Then
\[
c_{n,I}
=
\frac{1}{M_n}
\sum_{(i,j)\in\mathcal E_n}
w_n^{(ij)}\bar e_n^{(ij)},
\qquad
c_{n,I}=0
\ \text{if}\ 
\mathcal E_n=\varnothing,
\]
and localized interactions are encouraged through
\[
\mathcal L_{\mathrm{sparse}}
=
\frac{1}{N}
\sum_{n=1}^{N}
\frac{1}{M_n}
\sum_{(i,j)\in\mathcal E_n}
w_n^{(ij)}.
\]
The fixed normalization prevents trivial rescaling of interaction weights and
states.

\subsection{\textbf{\emph{Block-Structured Matrix Factorization}}}

For each factor $k\in\{L,A,I\}$, we stack all target-token coordinates across
the minibatch:
\[
C_k
=
\begin{bmatrix}
(c_{1,k,1})^\top\\[-1mm]
\vdots\\
(c_{1,k,m})^\top\\
\vdots\\
(c_{N,k,m})^\top
\end{bmatrix}
\in\mathbb{R}^{Nm\times r_k}.
\]
The factor coordinates and learned synthesis dictionaries are concatenated as
\[
C=
\begin{bmatrix}
C_L & C_A & C_I
\end{bmatrix},
\qquad
A=
\begin{bmatrix}
A_L & A_A & A_I
\end{bmatrix},
\qquad
A_k\in\mathbb{R}^{d\times r_k}.
\]

FactorJEPA predicts the complete token-level target matrix through
\[
\boxed{
\widehat Y
=
\underbrace{
\begin{bmatrix}
C_L & C_A & C_I
\end{bmatrix}
}_{\text{factor coordinates}}
\underbrace{
\begin{bmatrix}
A_L & A_A & A_I
\end{bmatrix}^{\!\top}
}_{\text{synthesis dictionaries}}
=
CA^\top
}
\]
or, equivalently,
\[
\widehat Y
=
\underbrace{C_LA_L^\top}_{Y_L}
+
\underbrace{C_AA_A^\top}_{Y_A}
+
\underbrace{C_IA_I^\top}_{Y_I}.
\]
Distinct pathways, factor supervision, and dictionaries make the decomposition
architectural rather than post hoc.

For stacked future targets
\[
Y^\star
=
\operatorname{col}
\left(
Y_{1,t+\Delta}^\star,\ldots,Y_{N,t+\Delta}^\star
\right)
\in\mathbb{R}^{Nm\times d},
\]
the JEPA objective is
\[
\mathcal L_{\mathrm{JEPA}}
=
\frac{1}{Nm}
\left\|
\widehat Y-Y^\star
\right\|_F^2.
\]
The corresponding $m$ rows are unstacked per clip, preserving the complete
future-token geometry.

\begin{figure*}[t]
    \centering
    \includegraphics[width=\textwidth]{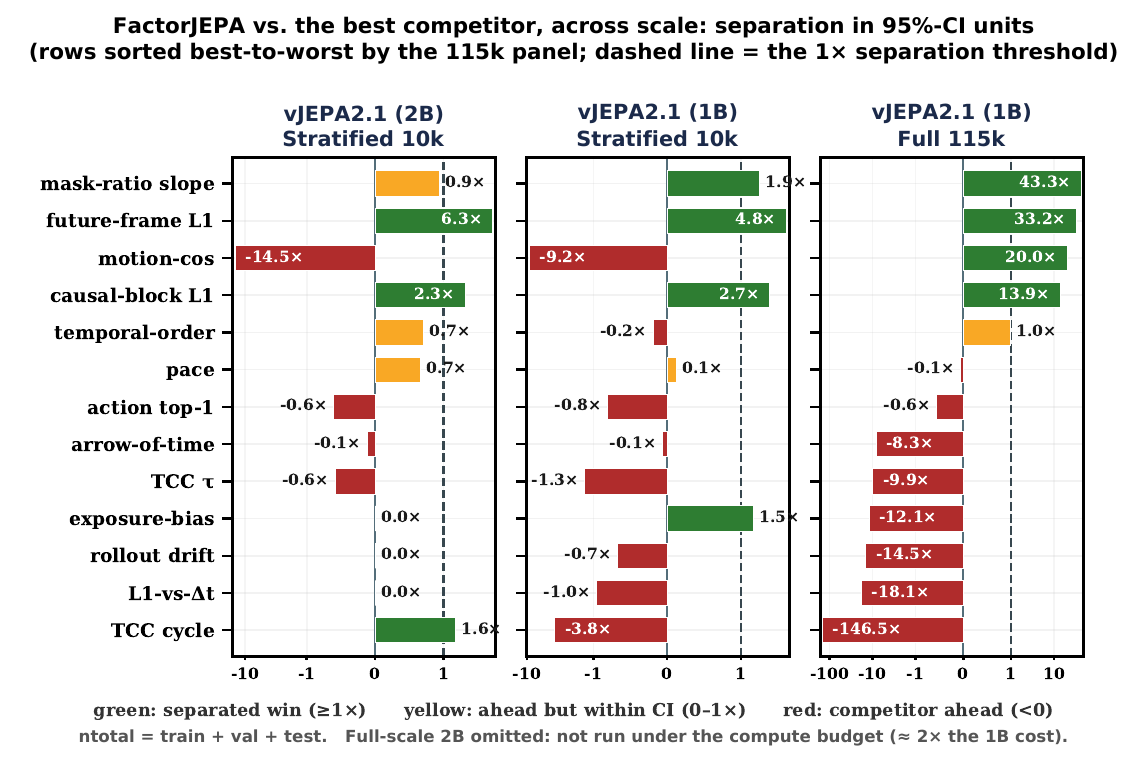}
    \caption{
\textbf{\emph{FactorJEPA vs. the strongest competitor across scale
and data regimes.}}
Bars report FactorJEPA's advantage in units of the paired-difference
95\% confidence interval; the dashed line marks statistical separation at
$1\times$. Under matched stratified 10k protocol, FactorJEPA separates on
Future-frame L1 and Causal L1 at both scales, and on Mask-ratio slope at 1B,
while trailing on Motion cosine (revealing a consistent prediction--motion
trade-off). With full 115k-clip training at 1B, it separates strongly on all
four primary diagnostics, reaching $43.3\times$ for Mask-ratio slope,
$33.2\times$ for Future-frame L1, $20.0\times$ for Motion cosine, and
$13.9\times$ for Causal L1.
}
    \label{fig:frozen_forest}
\end{figure*}

\begin{figure*}[t]
    \centering
    \includegraphics[width=\textwidth]{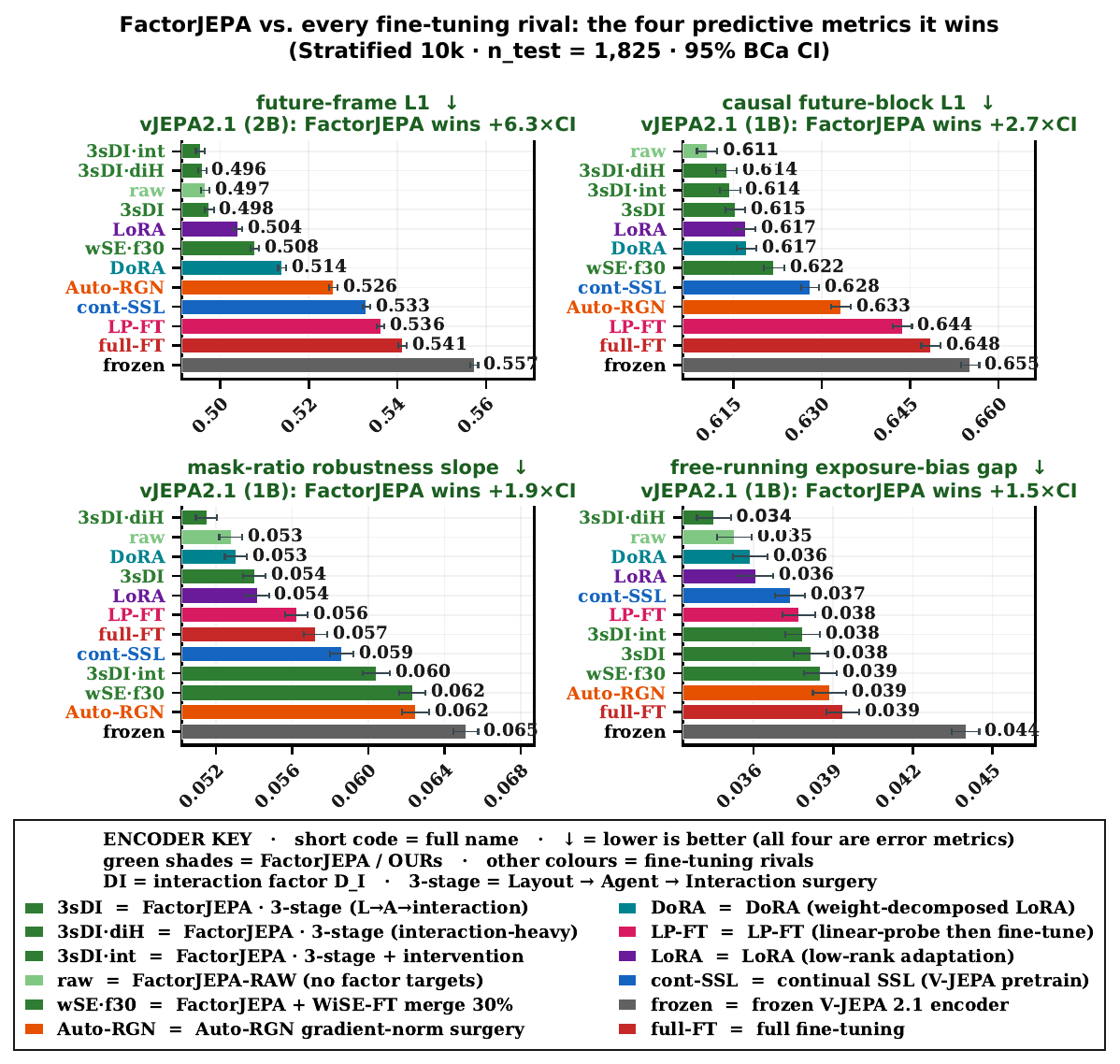}
    \caption{
\textbf{\emph{Evaluation scorecard across 2B and 1B scales.}}
Frozen, conventionally adapted, parameter-efficient, and factorized
V-JEPA~2.1 variants are compared across predictive, motion, semantic,
and temporal diagnostics. Bars report performance with 95\% BCa
intervals where available.
}
    \label{fig:eval_scorecard}
\end{figure*}

\noindent The head-to-head advantage above holds up when FactorJEPA is placed against every adaptation family at once: the full scorecard (Figure~\ref{fig:eval_scorecard}) spans predictive, motion, semantic, and temporal diagnostics at both scales.

\subsection{\textbf{\emph{Semantic Anchoring and Channel Separation}}}

The factorization $\widehat Y=CA^\top$ is non-unique: for any invertible
$R$,
\[
CA^\top=(CR)(AR^{-\top})^\top.
\]
Thus, the JEPA objective alone permits rotations and cross-block mixing.
Factor-specific heads
\[
\widehat T_k=C_kP_k^\top,
\qquad k\in\{L,A,I\},
\]
together with $\mathcal L_{\mathrm{factor}}$, anchor the blocks to layout,
agents, and interactions. FactorJEPA therefore claims semantic block
separation, not coordinate-level identifiability.

To suppress residual cross-channel shortcuts, let
$\Gamma_{kk'}$ denote the empirical covariance between channel outputs
$Y_k$ and $Y_{k'}$. We penalize linear and nonlinear leakage through
\[
\mathcal L_{\mathrm{sep}}
=
\underbrace{
2\sum_{k<k'}\|\Gamma_{kk'}\|_F^2
}_{\mathcal L_{\mathrm{cov}}}
+
\beta_{\mathrm{nlin}}
\underbrace{
\sum_{k<k'}
\frac{
\operatorname{tr}(\bar K_k\bar K_{k'})
}{
\|\bar K_k\|_F\|\bar K_{k'}\|_F+\epsilon
}
}_{\mathcal L_{\mathrm{nlin}}},
\]
where $\bar K_k$ is the centered RBF Gram matrix for channel $k$.
This suppresses cross-channel dependence while leaving within-channel
variation unconstrained; it does not imply statistical or causal
independence.

\paragraph{\textbf{\emph{Leakage diagnostic.}}}
We freeze FactorJEPA and train fixed-capacity probes from channel $Y_k$ to
factor target $T_{k'}$. With held-out probe score $S_{k\rightarrow k'}$ and
constant-predictor score $S_{0\rightarrow k'}$, normalized leakage is
\[
\operatorname{Leak}(k\!\rightarrow\!k')
=
\frac{
S_{k\rightarrow k'}-S_{0\rightarrow k'}
}{
S_{k'\rightarrow k'}-S_{0\rightarrow k'}+\epsilon
},
\qquad k\neq k'.
\]
Effective separation requires strong diagonal predictability and low
off-diagonal leakage. Full definitions and probe protocols are provided in
Appendix~\ref{app:factor_architecture}.

\begin{comment}

\begin{figure*}[ht!]
    \centering
    \includegraphics[width=0.8\textwidth]
    {figures/eval_scorecard_winbars.pdf}
    \caption{
    \textbf{\emph{Complete evaluation scorecard at 2B and 1B scales.}}
    We compare frozen, conventionally adapted, parameter-efficient, and
    factorized V-JEPA~2.1 variants across 15 semantic, predictive, motion, and
    temporal-coherence diagnostics. Bars report test performance with 95\% BCa
    confidence intervals where available. The four headline diagnostics are
    Future-frame L1, Causal L1, Motion cosine, and Mask-ratio slope.
    }
    \label{fig:eval_scorecard}
    \vspace{-0.6em}
\end{figure*}
\end{comment}

\begin{figure}[ht]
    \centering
    \includegraphics[width=\columnwidth]
    {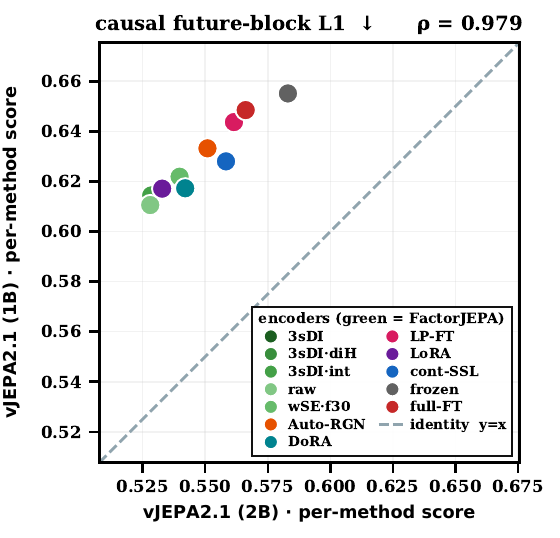}
    \caption{
    \textbf{\emph{Causal future-block rankings replicate across model scales.}}
    Each point represents one adaptation method, comparing its causal
    future-block L1 score with the ViT-G 2B backbone on the horizontal axis
    and the ViT-g 1B backbone on the vertical axis; lower values are better.
    FactorJEPA variants are shown in green, and the dashed line denotes
    equal scores across scales. The strong Spearman correlation
    ($\rho=0.979$) shows that the relative method ordering is highly
    preserved when scaling from 2B to 1B.
    }
    \label{fig:scale_replication}
    \vspace{-0.6em}
\end{figure}

\subsection{\textbf{\emph{Predictor Surgery}}}

FactorJEPA is initialized from a pretrained V-JEPA model. The online encoder,
momentum target encoder, target construction, and masking policy are retained.
Only the monolithic predictor is replaced:

\begin{equation*}
\Theta_{\mathrm{pred}}
\longrightarrow
\Theta_{\mathrm{fact}}
=
\left\{
\Theta_L,\Theta_A,\Theta_I,\Theta_V,\Theta_W,A
\right\}.
\end{equation*}

Training is restricted to the factorized predictor and the top $K$ encoder
blocks:

\begin{equation*}
\Theta_{\mathrm{train}}
=
\Theta_{\mathrm{fact}}
\cup
\Theta_{\mathrm{top}\text{-}K}.
\end{equation*}

The lower encoder blocks and momentum target network follow the frozen or
momentum-updated protocol of the underlying V-JEPA implementation.

The final objective is
\begin{equation*}
\boxed{
\begin{aligned}
\mathcal L
&=
\mathcal L_{\mathrm{JEPA}}
+
\lambda_{\mathrm{sep}}\mathcal L_{\mathrm{sep}}
+
\lambda_{\mathrm{sparse}}\mathcal L_{\mathrm{sparse}}\\
&\quad+
\lambda_v\mathcal L_v
+
\lambda_{\mathrm{sup}}\mathcal L_{\mathrm{factor}}.
\end{aligned}
}
\end{equation*}

Each term has a distinct role:
\begin{equation*}
\begin{array}{rcl}
\mathcal L_{\mathrm{JEPA}}
&:&
\text{future-latent prediction},\\
\mathcal L_{\mathrm{sep}}
&:&
\text{cross-channel leakage suppression},\\
\mathcal L_{\mathrm{sparse}}
&:&
\text{interaction locality},\\
\mathcal L_v
&:&
\text{visibility calibration},\\
\mathcal L_{\mathrm{factor}}
&:&
\text{semantic anchoring}.
\end{array}
\end{equation*}

\begin{figure*}[t]
    \centering
    \includegraphics[width=\textwidth]{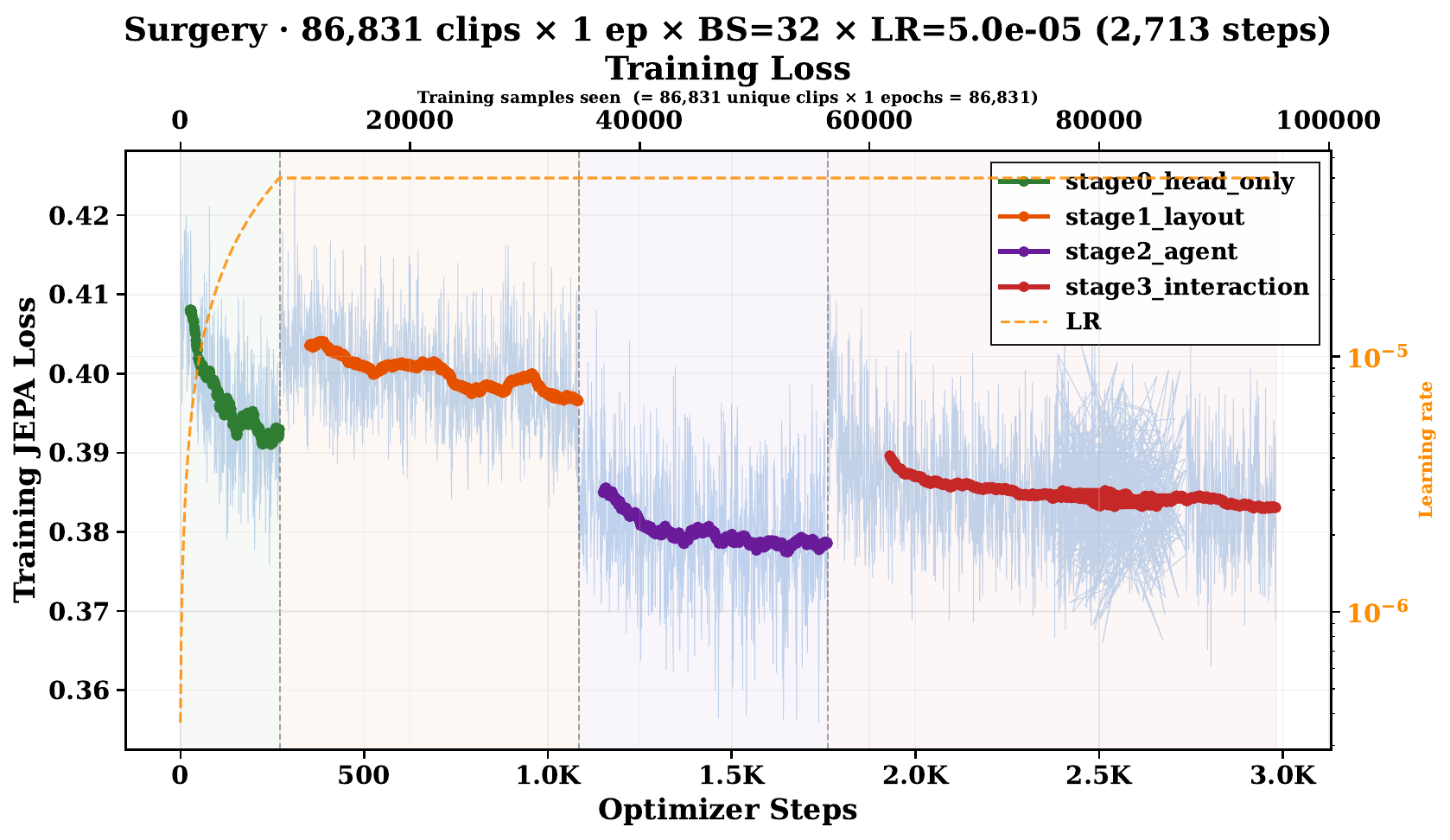}
    \caption{
    \textbf{\emph{Staged factor-curriculum surgery on the full corpus.}}
    Training JEPA loss for FactorJEPA predictor surgery on the full DENSEWORLD
    corpus ($\sim$115k clips; ViT-g 1B backbone, batch 32, learning rate
    $5{\times}10^{-5}$). Surgery proceeds in four shaded phases (dashed
    boundaries): a \textbf{\emph{head-only}} warmup, then progressive unfreezing
    of the top encoder blocks while the predictive emphasis cycles through
    \textbf{\emph{layout}}, \textbf{\emph{agents}}, and
    \textbf{\emph{interactions}}. The loss falls from $\approx 0.41$ to
    $\approx 0.38$; each transition briefly raises the loss as the target
    distribution shifts, after which training re-descends, with the interaction
    stage settling at a slightly higher floor consistent with its harder
    prediction target. The learning-rate schedule is dashed (right axis).
    }
    \label{fig:surgery_train_loss}
\end{figure*}

\noindent In practice, surgery runs as a staged factor curriculum (Figure~\ref{fig:surgery_train_loss}): a short head-only warmup precedes progressive unfreezing of the top-$K$ encoder blocks, while the predictive emphasis shifts from \textbf{\emph{layout}} to \textbf{\emph{agents}} to \textbf{\emph{interactions}}.

\subsection{\textbf{\emph{Depth-Wise Factor Realization}}}
\label{sec:depth_analysis}

Although the factorization acts at the predictor, we test how factor information
becomes linearly accessible across encoder depth. Let

\begin{equation*}
H^{(\ell)}
=
\begin{bmatrix}
h_1^{(\ell)}&\cdots&h_N^{(\ell)}
\end{bmatrix}^{\!\top}
\in\mathbb{R}^{N\times d_\ell}
\end{equation*}

denote the representations at layer $\ell$. For each factor
$k\in\{L,A,I\}$, we fit a regularized linear probe on the training split:

\begin{equation*}
\resizebox{0.98\columnwidth}{!}{$\displaystyle
W_k^{(\ell)}
=
\arg\min_W
\left\|
H_{\mathrm{train}}^{(\ell)}W-T_{k,\mathrm{train}}
\right\|_F^2
+
\lambda_{\mathrm{probe}}\|W\|_F^2
$}
\end{equation*}

The held-out depth-wise factor score is

\begin{equation*}
I_k(\ell)
=
\operatorname{score}_k
\left(
H_{\mathrm{test}}^{(\ell)}W_k^{(\ell)},
T_{k,\mathrm{test}}
\right).
\end{equation*}

Distinct onset and growth profiles indicate that layout, agent, and interaction
information becomes accessible at different depths. This is a representation
diagnostic, not a neuron-level or causal decomposition.

\section{\textbf{Experiments \& Evaluation}}
\label{sec:experiments}

We ask whether FactorJEPA improves V-JEPA's \emph{predictive structure}, rather
than only its downstream semantics or linear accessibility. We evaluate four
complementary diagnostics: \textbf{Future-frame L1}, \textbf{Causal L1},
\textbf{Mask-ratio slope}, and \textbf{Motion cosine}. Together, they probe
future-latent fidelity, intervention sensitivity, robustness under partial
observability, and accessible motion information, distinguishing improved
forecasting from semantic alignment or easier linear readout alone.

\subsection{\textbf{\emph{Evaluator Independence and Audit Split}}}
\label{sec:evaluator_independence}

To prevent agreement with the pseudo-label generator from being mistaken for
improved world modeling, we strictly separate training-target construction from
evaluation. DINOv2-derived targets are constructed only on the training split:
\begin{equation*}
\mathcal T_{\mathrm{train}}
=
\Psi_{\mathrm{struct}}
\left(
\mathcal F_{\mathrm{DINOv2}}(x)
\right),
\qquad
x\in\mathcal D_{\mathrm{train}}.
\end{equation*}

Primary factor-specific evaluation uses a city-disjoint audit split
$\mathcal D_{\mathrm{audit}}$, whose agent masks, visibility states,
interaction pairs, and intervention regions are independently annotated:
\begin{equation*}
\mathcal D_{\mathrm{train}}
\cap
\mathcal D_{\mathrm{audit}}
=
\varnothing,
\qquad
\mathcal T_{\mathrm{audit}}
=
\Psi_{\mathrm{human}}(x).
\end{equation*}

Audit annotations are excluded from training, model selection, threshold
selection, and hyperparameter tuning. City disjointness further limits
scene-specific or geographic memorization. Future-frame L1 and Mask-ratio slope
use the frozen V-JEPA target encoder; Motion cosine uses an independently
frozen motion estimator; and RGB Agent F1 uses an external detector supplying
neither FactorJEPA supervision nor DINOv2 preprocessing. Thus, no headline
evaluator reuses FactorJEPA's pseudo-label generator or training targets.

\subsection{\textbf{\emph{Protocol and Comparisons}}}

We evaluate V-JEPA~2.1 at two scales: ViT-G with approximately 2B parameters
and ViT-g with approximately 1B parameters. Both are evaluated under a matched
stratified 10k-clip regime; ViT-g is additionally trained on the full
115k-clip corpus. Full-scale ViT-G training is omitted under the declared
compute budget because it requires approximately twice the 1B cost.

Comparisons span the frozen backbone, full and parameter-efficient fine-tuning,
continual SSL, LP-FT, LoRA, DoRA, Auto-RGN, WiSE-FT,
\textbf{FactorJEPA-RAW}, and full \textbf{FactorJEPA}. FactorJEPA-RAW retains
the factorized predictor but removes factor-target supervision, yielding
\begin{equation*}
\resizebox{0.98\columnwidth}{!}{$\displaystyle
\underbrace{\{\text{LoRA},\text{DoRA},\text{Auto-RGN}\}}_{\text{generic adaptation}}
\;\longrightarrow\;
\underbrace{\text{FactorJEPA-RAW}}_{\text{factorized architecture}}
\;\longrightarrow\;
\underbrace{\text{FactorJEPA}}_{\text{architecture + factor curriculum}}
$}
\end{equation*}
This attribution chain separates gains from generic adaptation, predictor
factorization, and structured supervision. The experiments therefore test
whether explicit \textbf{layout--agent--interaction structure} improves future
prediction beyond a matched monolithic JEPA predictor.

Within each regime, all methods use identical clips, masking policies,
optimization steps, evaluation splits, and metric implementations. We report
per-encoder test results with paired 95\% BCa bootstrap confidence intervals.

\subsection{\textbf{\emph{Four Diagnostics of Predictive Structure}}}

\paragraph{\textbf{\emph{Future-frame L1 ($\downarrow$).}}}
Normalized L1 distance between predicted and target future embeddings; lower
values indicate \textbf{\emph{more accurate future-latent prediction}}.

\paragraph{\textbf{\emph{Causal L1 ($\downarrow$).}}}
For each independently annotated intervention region $a$, we apply the same
controlled edit $\mathcal I_a$ and compare its effect on predicted and target
future latents:
\begin{equation*}
\resizebox{0.98\columnwidth}{!}{$\displaystyle
\widehat\Delta_a
=
\widehat Y\!\left(\mathcal I_a(x)\right)-\widehat Y(x),
\qquad
\Delta_a^\star
=
\operatorname{sg}\!\left[
\bar f_{\bar\theta}\!\left(\mathcal I_a(x)\right)
-
\bar f_{\bar\theta}(x)
\right].
$}
\end{equation*}
\begin{equation*}
\operatorname{CausalL1}
=
\frac{1}{|\mathcal A|}
\sum_{a\in\mathcal A}
\frac{
\|\widehat\Delta_a-\Delta_a^\star\|_1
}{
\epsilon+\|\Delta_a^\star\|_1
}.
\end{equation*}
Intervention regions and types come from independent audit annotations, not
DINOv2 predictions. The metric evaluates consistency with
intervention-induced changes, not causal identification.

\paragraph{\textbf{\emph{Mask-ratio slope ($\downarrow$).}}}
Increase in prediction error as visual evidence is progressively masked; a
smaller slope indicates \textbf{\emph{greater robustness under partial
observability}}.

\paragraph{\textbf{\emph{Motion cosine ($\uparrow$).}}}
Cosine alignment between linearly decoded and target motion descriptors; higher
values indicate \textbf{\emph{more linearly accessible motion information}}.

Together, these diagnostics distinguish improved future modeling from gains
limited to semantic decoding, a single masking level, or convenient linear
representation.

\subsection{\textbf{\emph{Performance Across Scale and Data Regimes}}}

Figures~\ref{fig:frozen_forest} and~\ref{fig:eval_scorecard} summarize the
primary evidence. Under the matched stratified 10k protocol, FactorJEPA
separates from the strongest competitor on Future-frame L1
($6.3\times$/$4.8\times$ CI widths) and Causal L1
($2.3\times$/$2.7\times$) at 2B/1B. Mask-ratio slope remains within the
confidence interval at 2B ($0.9\times$) but separates at 1B
($1.9\times$). Motion cosine instead favors generic fine-tuning
($-14.5\times$/$-9.2\times$), exposing a limited-data
prediction--motion trade-off.

Full 115k-clip training changes this profile decisively. At 1B, FactorJEPA
separates on all four primary diagnostics, reaching
\textbf{\emph{$43.3\times$}} for Mask-ratio slope,
\textbf{\emph{$33.2\times$}} for Future-frame L1,
\textbf{\emph{$20.0\times$}} for Motion cosine, and
\textbf{\emph{$13.9\times$}} for Causal L1
(Figure~\ref{fig:frozen_forest}). These values denote separation in paired
confidence-interval units, not multiplicative performance gains.

Figure~\ref{fig:eval_scorecard} further shows that the gains are
\textbf{\emph{structured rather than universal}}. FactorJEPA leads on
predictive fidelity, intervention sensitivity, and robustness-oriented
diagnostics, while generic adaptation remains competitive on several
frame-timing and temporal-coherence measures. The improvement therefore targets
the information required to forecast crowded, partially observed scenes rather
than uniformly shifting unrelated metrics.

\subsection{\textbf{\emph{Prediction--Motion Trade-off}}}

Under stratified 10k training, FactorJEPA improves future prediction,
intervention consistency, and masking robustness while conceding maximally
linear motion readout to conventional fine-tuning. This suggests that, with
limited data, the factorized objective prioritizes motion information useful
for structured forecasting rather than motion that is easiest to recover
linearly.

Crucially, full 115k training reverses this deficit: Motion cosine becomes a
strongly separated win while the predictive gains increase further. The
trade-off is therefore \textbf{\emph{data- and optimization-dependent}}, not an
intrinsic limitation of layout--agent--interaction factorization. With
sufficient interaction coverage, FactorJEPA preserves both structured
prediction and accessible motion.

\subsection{\textbf{\emph{Cross-Scale Replication}}}

We test whether adaptation rankings survive the transition from the 2B backbone
to the approximately half-cost 1B backbone. For diagnostic $k$,
\begin{equation*}
\rho_k
=
\operatorname{Spearman}
\left(
\mathbf s_k^{2\mathrm B},
\mathbf s_k^{1\mathrm B}
\right),
\end{equation*}
where $\mathbf s_k^{2\mathrm B}$ and $\mathbf s_k^{1\mathrm B}$ contain
per-method scores at the two scales.

As Figure~\ref{fig:scale_replication} illustrates, Causal L1 yields
$\rho=0.979$, Motion cosine $\rho=0.952$, Future-frame L1
$\rho=0.938$, and Mask-ratio slope $\rho=0.895$. These are the four strongest
cross-scale correlations. Twelve of fifteen diagnostics retain broadly
consistent rankings, while Temporal-order, Teacher-free gap, and Rollout drift
do not transfer reliably.

The result shows that the relative behavior of adaptation strategies is largely
backbone-invariant. Both FactorJEPA's predictive gains and its limited-data
motion trade-off reflect stable method ordering rather than artifacts of one
model size. The high correlations also establish the 1B backbone as a faithful
lower-cost proxy for screening adaptation choices before full-scale 2B
training.

\section{\textbf{Conclusion}}

We introduced \textbf{DENSEWORLD}, a 1,000-hour benchmark from 22 cities for
world modeling under dense traffic, heterogeneity, occlusion, and partial
observability. \textbf{\emph{FactorJEPA}} decomposes future prediction into
visibility-aware \textbf{\emph{layout}}, \textbf{\emph{agent}}, and
\textbf{\emph{interaction}} channels, yielding
\textbf{\emph{(i) lower future-latent error}},
\textbf{\emph{(ii) stronger intervention sensitivity}}, and
\textbf{\emph{(iii) greater robustness to missing evidence}}
across 2B and 1B backbones, with stable cross-scale rankings
($\rho=0.895$--$0.978$). The results expose a consistent trade-off with
linearly accessible motion. A Cosmos-initialized decoder further renders
predicted futures, with an oracle control isolating forecasting error from
decoder limitations. FactorJEPA therefore moves JEPA world models from
monolithic latent prediction toward structured, interpretable forecasting of
complex urban dynamics.

% References directly after the conclusion (matches 0_main_AAAI.tex); the appendix follows.
\bibliography{aaai2027}

% ===== appendix material AFTER the references (natbib collects all \cite, so appendix refs still resolve) =====
\clearpage
% =====================================================================
% Main-paper figures/tables reproduced FULL WIDTH (figure*/table*) in the
% two-column supplement, in v1_main order, each with a short connective note.
% The teaser is single-column so it can sit at the TOP of page 1 (a full-width
% figure* cannot: the title holds the page-1 top). The dense Limitations +
% Appendix text that follows fills the columns around the figure* floats, so
% the notes stay concise without leaving blank pages.
% Order: demo_cards -> scene_types -> factor_visual -> factor_targets ->
%        frozen_scorecard(table) -> forest -> eval_scorecard -> scale_replication
% =====================================================================

% teaser (demo_cards) moved to page 1 (after \maketitle in arxiv_combined.tex)
\section{Main-Paper Figures and Tables, Enlarged}
\label{sec:mainpaper_fullsize}

For legibility, this section reproduces each main-paper figure and the ablation table at full width, in the order they appear in the main text, with a brief note linking them.

% scene-type coverage figure moved into the MAIN body (2_data.tex) per request

\begin{figure*}[t!]
    \centering
    \includegraphics[width=0.88\textwidth]{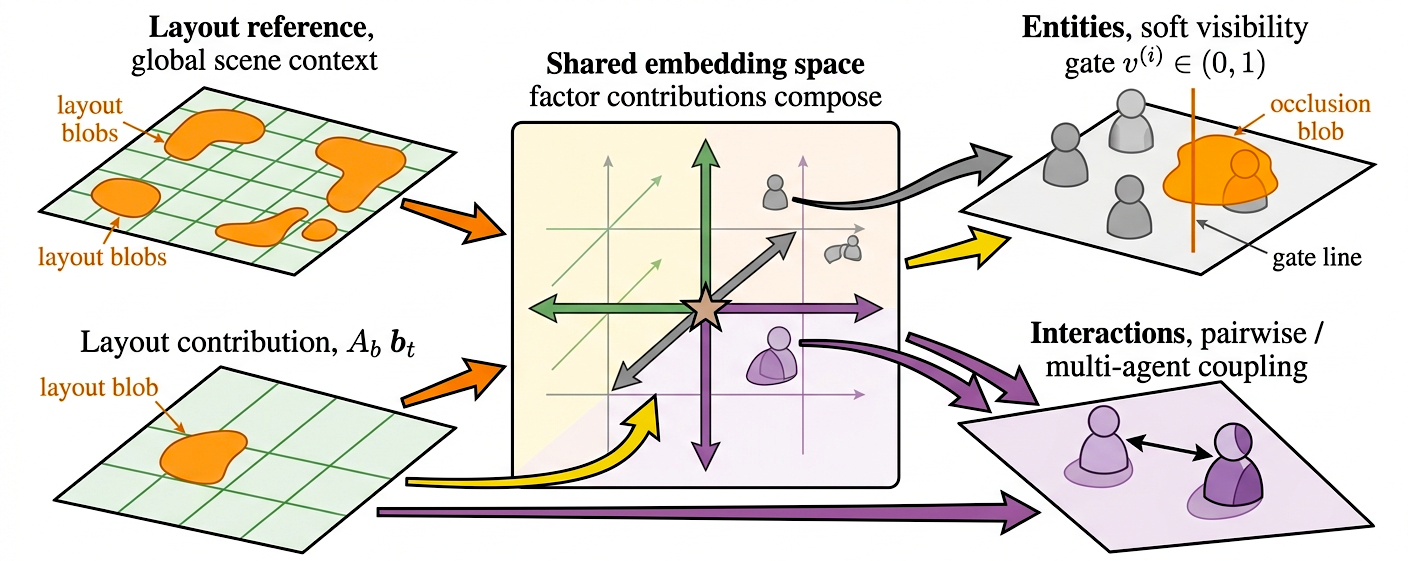}
    \caption{\textbf{\emph{Factorized predictive channels.}} Context is split into \emph{layout}, visibility-gated \emph{agent}, and sparse \emph{interaction} coordinates, then recomposed in the future JEPA embedding, limiting cross-factor leakage.}
    \label{fig:factorjepa_visual_full}
\end{figure*}

\noindent FactorJEPA meets this by decomposing the future into three predictive channels, layout, visibility-gated agents, and sparse interactions, recomposed in the JEPA embedding to discourage cross-factor shortcuts.

\begin{figure*}[t!]
    \centering
    \includegraphics[width=0.88\textwidth]{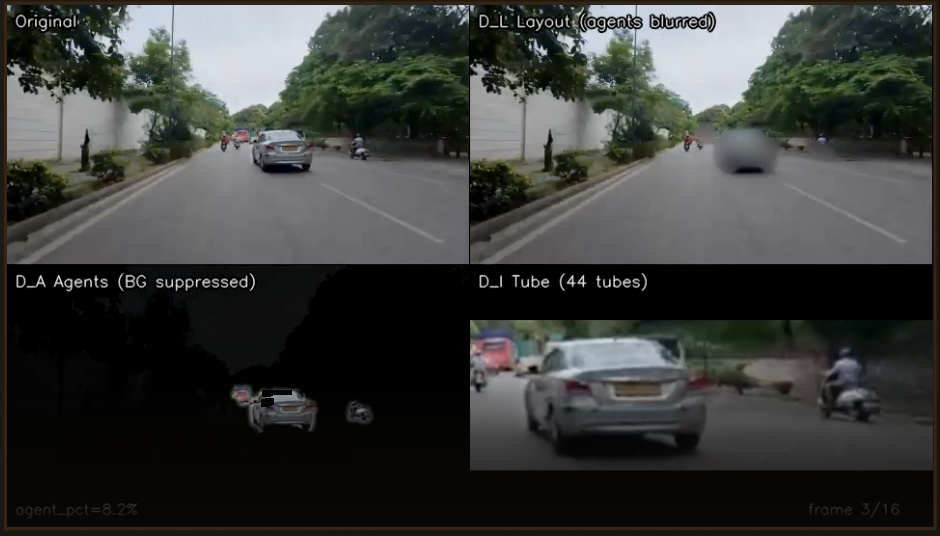}
    \caption{\textbf{\emph{Automatic factor-target construction.}} Per clip, the pipeline derives \emph{layout} $D_L$, \emph{agent} $D_A$, and \emph{interaction} $D_I$ supervision by masking foreground/background and linking agent tubes.}
    \label{fig:dino_factor_targets_full}
\end{figure*}

\noindent These three views are constructed automatically by a fixed detection-and-segmentation pipeline, so the factor supervision is reproducible and scales to the full corpus without manual labels.

% frozen-encoder scorecard table moved into the MAIN body (2_factor_jepa.tex) per request

% forest / eval-scorecard / scale figures REMOVED here: they are EXACT duplicates of
% main-body Fig 7 (forest), Fig 8 (scorecard), Fig 9 (scale). factor_visual and
% segmented_scene above are kept because they are NOT in the main body.
   % enlarged reproductions (forest/eval/scale removed as main-body duplicates)
\section{\textbf{Limitations}}

FactorJEPA demonstrates that structured predictive channels can improve world modeling in dense, heterogeneous, and partially observed urban scenes. The present evidence nevertheless leaves several important questions open, particularly around interaction grounding, semantic factorization, long-horizon evaluation, geographic transfer, and full-scale training.

\paragraph{\textbf{\emph{(i) Interaction grounding remains the principal open challenge.}}} Our interaction targets combine frozen DINOv2 regions, temporal association, relative motion, visibility, proximity, and reliability weighting. This design provides scalable supervision across 1,000 hours of video and captures a tractable subset of the interaction cues that matter for prediction. However, urban interactions are not directly segmentable visual objects: they are temporally extended relations shaped by anticipation, yielding, hesitation, shared right of way, informal signaling, and partially observed actors. Nearby agents may move independently, whereas distant or occluded agents may still influence the future. The current pathway therefore emphasizes sparse pairwise coupling and does not yet fully represent \textbf{\emph{(i) group-level motion}}, \textbf{\emph{(ii) multi-agent negotiation}}, or \textbf{\emph{(iii) temporally persistent interaction events}}. The independently annotated audit split ensures that the training-time interaction generator is not reused as the headline evaluator, while reliability weighting reduces the influence of uncertain targets. Even so, richer event-level annotation, trajectory-aware grounding, and higher-order relational representations could provide more complete supervision. Accordingly, FactorJEPA should be interpreted as learning \textbf{\emph{interaction-sensitive predictive channels}}, while richer semantic, intentional, and causal interaction modeling remains open.

\paragraph{\textbf{\emph{(ii) Factor semantics are anchored at the channel level.}}} Layout, agent, visibility, and interaction targets are produced by a fixed DINOv2-based pipeline rather than exhaustive human annotation. This enables supervision at dataset scale, but the resulting targets may still reflect missed small agents, fragmented tracks, uncertain boundaries, or reduced reliability under severe occlusion, blur, poor illumination, and camera motion. Some distinctions are also inherently contextual: parked vehicles, temporary barriers, vendors, crowds, and encroachments may behave as persistent layout in one scene and dynamic entities in another. Factor-specific pathways, heads, and separation losses provide \textbf{\emph{semantic anchoring and functional channel separation}}, but do not imply coordinate-level identifiability, complete statistical independence, or a unique decomposition of the future latent. Alternative teachers, taxonomies, or factor ranks may yield different yet comparably predictive partitions. Multi-teacher agreement, teacher-free factor discovery, and intervention-based equivalence tests are therefore natural extensions.

\paragraph{\textbf{\emph{(iii) The evaluation targets predictive structure rather than complete causal understanding.}}} The four primary diagnostics have deliberately bounded interpretations. Future-frame L1 measures fidelity in the frozen target-encoder space; Mask-ratio slope measures robustness under controlled evidence removal; Motion cosine measures linearly accessible motion rather than all motion information encoded by the model; and Causal L1 evaluates consistency between predicted and target-encoder changes under independently specified interventions. Causal L1 therefore measures \textbf{\emph{intervention sensitivity}}, not causal identification or unrestricted counterfactual reasoning. The experiments also emphasize short-horizon prediction from prerecorded video. Longer rollouts may accumulate identity, layout, and interaction errors, while action-conditioned forecasting, closed-loop planning, active perception, and online adaptation remain untested. The current results thus establish FactorJEPA as a \textbf{\emph{structured predictive representation}}; extending it to long-horizon and closed-loop world modeling remains future work.

\paragraph{\textbf{\emph{(iv) Geographic breadth does not imply universal coverage.}}} DENSEWORLD spans 22 Indian cities, three capture modes, and substantial variation in density, infrastructure, illumination, weather, road structure, and traffic composition. It should nevertheless be viewed as \textbf{\emph{one large-scale realization}} of populous, crowded, and chaotic Global South urban environments rather than an exhaustive characterization of the Global South. Mobility systems, infrastructure, road conventions, climate, and social negotiation differ across South Asia, Africa, Latin America, Southeast Asia, and the Middle East. The corpus also emphasizes outdoor urban public spaces; rural roads, indoor crowds, industrial environments, disaster settings, and other interaction regimes remain outside its present scope. Cross-region transfer, geographically held-out evaluation, and broader collection are needed to test how far the learned factorization generalizes.

\paragraph{\textbf{\emph{(v) Full-scale training and visual decoding remain incomplete.}}} Full 115k-clip training is reported for the 1B backbone, whereas the 2B model is evaluated under the matched stratified 10k regime. The strong cross-scale rank correlations indicate stable relative method behavior, but they do not replace a full-data 2B experiment. It therefore remains open whether larger-scale training further strengthens the predictive channels, changes their depth-wise allocation, or modifies the observed prediction--motion trade-off. The Cosmos-initialized latent-to-RGB decoder also introduces a separate rendering stage. Oracle controls isolate forecast-induced error from decoder reconstruction error, but rendering may still exhibit blur, deterministic averaging, missed fine-grained agents, or uncertainty collapsed into a single future. Visual plausibility should therefore be interpreted alongside latent-space and oracle-controlled evaluation rather than as standalone evidence of predictive correctness.

These limitations motivate three immediate directions: \textbf{\emph{(i) richer interaction grounding and event-level supervision}}, \textbf{\emph{(ii) higher-order and temporally persistent relational modeling}}, and \textbf{\emph{(iii) broader geographic, long-horizon, and full-scale validation}}.      % Limitations
\appendix

\section*{Appendix}

This appendix provides the complete evidence and implementation record
underlying the main paper. Its organization follows the scientific
progression of the study. Appendices~A--C establish the provenance,
distinctiveness, and reproducibility of DENSEWORLD, its factor targets,
and the executed training protocols. Appendices~D--F then examine the
central methodological claims: which components produce the observed
gains, whether the evaluation is statistically reliable, and whether
the learned layout, agent, and interaction channels are functionally
distinct. Appendix~G evaluates the temporal and downstream consequences
of the prediction--motion trade-off. Finally, Appendix~H documents the
latent-to-RGB decoder, separates forecast error from reconstruction
limitations, and presents quantitative and qualitative future
predictions.

The appendix distinguishes primary confirmatory evidence from
diagnostic and robustness analyses. Unless otherwise stated,
comparisons use identical evaluation clips, paired resampling units,
and the same independently defined audit signals. Factor separation is
interpreted operationally at the level of predictive channels rather
than as coordinate-level identifiability or complete statistical
independence. Intervention-based evaluation measures consistency with
controlled edits, while cross-scale replication refers specifically to
stability between the 1B and 2B V-JEPA~2.1 backbones.

\paragraph{Appendix index.}

\begin{itemize}

% ---------------------------------------------------------------------
\item \textbf{Appendix A: DENSEWORLD Construction, Splits, and Responsible Use}
\begin{itemize}

    \item \textbf{A.1 Acquisition protocol and capture modes}
    
    Shared collection specification; geographic and environmental
    sampling; non-scripted scene activity; and drive-through,
    walk-through, and aerial capture.

    \item \textbf{A.2 Source processing, clip construction, and provenance}
    
    FFmpeg decoding; adaptive shot-boundary detection; model-ready clip
    construction; quality filtering; and the retained
    city--session--source--shot--clip provenance hierarchy.

    \item \textbf{A.3 Partitioning and DINOv2 target provenance}
    
    Source-grouped \(80/10/10\) train--validation--test allocation;
    partition independence; permitted information flow; and
    DINOv2-derived training, validation, and teacher-relative test
    targets. No manually annotated semantic labels are used.

    \item \textbf{A.4 Corpus composition and coverage}
    
    City and capture-mode distributions; environmental and temporal
    coverage; scene families; traffic composition; and the automated
    dynamic-agent taxonomy, including locally distinctive agents.

    \item \textbf{A.5 Privacy processing, responsible use, and research artifacts}
    
    Automated face and registration-plate detection; temporal
    association and blurring; redaction quality control; governed
    research access; supporting code and aggregate statistics;
    intended and prohibited uses; and empirical scope and extensions.

\end{itemize}

% ---------------------------------------------------------------------
\item \textbf{Appendix B: Five-Axis Validation of the DENSEWORLD Regime}
\begin{itemize}

    \item \textbf{B.1 Shared measurement protocol and automated taxonomy}
    
    Common preprocessing, temporal sampling, valid-image regions,
    DINOv2-based extraction, dynamic-agent taxonomy, and source-level
    aggregation across DENSEWORLD, BDD100K, and nuScenes.

    \item \textbf{B.2 Definition of the five regime axes}
    
    Formal definitions of agent count density, agent occupancy,
    occlusion pressure, interaction pressure, and agent heterogeneity,
    including normalization, aggregation units, and edge-case handling.

    \item \textbf{B.3 Matched cross-dataset comparison}
    
    Road-level scene matching by capture configuration, scene type,
    illumination, weather, image resolution, field of view, and clip
    duration. Walk-through and aerial DENSEWORLD clips are analyzed
    separately from the primary driving-benchmark comparison.

    \item \textbf{B.4 Comparative results, uncertainty, and sensitivity}
    
    Five-axis estimates, scene-stratified effects,
    source-video-clustered confidence intervals, and hierarchical
    resampling over cities or sequences. Sensitivity is evaluated over
    detection threshold, minimum agent size, valid-image normalization,
    and interaction parameters. Results are interpreted as evidence for
    an Indian urban realization of the broader DENSEWORLD regime, not
    as an exhaustive characterization of Global-South mobility.

\end{itemize}

% ---------------------------------------------------------------------
\item \textbf{Appendix C: Segmentation and Training}

\begin{itemize}

    \item \textbf{C.1 Frozen Teacher and Factor-Target Construction}
    
    DINOv2 checkpoint, preprocessing, feature extraction, region and
    mask construction, confidence filtering, temporal association, and
    the exact definitions of the layout, agent, visibility, and
    interaction targets \(T_L,T_A,T_V,T_I\). This subsection also
    specifies target dimensions, interaction-pair construction,
    reliability weights, exclusion rules, and retained target coverage.

    \item \textbf{C.2 FactorJEPA Architecture and Prediction Protocol}
    
    V-JEPA~2.1 backbone scales, frozen and trainable components,
    predictor depth, factor projections, subspace ranks, visibility
    gating, masking policy, context duration, prediction horizon, and
    factor-to-predictor information flow.

    \item \textbf{C.3 Optimization, Model Selection, and Executed Algorithm}
    
    Objective terms, loss weights, optimizer, learning-rate schedule,
    momentum-target update, batch size, gradient clipping, training
    steps, random seeds, validation criteria, checkpoint selection, and
    end-to-end training pseudocode.

    \item \textbf{C.4 Baseline Matching and Attribution Controls}
    
    Full fine-tuning, LoRA, DoRA, Auto-RGN, FactorJEPA-RAW, and
    FactorJEPA configurations; trainable-parameter matching; shared
    data, masks, objectives, training steps, and evaluation protocol;
    and the controls separating generic adaptation, factorized
    architecture, and factor-target supervision.

    \item \textbf{C.5 Evaluation Provenance and Resource Accounting}
    
    Separation of training targets, validation signals, and frozen test
    evaluators; software and hardware configuration; trainable and total
    parameters; GPU-hours; peak memory; throughput; inference latency;
    and the reproducibility record for every reported experiment.

\end{itemize}

% ---------------------------------------------------------------------
\item \textbf{Appendix D: Attribution and Component Ablations}
\begin{itemize}

    \item \textbf{D.1 Attribution design and experimental contract}

    Defines the architectural and supervisory interventions, the
    parameter-matching protocol, the executed objectives, and the
    statistical estimands used throughout the appendix.

    \item \textbf{D.2 Architecture \(\times\) structured-supervision
    factorial}

    Compares matched monolithic and factorized predictors with and
    without DINOv2-derived structured supervision, isolating
    architecture, supervision, and their interaction.

    \item \textbf{D.3 Factor pathways and objective ablations}

    Tests the layout--agent target loss, visibility pathway,
    interaction pathway, sparsity regularization, and separation
    objective through controlled one-factor removals.

    \item \textbf{D.4 Falsification controls and sensitivity}

    Evaluates temporally misaligned and clip-shuffled teacher targets,
    reliability weighting, factor ranks, and loss coefficients to
    distinguish structured information from generic regularization.

    \item \textbf{D.5 Attribution synthesis}

    Consolidates the factorial contrasts and component ablations,
    reporting which mechanisms account for each prediction,
    visibility, interaction, motion, and RGB result.

\end{itemize}

% ---------------------------------------------------------------------
\item \textbf{Appendix E: Metrics, Clustered Inference, and Robustness}
\begin{itemize}

    \item \textbf{E.1 Evaluation contract and metric registry}

    Defines every reported diagnostic by formula, direction, evaluation
    unit, aggregation rule, frozen evaluator, provenance, and
    primary or exploratory status.

    \item \textbf{E.2 Primary predictive diagnostics}

    Specifies Future-frame MSE, intervention-consistency L1,
    Mask-ratio slope, and Motion cosine, including token aggregation,
    normalization, probe training, and horizon or masking support.

    \item \textbf{E.3 Secondary semantic and temporal diagnostics}

    Defines action, taxonomy, rollout, temporal-order, arrow-of-time,
    playback-pace, temporal-correspondence, and exposure-bias
    diagnostics under a common aggregation protocol.

    \item \textbf{E.4 Paired clustered inference and multiplicity}

    Defines the city--source-video estimand, hierarchical paired
    bootstrap, treatment of training seeds, mixed-effects sensitivity,
    and multiple-comparison control.

    \item \textbf{E.5 Metric sensitivity and complete numerical results}

    Reports horizon and masking curves, intervention-normalization
    sensitivity, cross-scale ranking robustness, leave-one-method and
    leave-one-family analyses, and the complete ViT-G and ViT-g
    numerical scorecards.

\end{itemize}

% ---------------------------------------------------------------------
\item \textbf{Appendix H: Latent-to-RGB Decoding and Qualitative Results}
\begin{itemize}
    \item \textbf{H.1 Cosmos checkpoint and decoder components}
    
    Exact initialization, frozen modules, trainable modules, and
    adaptation configuration.

    \item \textbf{H.2 Latent transport architecture}
    
    Token projection, spatial reshaping, positional encoding,
    normalization, and transport dimensions.

    \item \textbf{H.3 Decoder training objective}
    
    Pixel, perceptual, structural, motion, and regularization losses.

    \item \textbf{H.4 Decoder-neutrality controls}
    
    Target-latent-only, balanced-prediction, and method-specific
    decoder protocols.

    \item \textbf{H.5 Oracle, forecast-gap, and end-to-end evaluation}
    
    Operational decomposition of reconstruction and predicted-latent
    errors.

    \item \textbf{H.6 Quantitative RGB results}
    
    PSNR, SSIM, LPIPS, Flow EPE, Agent F1, dynamic-region metrics, and
    horizon-wise results.

    \item \textbf{H.7 Density- and occlusion-stratified RGB evaluation}
    
    Rendering quality across low-density, high-density, and heavily
    occluded conditions.

    \item \textbf{H.8 Factor-sensitive visual diagnostics}
    
    Layout-, agent-, and interaction-removed decoding and influence
    maps.

    \item \textbf{H.9 Qualitative future-prediction gallery}
    
    Context, target, oracle, V-JEPA, FactorJEPA-RAW, and FactorJEPA
    comparisons.

    \item \textbf{H.10 Failure taxonomy}
    
    Layout drift, missed agents, identity errors, incorrect
    interactions, motion errors, deterministic averaging, and decoder
    hallucination.

    \item \textbf{H.11 Additional qualitative examples}
    
    Examples across cities, scene types, viewpoints, weather,
    illumination, density, and prediction horizons.
\end{itemize}

\end{itemize}

% page break removed for single-column continuous flow

\section{DENSEWORLD Construction, Splits, and Responsible Use}
\label{app:dataset}

DENSEWORLD comprises approximately \textbf{1,000 hours} of
drive-through, walk-through, and aerial video acquired across
\textbf{22 Indian cities}. It targets predictive modeling under high
agent density, heterogeneous traffic, persistent occlusion, soft spatial
boundaries, and frequent local interaction. DENSEWORLD contains
\emph{no manually annotated semantic labels}; all layout, agent,
visibility, and interaction targets are produced automatically using a
fixed DINOv2-based pipeline.

% =====================================================================
\subsection{Acquisition Protocol and Capture Modes}
\label{app:dataset_acquisition}

DENSEWORLD was acquired through two professional video-collection
organizations following a \textbf{shared acquisition specification}.
The specification defined coverage targets over geographic sites,
scene types, times of day, weather, visibility, road structure, and
infrastructure conditions. It controlled \emph{where and how} videos
were acquired without scripting traffic, pedestrian, or animal
behavior.

The collection uses three capture modes. \textbf{Drive-through}
recordings emphasize road-level ego-motion, mixed traffic, and
near-field interactions. \textbf{Walk-through} recordings capture
pedestrian-scale navigation, shared spaces, markets, commercial
corridors, and transit areas. \textbf{Aerial} recordings provide broader
context for crowd flow, junction organization, traffic structure, and
interaction topology. City, collection session, capture mode, and
source timestamps are retained as acquisition metadata.

% =====================================================================
\subsection{Source Processing, Clip Construction, and Provenance}
\label{app:clip_construction}

Source videos are decoded using
FFmpeg\footnote{\url{https://ffmpeg.org/}} while retaining frame
timestamps and source identifiers. Shot boundaries are detected using
the PySceneDetect
\texttt{AdaptiveDetector}\footnote{\url{https://www.scenedetect.com/docs/latest/api/detectors.html}},
which normalizes adjacent-frame changes in hue, saturation, and
luminance by a rolling temporal average. This reduces false boundaries
caused by rapid ego-motion and camera shake.

We use a fixed corpus-wide configuration:
\texttt{adaptive\_threshold}$=3.0$,
\texttt{min\_content\_val}$=15.0$,
\texttt{window\_width}$=2$, and
\texttt{min\_scene\_len}$=15$ frames. For a source video
\(v=\{x^{(v)}_t\}_{t=1}^{T_v}\), the detector returns ordered
boundaries
\begin{equation*}
\mathcal B_v
=
\left\{
1=b^{(v)}_0
<
b^{(v)}_1
<
\cdots
<
b^{(v)}_{J_v}=T_v
\right\},
\end{equation*}
which partition the source into candidate visually continuous shots.
Shots shorter than the minimum usable duration are excluded; retained
shots yield model-ready clips of approximately \(6\)--\(13\) seconds.
No clip crosses a detected shot boundary.

The retained provenance hierarchy is
\begin{equation*}
\resizebox{0.98\columnwidth}{!}{$\displaystyle
\text{city}
\;\longrightarrow\;
\text{collection session}
\;\longrightarrow\;
\text{source video}
\;\longrightarrow\;
\text{shot}
\;\longrightarrow\;
\text{clip}
$}
\end{equation*}
For every clip \(c\), we store
\begin{equation*}
\pi(c)=
\bigl(
i_{\mathrm{city}},
i_{\mathrm{session}},
i_{\mathrm{source}},
i_{\mathrm{shot}},
m_{\mathrm{capture}},
t_{\mathrm{start}},
t_{\mathrm{end}}
\bigr).
\end{equation*}
Thus, the clip is the \textbf{computational input}, while its source,
session, and city identifiers preserve higher-level geographic and
temporal dependence.

% =====================================================================
\subsection{Partitioning and DINOv2 Target Provenance}
\label{app:dataset_splits}

Source videos are assigned as indivisible groups using an
\textbf{\(80/10/10\) train/validation/test allocation}, stratified by
city and capture mode. Split assignment precedes clip extraction, so
every shot and clip derived from one source video inherits the same
partition. Let \(\mathcal V_s\) denote the source-video set assigned to
partition \(s\). The executed grouping satisfies
\begin{equation*}
\mathcal V_{\mathrm{train}}\cap\mathcal V_{\mathrm{val}}
=
\mathcal V_{\mathrm{train}}\cap\mathcal V_{\mathrm{test}}
=
\mathcal V_{\mathrm{val}}\cap\mathcal V_{\mathrm{test}}
=
\varnothing.
\end{equation*}

DENSEWORLD does not use manually annotated factor labels. For each
partition, automated structural targets are generated by the fixed
DINOv2 pipeline:

\begin{equation*}
\begin{aligned}
\mathcal T_s^{\mathrm{DINO}}
&=
\left\{
\Psi_{\mathrm{struct}}
\!\left(
F_{\mathrm{DINOv2}}(x)
\right)
:
x\in\mathcal D_s
\right\},
\\[-1pt]
s
&\in
\left\{
\mathrm{train},
\mathrm{val},
\mathrm{test}
\right\}.
\end{aligned}
\end{equation*}

Training targets may provide factor supervision; validation targets may
support prespecified selection and calibration; and test targets are
used only for held-out \emph{teacher-relative} factor diagnostics.
Test-derived targets, thresholds, and metrics never influence gradient
updates or model selection.

\begin{table*}[t]
\centering
\caption{
\textbf{Composition, information flow, and independence constraints of
the DENSEWORLD partitions.}
Allocation is performed over source-video groups. Durations are
approximate because grouped recordings vary in length and usable-shot
yield. DINOv2-based test measurements are reported as teacher-relative;
headline prediction, motion, and RGB metrics use fixed evaluators that
do not provide FactorJEPA's factor-supervision targets.
}
\label{tab:dataset_partitions}
\scriptsize
\setlength{\tabcolsep}{3pt}
\renewcommand{\arraystretch}{1.14}
\begin{tabular}{
    p{0.085\textwidth}
    p{0.055\textwidth}
    p{0.075\textwidth}
    p{0.065\textwidth}
    p{0.065\textwidth}
    p{0.070\textwidth}
    p{0.225\textwidth}
    p{0.230\textwidth}
}
\toprule
\textbf{Partition}
&
\textbf{Share}
&
\textbf{Duration}
&
\shortstack{\textbf{Gradient}\\\textbf{updates}}
&
\shortstack{\textbf{Model}\\\textbf{selection}}
&
\shortstack{\textbf{Threshold}\\\textbf{calibration}}
&
\textbf{Signal exposed}
&
\textbf{Isolation constraint}
\\
\midrule

Training
&
\(80\%\)
&
\(\approx800\) h
&
Yes
&
No
&
No
&
JEPA prediction targets and DINOv2-derived layout, agent, and
interaction targets.
&
Source-video disjoint from validation and test; all descendants of a
source remain grouped.
\\

Validation
&
\(10\%\)
&
\(\approx100\) h
&
No
&
Yes
&
Yes
&
Prespecified validation metrics and DINOv2-derived targets used only
for selection and calibration.
&
Source-video disjoint from training and test; never used for gradient
optimization.
\\

Test
&
\(10\%\)
&
\(\approx100\) h
&
No
&
No
&
No
&
Frozen headline evaluators and DINOv2-derived targets used only for
explicitly teacher-relative diagnostics.
&
Source-video disjoint from training and validation; all configurations
are frozen before test access.
\\

\midrule

\textbf{Total}
&
\(\mathbf{100\%}\)
&
\(\approx1{,}000\) h
&
-- & -- & -- & -- & --
\\

\bottomrule
\end{tabular}
\end{table*}

% =====================================================================
\subsection{Corpus Composition and Coverage}
\label{app:dataset_composition}

\begin{figure*}[ht!]
\centering
\includegraphics[width=0.8\textwidth]{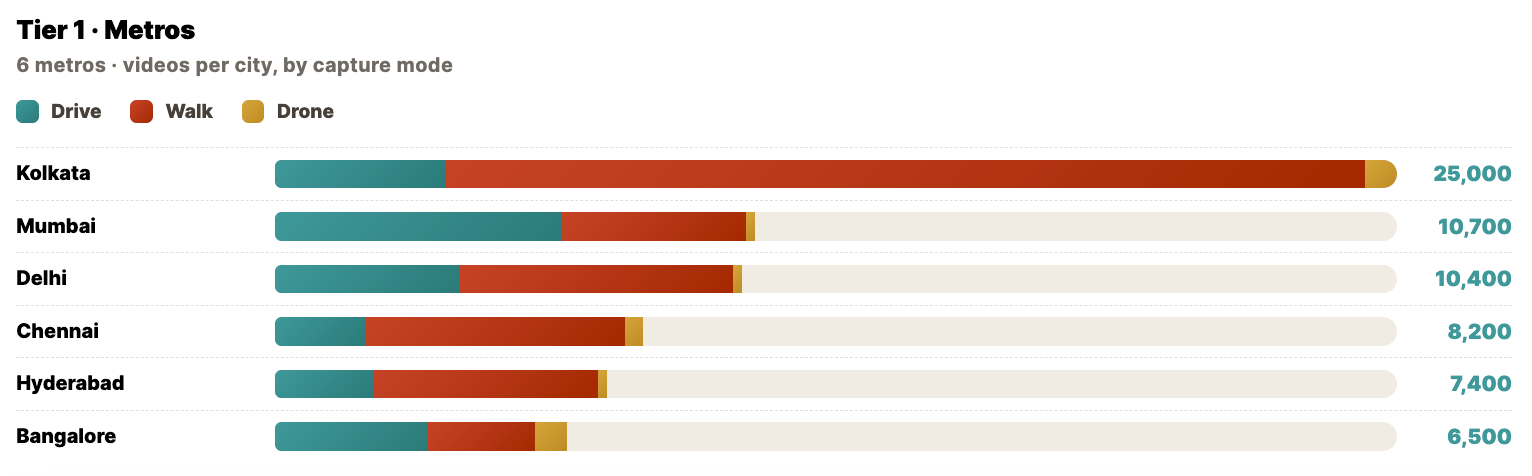}

\vspace{4pt}

\includegraphics[width=0.8\textwidth]{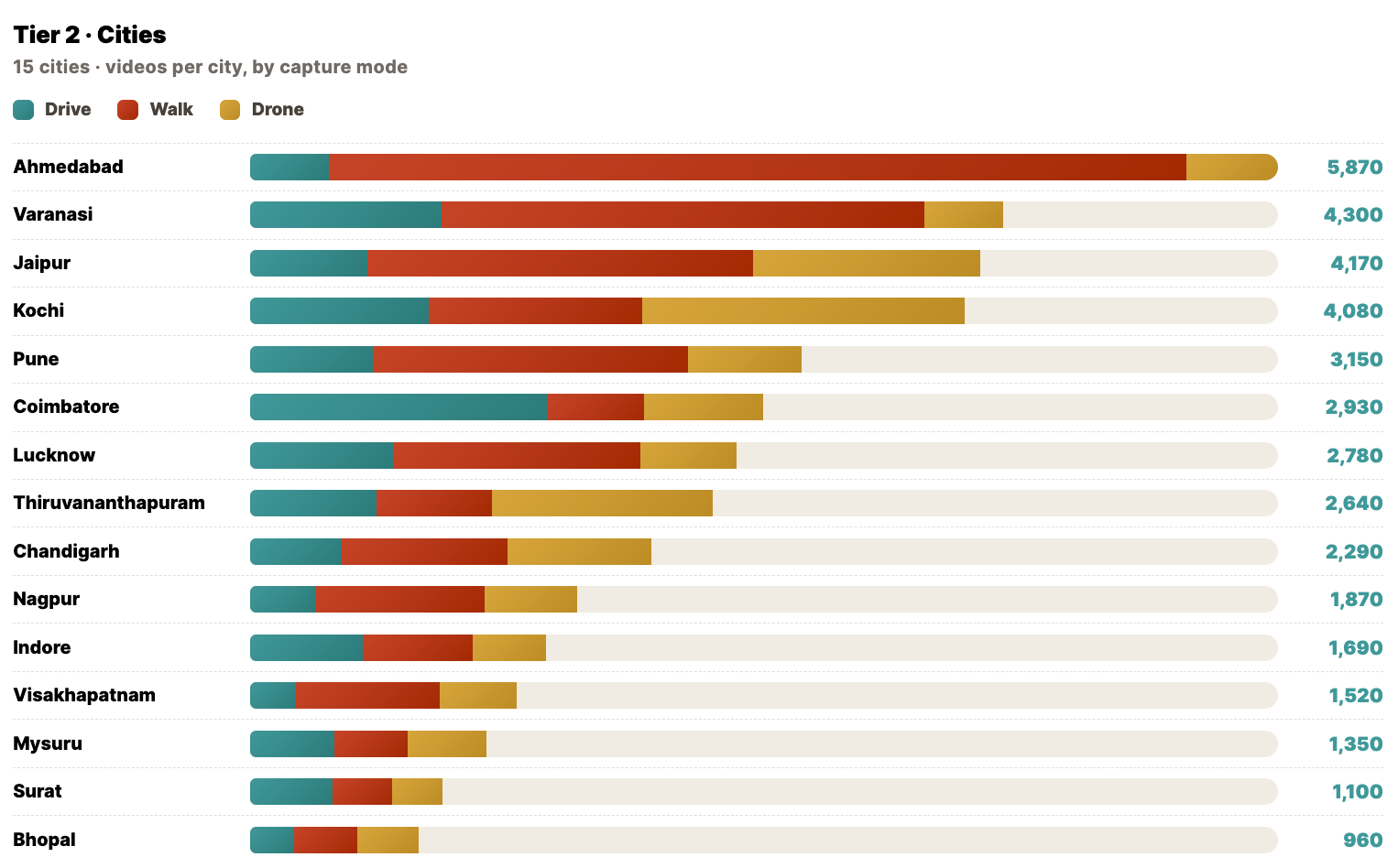}
\caption{
\textbf{City-level composition by capture mode.}
The upper panel shows metropolitan collection sites and the lower panel
shows additional urban sites. Each bar decomposes a city's model-ready
clips into drive-through, walk-through, and aerial capture. The value
at the right gives the city total; both panels use a common horizontal
scale.
}
\label{fig:city_capture_distribution}
\end{figure*}

\paragraph{Geographic and capture-mode composition.}
The collection spans metropolitan and additional urban sites across
India. These groups are \textbf{collection strata}, not claims of an
official administrative taxonomy or within-stratum homogeneity.
Figure~\ref{fig:city_capture_distribution} reports the contribution of
drive-through, walk-through, and aerial clips for each city. The
city-by-mode distribution is observational and intentionally
heterogeneous; bar length describes corpus composition, not evaluation
weight or statistical independence.

\paragraph{Environmental and agent coverage.}
DENSEWORLD includes commercial, residential, transit, heritage,
coastal, and high-density street environments, spanning markets,
junctions, flyovers, highways, promenades, bazaars, ghats, beaches, and
shared roads. It also captures variation in illumination, weather,
traffic composition, crowd density, road geometry, surface quality,
encroachment, and pedestrian--vehicle separation.

\begin{figure*}[t]
\centering
\includegraphics[width=\textwidth]{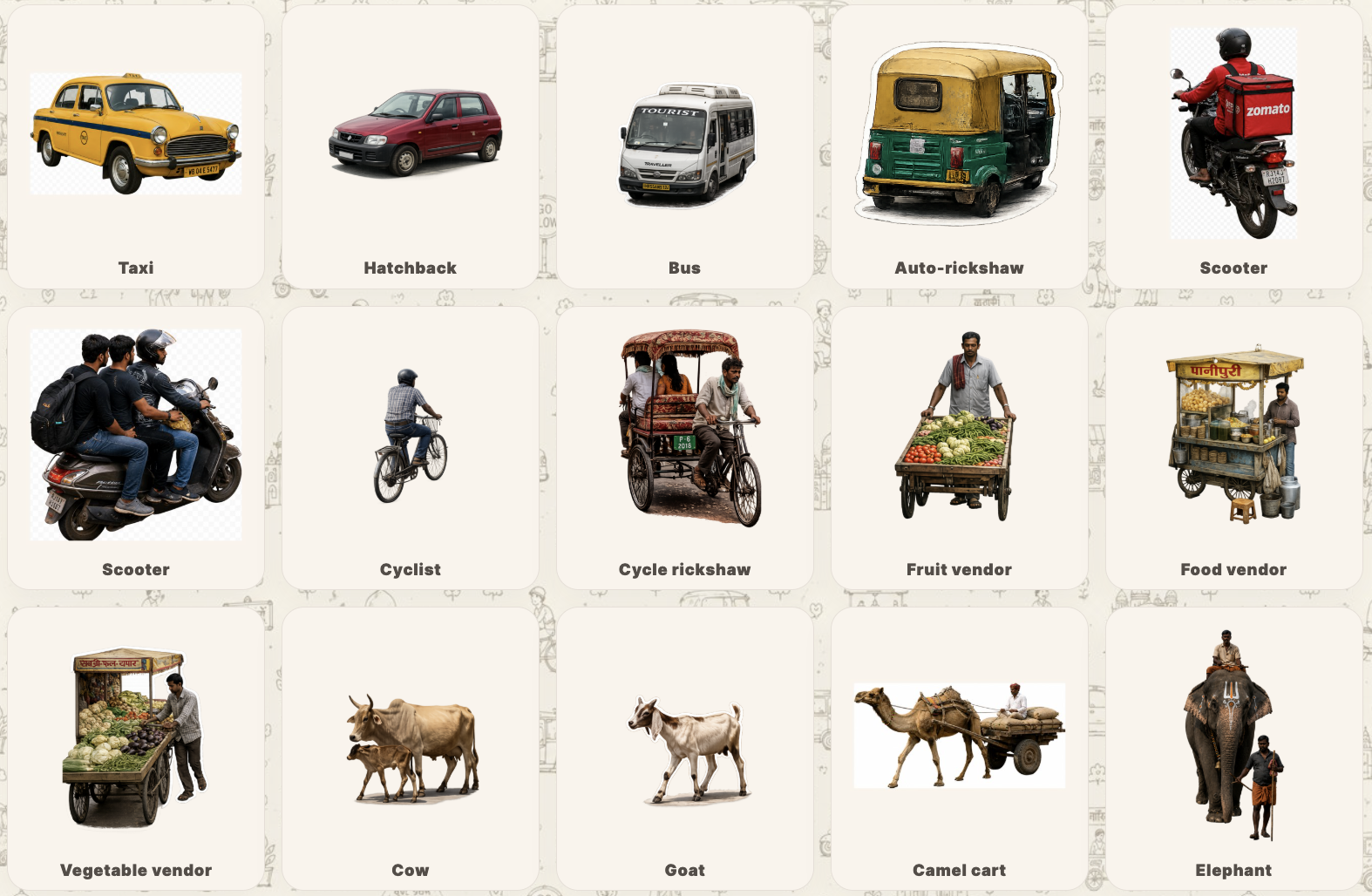}
\caption{
\textbf{Illustrative DENSEWORLD dynamic-agent taxonomy.}
The examples communicate the semantic scope of the automated taxonomy,
including conventional transport, intermediate mobility, mobile
vendors, animal-drawn transport, and independently moving animals.
They are illustrative category renderings rather than
frequency-proportional dataset samples.
}
\label{fig:agent_taxonomy}
\end{figure*}

The dynamic-agent taxonomy extends beyond passenger vehicles and
pedestrians to include auto-rickshaws, cycle rickshaws, delivery riders,
multi-rider two-wheelers, cyclists, mobile vendors, animal-drawn
vehicles, and independently moving animals. Figure~\ref{fig:agent_taxonomy}
illustrates category semantics; it is neither a frequency-proportional
sample nor manually annotated evidence. Quantitative prevalence and
heterogeneity statistics derived from the automated pipeline are
reported as \emph{DINOv2-estimated} quantities.

% =====================================================================
\subsection{Privacy Processing, Responsible Use, and Research Artifacts}
\label{app:dataset_responsible_use}

\paragraph{Privacy processing.}
Before downstream use, clips pass through an automated privacy pipeline.
Faces are localized with
SCRFD\footnote{\url{https://github.com/deepinsight/insightface/tree/master/detection/scrfd}},
and vehicle-registration plates are localized with a custom-trained
YOLO detector\footnote{\url{https://docs.ultralytics.com/tasks/detect/}}.
Detections are associated temporally using
ByteTrack\footnote{\url{https://github.com/FoundationVision/ByteTrack}},
expanded by a fixed spatial safety margin, and redacted using OpenCV
\texttt{GaussianBlur}\footnote{\url{https://docs.opencv.org/4.x/d4/d86/group__imgproc__filter.html}}.

The policy covers frontal, profile, small, and partially occluded faces,
detectable faces in reflections, and readable or partially readable
vehicle-registration plates. Let
\(\mathcal R_t^{\mathrm{face}}\) and
\(\mathcal R_t^{\mathrm{plate}}\) denote the temporally associated and
expanded regions in frame \(x_t\). The processed frame is
\begin{equation*}
\widetilde{x}_t
=
\mathcal B_{\sigma}
\left(
x_t,\,
\mathcal R_t^{\mathrm{face}}
\cup
\mathcal R_t^{\mathrm{plate}}
\right),
\end{equation*}
where \(\mathcal B_{\sigma}\) applies region-scaled Gaussian filtering.
Automated quality control checks track discontinuities, unmatched
detections, frame-boundary truncation, and processing failures.
Segments with incomplete or failed redaction are excluded. Only
\(\widetilde{x}_t\) enters target construction, training, evaluation,
or external illustration.

\paragraph{Responsible use and research artifacts.}
DENSEWORLD is intended for predictive world modeling, video
representation learning, multi-agent forecasting, and robustness under
partial observability. It is not intended for person identification,
persistent individual tracking, sensitive-attribute inference, or
surveillance.

The complete corpus is maintained within the \textbf{governed research
environment} used for the reported experiments. Supporting artifacts
include the FactorJEPA implementation, preprocessing configuration,
DINOv2 target-construction pipeline, metric definitions, automated
taxonomy, aggregate corpus statistics, and representative
privacy-filtered material, subject to applicable rights and privacy
review.

\section{Five-Axis Validation of the DENSEWORLD Regime}
\label{app:benchmark_validation}

We ask whether DENSEWORLD remains quantitatively distinct from
BDD100K~\cite{yu2020bdd100k} and
nuScenes~\cite{caesar2020nuscenes} after controlling for observable
differences in capture and scene composition. We characterize this
regime along five axes: \textbf{agent count density},
\textbf{agent occupancy}, \textbf{visibility pressure},
\textbf{interaction pressure}, and \textbf{agent heterogeneity}.

The comparison is \textbf{outcome-blind}, \textbf{scene-matched}, and
\textbf{dependence-aware}. All datasets are processed using the same
frozen DINOv2-based measurement pipeline; none of the five axes enters
the matching procedure; and uncertainty is clustered above the frame
and clip levels. The resulting quantities are therefore interpreted as
\emph{teacher-relative dataset statistics}, not manually annotated
ground truth or causal effects of dataset membership.

% =====================================================================
\subsection{Shared Measurement Protocol and Automated Taxonomy}
\label{app:shared_taxonomy}

All three datasets are processed through the common sequence

\begin{equation*}
\resizebox{0.98\columnwidth}{!}{$\displaystyle
\text{video}
\;\longrightarrow\;
\text{sampled frames}
\;\longrightarrow\;
\text{DINOv2 features}
\;\longrightarrow\;
\text{structural pseudo-annotations}
\;\longrightarrow\;
\text{five-axis statistics}
$}
\end{equation*}

For frame \(x_t^{(d)}\) from dataset
\(d\in\{\mathrm{DW},\mathrm{BDD},\mathrm{NS}\}\), the frozen DINOv2
encoder~\cite{oquab2024dinov2} and structural postprocessor produce

\begin{equation*}
\begin{aligned}
\mathcal P_t^{(d)}
&=
\Psi_{\mathrm{struct}}
\left(
F_{\mathrm{DINOv2}}(x_t^{(d)})
\right)
\\
&=
\left\{
\left(
\hat c_{i,t},
\hat s_{i,t},
\hat B_{i,t},
\hat M_{i,t},
\hat\ell_{i,t}
\right)
\right\}_{i\in\mathcal I_t^{(d)}} .
\end{aligned}
\end{equation*}

Here, \(\mathcal I_t^{(d)}\) indexes candidate instances;
\(\hat c_{i,t}\), \(\hat s_{i,t}\), \(\hat B_{i,t}\), and
\(\hat M_{i,t}\) denote the predicted class, confidence, bounding box,
and visible instance mask; and \(\hat\ell_{i,t}\) is the temporally
associated track identifier.

DINOv2 itself is not an instance detector. Accordingly,
\(\Psi_{\mathrm{struct}}\) denotes the executed frozen sequence of
proposal extraction, class assignment, mask construction, duplicate
suppression, and temporal association. Its component architectures,
checkpoint identifiers, inference resolution, thresholds, suppression
rule, and association parameters are fixed before comparison and
applied identically to all datasets.

Let \(\Omega_t\) denote the valid, unpadded image region. The retained
agent set is

\begin{equation*}
\mathcal A_t^{(d)}
=
\left\{
i\in\mathcal I_t^{(d)}:
\hat s_{i,t}\ge\tau_{\mathrm{det}},
\quad
\frac{
|\hat M_{i,t}\cap\Omega_t|
}{
|\Omega_t|
}
\ge a_{\min}
\right\}.
\end{equation*}

The same temporal stride, resizing rule, confidence threshold,
minimum-area criterion, and temporal-association configuration are used
for DENSEWORLD, BDD100K, and nuScenes. Predictions that cannot be
mapped unambiguously are excluded rather than reassigned using
dataset-specific heuristics.

Predicted classes are mapped to seven shared superclasses:

\begin{enumerate}[
    label=(\roman*),
    leftmargin=2.8em,
    itemsep=1pt,
    topsep=2pt,
    parsep=0pt
]
    \item \textbf{pedestrian}
    \item \textbf{pedal cycle}
    \item \textbf{powered two-wheeler}
    \item \textbf{light passenger vehicle}
    \item \textbf{heavy or public vehicle}
    \item \textbf{intermediate or non-motorized transport}
    \item \textbf{animal or animal-drawn transport}
\end{enumerate}
Native BDD100K and nuScenes annotations are not used to compute,
calibrate, or select any primary statistic. To expose possible
teacher-induced domain effects, we report the retained-instance rate,
unresolved-class rate, confidence distribution, small-instance
exclusion rate, and track-fragmentation rate separately for each
dataset. The primary analysis is additionally repeated over a
prespecified grid of confidence, area, and association settings.

% =====================================================================
\subsection{Definition of the Five Regime Axes}
\label{app:five_axis_definitions}

Suppressing the dataset superscript for clarity, let

\[
N_t=|\mathcal A_t|,
\qquad
\mathcal T_1=\{t:N_t\ge1\},
\qquad
\mathcal T_2=\{t:N_t\ge2\}.
\]

All boxes and masks are clipped to \(\Omega_t\) before measurement.
Count and occupancy are evaluated over all sampled frames. Visibility
pressure is evaluated over \(\mathcal T_1\), whereas interaction and
heterogeneity are evaluated over \(\mathcal T_2\). We report the
eligible-frame rate for every conditional statistic.

\begin{enumerate}

% ---------------------------------------------------------------------
\item \textbf{Agent count density.}

The primary statistic is the mean number of retained dynamic agents per
sampled frame:

\[
D_{\mathrm{count}}
=
\frac{1}{T}
\sum_{t=1}^{T}N_t.
\]

We additionally report valid-support-normalized count,

\[
D_{\mathrm{count}}^{\mathrm{area}}
=
\frac{1}{T}
\sum_{t=1}^{T}
\frac{N_t}{|\Omega_t|/10^6},
\]

as a sensitivity measure for cropping, padding, and inference
resolution. It is not interpreted as physical agent density.

% ---------------------------------------------------------------------
\item \textbf{Agent occupancy.}

Occupancy is the fraction of valid image support covered by the union
of visible agent masks:

\[
D_{\mathrm{occ}}
=
\frac{1}{T}
\sum_{t=1}^{T}
\frac{
\left|
\displaystyle
\bigcup_{i\in\mathcal A_t}
(\hat M_{i,t}\cap\Omega_t)
\right|
}{
|\Omega_t|
}.
\]

Taking the union prevents crowded or overlapping regions from being
counted multiple times.

% ---------------------------------------------------------------------
\item \textbf{Visibility pressure.}

Because no amodal annotations are available, this axis is defined as an
automated \emph{visibility-pressure proxy}. It combines three
observable cues: substantial pairwise box overlap, valid-image-boundary
truncation, and short interior track disappearance followed by
reappearance.

For boxes \(B_i\) and \(B_j\), define intersection over minimum area as

\[
\operatorname{IoM}(B_i,B_j)
=
\frac{
|B_i\cap B_j|
}{
\min(|B_i|,|B_j|)+\epsilon
}.
\]

The agent-level visibility indicator is

\begin{equation*}
\resizebox{0.98\columnwidth}{!}{$\displaystyle
z_{i,t}
=
\mathbf 1
\left[
\max_{j\ne i}
\operatorname{IoM}
\bigl(\hat B_{i,t},\hat B_{j,t}\bigr)
\ge\tau_{\mathrm{ov}}
\;\lor\;
e_{i,t}=1
\;\lor\;
g_{i,t}=1
\right]
$}
\end{equation*}

where \(e_{i,t}\) indicates boundary truncation and \(g_{i,t}\)
indicates an interior track gap followed by reappearance within the
prespecified temporal window. Visibility pressure is

\[
D_{\mathrm{vis}}
=
\frac{1}{|\mathcal T_1|}
\sum_{t\in\mathcal T_1}
\frac{
\sum_{i\in\mathcal A_t}z_{i,t}
}{
N_t
}.
\]

Overlap, boundary-truncation, and track-gap rates are also examined
separately to verify that the composite is not dominated by one proxy.

% ---------------------------------------------------------------------
\item \textbf{Interaction pressure.}

We construct a frame-level interaction graph
\(\mathcal G_t=(\mathcal A_t,\mathcal E_t)\). Let
\(\mathbf p_{i,t}\) denote the normalized image-plane center of agent
\(i\). Velocities \(\widetilde{\mathbf u}_{i,t}\) are computed after
removing estimated global image motion. Define

\[
\mathbf r_{ij,t}
=
\mathbf p_{j,t}-\mathbf p_{i,t},
\qquad
\mathbf w_{ij,t}
=
\widetilde{\mathbf u}_{j,t}
-
\widetilde{\mathbf u}_{i,t}.
\]

For approaching pairs, the predicted time and distance at closest
approach are

\begin{equation*}
\begin{aligned}
t_{ij,t}^{*}
&=
-\frac{
\mathbf r_{ij,t}^{\top}\mathbf w_{ij,t}
}{
\|\mathbf w_{ij,t}\|_2^2+\epsilon
},
\\
d_{ij,t}^{*}
&=
\left\|
\mathbf r_{ij,t}
+
t_{ij,t}^{*}\mathbf w_{ij,t}
\right\|_2 .
\end{aligned}
\end{equation*}

An interaction edge is present when the pair is currently proximate or
is predicted to approach within both the temporal and spatial
thresholds:

\[
(i,j)\in\mathcal E_t
\iff
\|\mathbf r_{ij,t}\|_2\le\rho
\;\lor\;
\left(
0<t_{ij,t}^{*}\le\tau_t
\;\land\;
d_{ij,t}^{*}\le\tau_d
\right).
\]

Interaction pressure is the conditional mean graph degree:

\[
D_{\mathrm{int}}
=
\frac{1}{|\mathcal T_2|}
\sum_{t\in\mathcal T_2}
\frac{2|\mathcal E_t|}{N_t}.
\]

Per-pair edge density is retained as a sensitivity statistic to
separate interaction structure from the mechanical effect of agent
count.

% ---------------------------------------------------------------------
\item \textbf{Agent heterogeneity.}

Let \(N_{c,t}\) be the number of retained agents assigned to superclass
\(c\in\mathcal C\). The primary statistic is the probability that two
agents drawn without replacement from the same frame belong to
different superclasses:

\[
h_t
=
1-
\sum_{c\in\mathcal C}
\frac{
N_{c,t}(N_{c,t}-1)
}{
N_t(N_t-1)
},
\qquad
t\in\mathcal T_2.
\]

Dataset-level heterogeneity is

\[
D_{\mathrm{het}}
=
\frac{1}{|\mathcal T_2|}
\sum_{t\in\mathcal T_2}h_t.
\]

This pairwise definition lies in \([0,1]\) and is less sensitive to
small per-frame counts than plug-in entropy. Normalized entropy, class
richness, and the effective number of classes are retained as secondary
measures.

\end{enumerate}

\begin{table*}[ht!]
\centering
\caption{
\textbf{Scene-matched agent density and occupancy.}
Count density is measured in agents per sampled frame; occupancy is the
percentage of valid image area covered by the union of visible agent
masks. \(\Delta=\mathrm{DW}-\mathrm{reference}\), and \(R\) is the
ratio of the two means. DENSEWORLD results are aggregated over
\(K\) executed folds. The final row gives the unweighted mean across
the six matched scene strata.
}
\label{tab:scene_stratified_density_occupancy}

\scriptsize
\setlength{\tabcolsep}{3.8pt}
\renewcommand{\arraystretch}{1.20}
\arrayrulecolor{RuleBlue}

\begin{tabular}{
    >{\raggedright\arraybackslash}p{0.155\textwidth}
    >{\centering\arraybackslash}p{0.075\textwidth}
    >{\centering\arraybackslash}p{0.095\textwidth}
    >{\centering\arraybackslash}p{0.065\textwidth}
    >{\centering\arraybackslash}p{0.065\textwidth}
    >{\centering\arraybackslash}p{0.075\textwidth}
    >{\centering\arraybackslash}p{0.095\textwidth}
    >{\centering\arraybackslash}p{0.065\textwidth}
    >{\centering\arraybackslash}p{0.065\textwidth}
}
\toprule

\rowcolor{PanelBlue}
&
\multicolumn{4}{c}{
\textbf{Agent count density}
\quad\emph{(agents/frame)}
}
&
\multicolumn{4}{c}{
\textbf{Agent occupancy}
\quad\emph{(\% valid area)}
}
\\

\cmidrule(lr){2-5}
\cmidrule(lr){6-9}

\rowcolor{PanelBlue}
\textbf{Matched scene stratum}
&
\shortstack{\textbf{Matched}\\\textbf{reference}}
&
\shortstack{\textbf{DENSEWORLD}\\\textbf{[\(K\)-fold]}}
&
\(\boldsymbol{\Delta}\)
&
\(\boldsymbol{R}\)
&
\shortstack{\textbf{Matched}\\\textbf{reference}}
&
\shortstack{\textbf{DENSEWORLD}\\\textbf{[\(K\)-fold]}}
&
\shortstack{\(\boldsymbol{\Delta}\)\\\textbf{pp}}
&
\(\boldsymbol{R}\)
\\
\midrule

Residential lane
&
\cellcolor{RefGray}\(1.8\)
&
\cellcolor{DWBlue}\(\mathbf{3.3}\)
&
\(+1.5\)
&
\(1.83{\times}\)
&
\cellcolor{RefGray}\(2.0\%\)
&
\cellcolor{DWBlue}\(\mathbf{3.5\%}\)
&
\(+1.5\)
&
\(1.75{\times}\)
\\

\rowcolor{RowGray}
Promenade
&
\cellcolor{RefGray}\(1.4\)
&
\cellcolor{DWBlue}\(\mathbf{2.5}\)
&
\(+1.1\)
&
\(1.79{\times}\)
&
\cellcolor{RefGray}\(1.5\%\)
&
\cellcolor{DWBlue}\(\mathbf{2.8\%}\)
&
\(+1.3\)
&
\(1.87{\times}\)
\\

Market
&
\cellcolor{RefGray}\(4.6\)
&
\cellcolor{DWBlue}\(\mathbf{12.4}\)
&
\(+7.8\)
&
\(2.70{\times}\)
&
\cellcolor{RefGray}\(4.2\%\)
&
\cellcolor{DWBlue}\(\mathbf{11.1\%}\)
&
\(+6.9\)
&
\(2.64{\times}\)
\\

\rowcolor{RowGray}
Heritage / tourist
&
\cellcolor{RefGray}\(1.7\)
&
\cellcolor{DWBlue}\(\mathbf{3.0}\)
&
\(+1.3\)
&
\(1.76{\times}\)
&
\cellcolor{RefGray}\(0.9\%\)
&
\cellcolor{DWBlue}\(\mathbf{1.4\%}\)
&
\(+0.5\)
&
\(1.56{\times}\)
\\

Flyover / underpass
&
\cellcolor{RefGray}\(2.6\)
&
\cellcolor{DWBlue}\(\mathbf{5.0}\)
&
\(+2.4\)
&
\(1.92{\times}\)
&
\cellcolor{RefGray}\(8.8\%\)
&
\cellcolor{DWBlue}\(\mathbf{20.8\%}\)
&
\(+12.0\)
&
\(2.36{\times}\)
\\

\rowcolor{RowGray}
Commercial
&
\cellcolor{RefGray}\(4.1\)
&
\cellcolor{DWBlue}\(\mathbf{11.0}\)
&
\(+6.9\)
&
\(2.68{\times}\)
&
\cellcolor{RefGray}\(3.9\%\)
&
\cellcolor{DWBlue}\(\mathbf{9.0\%}\)
&
\(+5.1\)
&
\(2.31{\times}\)
\\

\midrule

\rowcolor{PanelBlue}
\textbf{Equal-stratum mean}
&
\(\mathbf{2.70}\)
&
\(\mathbf{6.20}\)
&
\(\mathbf{+3.50}\)
&
\(\mathbf{2.30{\times}}\)
&
\(\mathbf{3.55\%}\)
&
\(\mathbf{8.10\%}\)
&
\(\mathbf{+4.55}\)
&
\(\mathbf{2.28{\times}}\)
\\

\bottomrule
\end{tabular}

\arrayrulecolor{black}

\vspace{3pt}
\begin{minipage}{0.97\textwidth}
\footnotesize
\textit{Interpretation.}
The ratios in the final row are ratios of the equal-stratum means, not
means of the six stratum-specific ratios. These quantities characterize
the measured operating regime; larger values do not denote conventional
benchmark performance.
\end{minipage}
\end{table*}

% =====================================================================
\subsection{Matched Cross-Dataset Comparison}
\label{app:matched_comparison}

The primary analysis matches DENSEWORLD drive-through clips separately
against road-level clips from BDD100K~\cite{yu2020bdd100k} and
nuScenes~\cite{caesar2020nuscenes}. Walk-through and aerial clips are
summarized separately and do not enter the matched driving-benchmark
comparison.

Matching is performed without access to the five axes or their
component measurements. Clips are first assigned to common strata
defined by capture mode, scene family, illumination, weather, and road
context. These nuisance attributes are obtained using the same fixed
metadata rules across datasets. Within each stratum, we perform
\(1{:}1\) nearest-neighbor matching without replacement over
standardized clip duration, valid image support, aspect ratio,
resolution, and common camera metadata. Covariates unavailable
consistently across all datasets are excluded from the primary matching
model.

Let \(\mathbf z_v\) denote the matching covariates for clip \(v\).
Post-matching balance is measured using the standardized mean
difference

\[
\operatorname{SMD}(z)
=
\frac{
\overline z_{\mathrm{DW}}
-
\overline z_{\mathrm{ref}}
}{
\sqrt{
(s_{\mathrm{DW}}^2+s_{\mathrm{ref}}^2)/2
}
}.
\]

The primary analysis retains common-support strata satisfying

\[
\max_{z\in\mathbf z}
|\operatorname{SMD}(z)|<0.1.
\]

We report the number of candidate and retained clips, independent source
groups, unmatched fraction, and maximum absolute SMD before and after
matching. Unmatched clips are excluded from the primary cross-dataset
estimate and retained only in DENSEWORLD-specific coverage summaries.

For axis \(k\), let \(\mathcal T_{k,u}\) be the eligible sampled frames
from matched clips belonging to source group \(u\). The source-level
summary is

\[
\overline m_{k,u}^{(d)}
=
\frac{1}{|\mathcal T_{k,u}|}
\sum_{t\in\mathcal T_{k,u}}
m_{k,t}^{(d)}.
\]

The matched-population estimate gives equal weight to every retained
source group:

\[
\widehat\mu_{r,k}^{(d)}
=
\frac{1}{|\mathcal U_{d,r}|}
\sum_{u\in\mathcal U_{d,r}}
\overline m_{k,u}^{(d)},
\]

where \(r\in\{\mathrm{BDD100K},\mathrm{nuScenes}\}\) identifies the
reference-specific matched population. The DENSEWORLD estimate is
therefore recomputed for its BDD100K-matched and nuScenes-matched
subsets.

For each reference \(r\), we report the absolute difference and ratio

\[
\Delta_{r,k}
=
\widehat\mu_{r,k}^{(\mathrm{DW})}
-
\widehat\mu_{r,k}^{(r)},
\qquad
R_{r,k}
=
\frac{
\widehat\mu_{r,k}^{(\mathrm{DW})}+\epsilon
}{
\widehat\mu_{r,k}^{(r)}+\epsilon
}.
\]

These contrasts describe separation under matched observable
conditions; they are not interpreted causally.

% =====================================================================
\subsection{Comparative Results, Uncertainty, and Sensitivity}
\label{app:five_axis_results}

Table~\ref{tab:scene_stratified_density_occupancy} reports the primary
scene-matched density and occupancy comparison. Entries follow the form \(X\,[X,X]\): a fold-aggregated
mean followed by its hierarchical \(95\%\) confidence interval. The
bracketed value after \textbf{DENSEWORLD} denotes the number of
executed folds.

Confidence intervals are estimated using a
\textbf{hierarchical clustered bootstrap}. DENSEWORLD resamples cities
and then source videos within cities; BDD100K resamples source videos;
and nuScenes resamples collection logs and scenes. All clips from a
sampled source group remain together, and matching is re-estimated
inside every bootstrap replicate. We report \(95\%\)
bias-corrected and accelerated (BCa) intervals for dataset means,
absolute differences, and ratios. Holm correction is applied jointly
to the ten primary contrasts: five axes against two reference datasets.

Robustness is evaluated over prespecified variations of
\(\tau_{\mathrm{det}}\), \(a_{\min}\), \(\tau_{\mathrm{ov}}\),
\(\rho\), \((\tau_t,\tau_d)\), and temporal sampling rate. We also
compare source and equal-stratum weighting, conditional and
unconditional normalization, and alternative common-support calipers.
Leave-one-city-out analysis tests whether any individual DENSEWORLD
city determines the reported effects.

Two automated negative controls test metric specificity.
\emph{Spatial permutation} preserves frame-level counts and class
frequencies while disrupting proximity and closest-approach edges.
\emph{Temporal permutation} preserves frame-level count and occupancy
while disrupting motion consistency and track-gap structure. Each
metric must attenuate under its corresponding control without
mechanically altering the axes that the control is intended to preserve.

An axis is interpreted as exhibiting stable regime separation only
when its effect direction is preserved across the prespecified
sensitivity grid, its clustered interval excludes zero, and the
conclusion survives leave-one-city-out analysis. The resulting evidence
is specific to the shared automated measurement system and matched
observable conditions. It does not imply exhaustive geographic
coverage, causal effects of dataset membership, or universal
superiority over existing driving benchmarks.

\section{Segmentations, and Training}
\label{app:factor_reproducibility}

This appendix specifies the model-side supervision and executed training
protocol for FactorJEPA. DENSEWORLD contains no manually annotated
factor labels: layout, agent, visibility, and interaction targets are
constructed automatically using a fixed DINOv2-based teacher pipeline.
These targets provide structured supervision during training and
teacher-relative diagnostics during evaluation; they are distinct from
the frozen evaluators used for headline future-prediction, motion, and
RGB metrics.

We first document target construction and reliability weighting, then
specify the FactorJEPA architecture, optimization procedure, controlled
baselines, and resource accounting. Table~\ref{tab:factor_targets}
defines the complete target contract, while
Algorithm~\ref{alg:factorjepa_training} records the executed training
path.

% =====================================================================
\subsection{Frozen Teacher and Factor-Target Construction}
\label{app:factor_targets}

\paragraph{Region extraction.}
For each privacy-filtered clip \(\widetilde x_n\), the frozen DINOv2
encoder~\cite{oquab2024dinov2} produces spatiotemporal visual features.
The executed preprocessing operator converts these features into
region-level pseudo-annotations:

\begin{equation*}
\begin{aligned}
\mathcal P_n
&=
\mathcal D_{\mathrm{pre}}
\left(
F_{\mathrm{DINOv2}}(\widetilde x_n)
\right)
\\
&=
\left\{
\left(
m_{n,t}^{(i)},
b_{n,t}^{(i)},
u_{n,t}^{(i)},
p_{n,t}^{(i)}
\right)
\right\}_{t,i}.
\end{aligned}
\end{equation*}

Here, \(m_{n,t}^{(i)}\), \(b_{n,t}^{(i)}\),
\(u_{n,t}^{(i)}\), and \(p_{n,t}^{(i)}\) denote the visible mask,
bounding box, visual descriptor, and confidence of region \(i\) at
time \(t\). The DINOv2 checkpoint, extracted feature layers, inference
resolution, spatial stride, normalization, region-construction rule,
and confidence filtering are fixed before model training and shared by
all splits.

\paragraph{Temporal association.}
The association operator \(\mathcal A\) links compatible regions into
tracklets:

\begin{equation*}
\mathcal A(\mathcal P_n)
=
\left\{
\tau_n^{(i)}
\right\}_{i=1}^{N_n^A},
\qquad
\tau_n^{(i)}
=
\left\{
m_{n,t}^{(i)},
b_{n,t}^{(i)},
u_{n,t}^{(i)}
\right\}_{t=t_s^{(i)}}^{t_e^{(i)}} .
\end{equation*}

Association combines region appearance, mask or box overlap, and
motion continuity. Track initiation, termination, admissible temporal
gaps, and re-identification thresholds are held fixed across training,
validation, and test. Short gaps may preserve a track identity; gaps
beyond the admissible window terminate the tracklet.

\paragraph{Structural targets.}
A deterministic structural map converts region geometry, temporal
continuity, visibility evidence, and relative motion into four targets:

\[
\Psi_{\mathrm{struct}}
\left(
\mathcal P_n,\mathcal A(\mathcal P_n)
\right)
\longmapsto
\left\{
\left(T_{n,k},q_{n,k}\right)
\right\}_{k\in\{L,A,V,I\}} .
\]

The targets \(T_{n,L}\), \(T_{n,A}\), \(T_{n,V}\), and \(T_{n,I}\)
encode layout, agents, visibility, and tracklet-derived interactions.
Each \(q_{n,k}\in[0,1]\) is a reliability weight constructed from the
underlying region confidence and temporal consistency. Low-confidence
targets are down-weighted rather than converted into negative labels;
targets below the exclusion threshold do not contribute to the
corresponding factor loss.

\begin{table*}[ht!]
\centering
\caption{
\textbf{Frozen teacher-to-training contract for FactorJEPA.}
DENSEWORLD contains no manually annotated factor labels. All targets
are generated automatically by the fixed DINOv2-based pipeline before
optimization. The table identifies, for each factor, the teacher
evidence, deterministic target operator, output support, reliability
contract, and trainable consumer. Here, \(p_L,p_A,p_I\) are the
executed target dimensions, \(N_n^A\) is the number of retained agent
tracklets, and \(\mathcal E_n\) is the candidate interaction graph.
}
\label{tab:factor_targets}

\scriptsize
\setlength{\tabcolsep}{3pt}
\renewcommand{\arraystretch}{1.22}
\arrayrulecolor{FTBorder}

\begin{tabular}{
    >{\raggedright\arraybackslash}p{0.088\textwidth}
    >{\raggedright\arraybackslash}p{0.165\textwidth}
    >{\raggedright\arraybackslash}p{0.215\textwidth}
    >{\centering\arraybackslash}p{0.115\textwidth}
    >{\raggedright\arraybackslash}p{0.185\textwidth}
    >{\raggedright\arraybackslash}p{0.120\textwidth}
}
\toprule

\rowcolor{FTHeader}
\multicolumn{1}{c}{
\textcolor{white}{\textbf{Factor}}
}
&
\multicolumn{4}{c}{
\textcolor{white}{
\textbf{Frozen teacher-side target contract}
}
}
&
\multicolumn{1}{c}{
\textcolor{white}{
\textbf{Training side}
}
}
\\

\rowcolor{FTSubheader}
\textcolor{white}{\textbf{Target}}
&
\textcolor{white}{\textbf{Teacher evidence}}
&
\textcolor{white}{\textbf{Executed operator}}
&
\textcolor{white}{\textbf{Output support}}
&
\textcolor{white}{\textbf{Reliability contract}}
&
\textcolor{white}{\textbf{Consumer}}
\\
\midrule

% ---------------------------------------------------------------------
\rowcolor{LayoutRow}
\cellcolor{LayoutCell}
\textbf{Layout}
\newline
\(T_{n,L}\)
&
DINOv2 spatial-feature grid, valid-image support, dynamic-region union,
and temporal feature persistence.
&
\(\Psi_L\) suppresses pixels assigned to retained dynamic tracklets,
aggregates the remaining temporally persistent support, and applies the
fixed layout-target projection.
&
\(
T_{n,L}
\in
\mathbb R^{p_L}
\)
\newline
\emph{clip-level}
&
\(\omega_{n,L}\in[0,1]\) combines valid background coverage, feature
confidence, and temporal stability. The target is retained only when
\(r_{n,L}=\mathbf 1[\omega_{n,L}\ge\tau_L]\).
&
\(P_L(C_L)\)
\newline
via
\(\mathcal L_{\mathrm{factor}}\)
\\

\addlinespace[1.5pt]

% ---------------------------------------------------------------------
\rowcolor{AgentRow}
\cellcolor{AgentCell}
\textbf{Agent}
\newline
\(T_{n,A}\)
&
Retained masks, boxes, region descriptors, confidences, and temporally
associated tracklets
\(\{\tau_n^{(i)}\}_{i=1}^{N_n^A}\).
&
\(\Psi_A\) constructs an object-centric state for each valid tracklet
and aggregates the retained states after confidence, visible-support,
and temporal-continuity filtering.
&
\(
T_{n,A}
\in
\mathbb R^{p_A}
\)
\newline
\emph{clip-level}
&
\(\omega_{n,A}\in[0,1]\) combines retained-region confidence, visible
mask support, track continuity, and valid temporal coverage. Missing
or rejected tracks are not converted into negative agents.
&
\(P_A(C_A)\)
\newline
via
\(\mathcal L_{\mathrm{factor}}\)
\\

\addlinespace[1.5pt]

% ---------------------------------------------------------------------
\rowcolor{VisibilityRow}
\cellcolor{VisibilityCell}
\textbf{Visibility}
\newline
\(T_{n,V}\)
&
Visible-mask support, valid-image-boundary truncation, region
confidence, track continuity, and short interior track gaps.
&
\(\Psi_V\) produces a soft visibility value for every retained agent.
Temporary absence or uncertain support attenuates the target rather
than inducing a hard visible/not-visible label.
&
\(
\left\{
T_{n,V}^{(i)}
\right\}_{i=1}^{N_n^A}
\)
\newline
\(T_{n,V}^{(i)}\in[0,1]\)
&
Each agent receives
\(\omega_{n,V}^{(i)}\in[0,1]\), determined by region confidence,
temporal support, and track consistency. Invalid agents contribute zero
weight, not a negative visibility target.
&
\(g_V\)
\newline
via weighted
\(\mathcal L_V\)
\\

\addlinespace[1.5pt]

% ---------------------------------------------------------------------
\rowcolor{InteractionRow}
\cellcolor{InteractionCell}
\textbf{Interaction}
\newline
\(T_{n,I}\)
&
Pairs of retained tracklets, relative position, relative motion,
visible support, temporal overlap, and pairwise proximity.
&
\(\Psi_I\) constructs
\(\mathcal E_n\subseteq
\{(i,j):i\ne j\}\), forms normalized pair descriptors, and aggregates
valid pair states into the interaction target.
&
\(
\mathcal E_n,
\quad
T_{n,I}
\in
\mathbb R^{p_I}
\)
\newline
\emph{graph + clip target}
&
Pair reliability
\(\omega_{n,I}^{(ij)}\in[0,1]\) combines the two endpoint confidences,
joint temporal support, and pair validity. These weights induce the
clip-level reliability \(\omega_{n,I}\).
&
\(P_I(C_I)\)
via
\(\mathcal L_{\mathrm{factor}}\);
\(g_I,g_W\) use
\(\mathcal E_n\)
\\

\midrule

\rowcolor{FTContract}
\multicolumn{6}{
>{\raggedright\arraybackslash}p{0.955\textwidth}
}{
\textbf{Shared reliability contract.}
All reliability variables are stop-gradient quantities in \([0,1]\).
For \(k\in\{L,A,I\}\), the effective loss weight is
\(r_{n,k}\omega_{n,k}\), where
\(r_{n,k}=\mathbf 1[\omega_{n,k}\ge\tau_k]\).
Factor losses are normalized by the total effective weight within the
minibatch, so variations in retained coverage do not directly rescale
their contribution. Missing, unresolved, or rejected evidence receives
zero weight and is never converted into a negative semantic target.
}
\\

\addlinespace[1pt]

\rowcolor{FTContract}
\multicolumn{6}{
>{\raggedright\arraybackslash}p{0.955\textwidth}
}{
\textbf{Gradient and provenance contract.}
The DINOv2 encoder, structural operators
\(\Psi_L,\Psi_A,\Psi_V,\Psi_I\), tracklets, candidate graph, targets,
and reliability weights remain frozen. Gradients propagate only through
the FactorJEPA branches, factor heads, synthesis dictionaries, and the
selected online-encoder blocks. Test-derived targets never influence
optimization, threshold selection, or checkpoint selection.
}
\\

\bottomrule
\end{tabular}

\arrayrulecolor{black}

\vspace{3pt}
\begin{minipage}{0.97\textwidth}
\footnotesize
\textit{Interaction-loss clarification.}
The interaction target \(T_{n,I}\) supervises \(P_I(C_I)\) through
\(\mathcal L_{\mathrm{factor}}\). The candidate graph
\(\mathcal E_n\) defines the support of the interaction branch, while
\(\mathcal L_{\mathrm{sparse}}\) regularizes the predicted soft edge
weights \(w_n^{(ij)}\); it does not directly consume
\(T_{n,I}\).
\end{minipage}
\end{table*}
\paragraph{Reliability-normalized supervision.}
For \(k\in\{L,A,I\}\), the factor heads are trained using

\[
\mathcal L_{\mathrm{factor}}
=
\sum_{k\in\{L,A,I\}}
\frac{
\sum_{n=1}^{N}
q_{n,k}\,
\ell_k\!\left(\widehat T_{n,k},T_{n,k}\right)
}{
\epsilon+\sum_{n=1}^{N}q_{n,k}
}.
\]

The denominator makes the scale of each factor loss insensitive to its
retained target coverage. Visibility is optimized separately through
\(\mathcal L_V\), using per-agent targets and reliability weights.

\paragraph{Target coverage and sensitivity.}
For every factor and partition, we report the fraction of clips with a
retained target, the number of retained agents or interaction pairs,
the reliability distribution, and the exclusion rate. Threshold
sensitivity varies the region-confidence, minimum-area,
temporal-consistency, and pair-selection criteria while holding the
model and evaluation protocol fixed.

\begin{figure*}[t]
\centering
\includegraphics[
    width=\textwidth
]{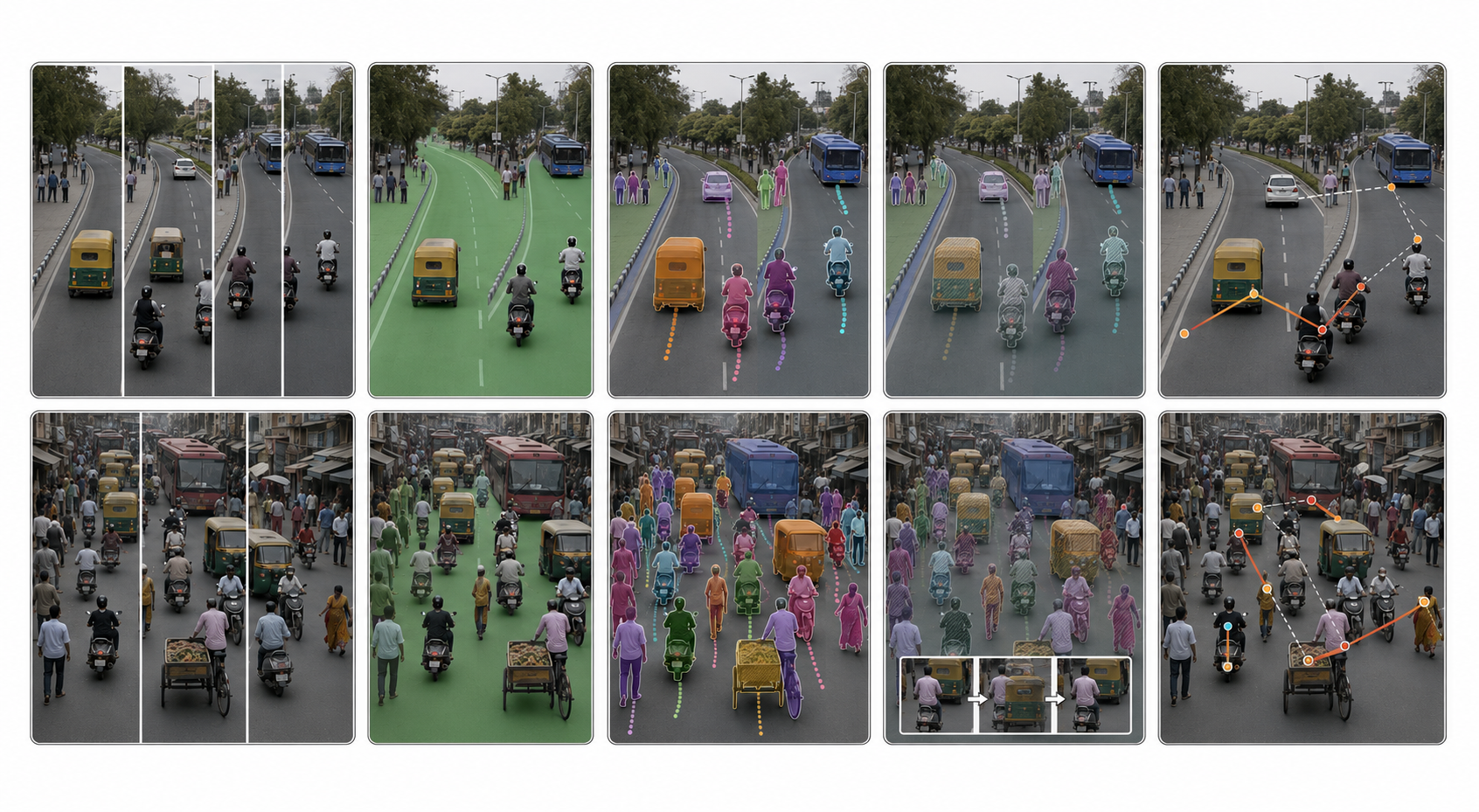}
\caption{
\textbf{Conceptual end-to-end construction of FactorJEPA targets.}
The upper band shows the frozen teacher-side pipeline:
a privacy-filtered clip is processed by DINOv2 to obtain region
evidence---masks, boxes, descriptors, and confidence---which is linked
into tracklets by temporal association. The deterministic structural
map then produces layout, agent, visibility, and interaction targets
\(\{T_L,T_A,T_V,T_I\}\) together with stop-gradient reliability weights
\(\omega\). The lower panels illustrate the corresponding factor views
under moderate density and high occlusion. Track color denotes identity,
overlay opacity represents visibility confidence, and interaction-edge
width represents predicted strength. The scenes are synthetic and
illustrate the target semantics; they are not empirical DINOv2 outputs.
}
\label{fig:factor_target_examples}
\end{figure*}

% =====================================================================
\subsection{FactorJEPA Architecture and Prediction Protocol}
\label{app:factor_architecture}

FactorJEPA retains the V-JEPA~2.1 online encoder, momentum target
encoder, context and target masking policy, and future-latent prediction
objective. It replaces the monolithic predictor with layout, agent,
interaction, visibility, and interaction-strength branches.

For target token \(p\), the context encoder first produces the
target-conditioned query

\[
q_{n,p}
=
Q(h_n,\pi_p,M_t),
\qquad
h_n=f_\theta(M_c\odot x_n),
\]

where \(\pi_p\) is the target-token positional encoding. The predictor
then produces three coordinate blocks:

\[
c_{n,p}
=
\operatorname{col}
\left(
c_{n,L,p},
c_{n,A,p},
c_{n,I,p}
\right)
\in
\mathbb R^{r_L+r_A+r_I}.
\]

\paragraph{Layout coordinates.}
The layout branch extracts slowly varying spatial support:

\[
c_{n,L}=g_L(q_n)\in\mathbb R^{r_L}.
\]

Its factor head predicts
\(\widehat T_{n,L}=P_L(c_{n,L})\).

\paragraph{Visibility-gated agent coordinates.}
For object-centric agent state \(o_n^{(i)}\), the agent branch and
visibility gate produce

\begin{equation*}
\begin{aligned}
s_n^{(i)}
&=
g_A\!\left(q_n,o_n^{(i)}\right)
\in\mathbb{R}^{r_A},
\\
v_n^{(i)}
&=
\sigma\!\left(
g_V\!\left(q_n,o_n^{(i)}\right)
\right)
\in(0,1).
\end{aligned}
\end{equation*}

The aggregated agent coordinate is

\[
c_{n,A}
=
\frac{
\sum_{i=1}^{N_n^A}
v_n^{(i)}s_n^{(i)}
}{
\epsilon+
\sum_{i=1}^{N_n^A}v_n^{(i)}
}.
\]

Normalization limits direct sensitivity to the detected-agent count,
while the soft gate attenuates uncertain or occluded observations
without removing them discontinuously.

\paragraph{Sparse interaction coordinates.}
Let \(\mathcal E_n\) be the retained candidate-pair set. For
\((i,j)\in\mathcal E_n\), the interaction branch predicts a normalized
pair state and soft interaction strength:

\[
\overline e_n^{(ij)}
=
\frac{
g_I(q_n,o_n^{(i)},o_n^{(j)})
}{
\epsilon+
\left\|
g_I(q_n,o_n^{(i)},o_n^{(j)})
\right\|_2
},
\]

\[
w_n^{(ij)}
=
\sigma
\left(
g_W(q_n,o_n^{(i)},o_n^{(j)})
\right).
\]

With \(M_n=\max\{1,|\mathcal E_n|\}\),

\[
c_{n,I}
=
\frac{1}{M_n}
\sum_{(i,j)\in\mathcal E_n}
w_n^{(ij)}
\overline e_n^{(ij)},
\]

and \(c_{n,I}=0\) when \(\mathcal E_n=\varnothing\). The corresponding
locality penalty is

\[
\mathcal L_{\mathrm{sparse}}
=
\frac{1}{N}
\sum_{n=1}^{N}
\frac{1}{M_n}
\sum_{(i,j)\in\mathcal E_n}
w_n^{(ij)}.
\]

\paragraph{Factor synthesis.}
For a minibatch of \(N\) clips and \(m\) target tokens, stack the
coordinate blocks as \(C_L,C_A,C_I\) and let
\(A_L,A_A,A_I\) be the learned synthesis dictionaries. The predicted
future-token matrix is

\[
\widehat Y
=
C_LA_L^\top
+
C_AA_A^\top
+
C_IA_I^\top .
\]

Thus, each factor follows a distinct computational path, receives
factor-specific supervision, and contributes through its own synthesis
dictionary. This establishes architectural block separation while
preserving the complete future-token geometry.

The token-level JEPA objective is

\[
\mathcal L_{\mathrm{JEPA}}
=
\frac{1}{Nm}
\left\|
\widehat Y-Y^\star
\right\|_F^2,
\]

where \(Y^\star\) is produced by the momentum target encoder under
stop-gradient.

\paragraph{Cross-factor separation.}
Linear leakage is penalized through the off-block covariance norm
\(\mathcal L_{\mathrm{cov}}\). Nonlinear leakage is penalized through
normalized RBF-kernel alignment \(\mathcal L_{\mathrm{nlin}}\), using
the stop-gradient median pairwise distance as the bandwidth. The
combined separation objective is

\[
\mathcal L_{\mathrm{sep}}
=
\mathcal L_{\mathrm{cov}}
+
\beta_{\mathrm{nlin}}
\mathcal L_{\mathrm{nlin}}.
\]

These losses suppress cross-channel shortcuts but do not assert
statistical independence, causal disentanglement, or coordinate-level
identifiability.

% =====================================================================
\subsection{Optimization, Model Selection, and Executed Algorithm}
\label{app:training_algorithm}

The complete objective is

\[
\mathcal L
=
\mathcal L_{\mathrm{JEPA}}
+
\lambda_{\mathrm{sep}}\mathcal L_{\mathrm{sep}}
+
\lambda_{\mathrm{sparse}}\mathcal L_{\mathrm{sparse}}
+
\lambda_V\mathcal L_V
+
\lambda_{\mathrm{sup}}\mathcal L_{\mathrm{factor}}.
\]

The five terms respectively enforce future-latent prediction,
cross-factor separation, interaction locality, visibility calibration,
and semantic anchoring. FactorJEPA trains the factorized predictor and
the selected top-\(K\) online-encoder blocks:

\[
\Theta_{\mathrm{train}}
=
\Theta_{\mathrm{fact}}
\cup
\Theta_{\mathrm{top}\text{-}K},
\qquad
\Theta_{\mathrm{fact}}
=
\{
\Theta_L,\Theta_A,\Theta_I,\Theta_V,\Theta_W,A
\}.
\]

All lower online-encoder blocks remain frozen. The target encoder is
updated only through the executed exponential-moving-average rule.

% Requires:
% \usepackage{algorithm}
% \usepackage{algpseudocode}

\refstepcounter{factoralgorithm}
\label{alg:factorjepa_training}

\begin{tcolorbox}[
    enhanced jigsaw,
    breakable,
    colback=AlgBackground,
    colframe=AlgBorder,
    colbacktitle=AlgTitleBackground,
    coltitle=AlgAccent,
    fonttitle=\bfseries\small,
    title={
        Algorithm~\thefactoralgorithm:
        Executed FactorJEPA Training Procedure
    },
    title after break={
        Algorithm~\thefactoralgorithm\ (continued):
        Executed FactorJEPA Training Procedure
    },
    boxrule=0.55pt,
    arc=1.2mm,
    outer arc=1.2mm,
    left=5pt,
    right=5pt,
    top=5pt,
    bottom=5pt,
    before skip=8pt,
    after skip=8pt,
    pad at break*=2mm,
    fontupper=\footnotesize
]

\textbf{Inputs.}
Source-grouped training clips
\(\mathcal D_{\mathrm{train}}\);
cached DINOv2 target store
\(\mathcal Z_{\mathrm{DINO}}\);
mask sampler \(\mathcal M\);
online encoder \(f_\theta\);
momentum encoder \(\overline f_{\bar\theta}\);
factorized predictor \(g_\phi\);
optimizer and learning-rate schedule;
loss coefficients
\(\lambda_{\mathrm{sep}},
\lambda_{\mathrm{sparse}},
\lambda_V,
\lambda_{\mathrm{sup}}\);
EMA schedule \(\mu_s\);
gradient bound \(\gamma\);
and validation interval \(S_{\mathrm{val}}\).

\smallskip

\textbf{Output.}
Validation-selected checkpoint
\((\theta^\star,\phi^\star)\), together with its executed
configuration, threshold set, optimizer state, and random-state record.

\begin{algsteps}

% =====================================================================
\AlgPhase{Initialization and parameter restriction}

\item
Initialize the online encoder \(f_\theta\) and momentum encoder
\(\overline f_{\bar\theta}\) from the same pretrained V-JEPA
checkpoint.

\item
Replace the monolithic V-JEPA predictor with

\[
g_\phi
=
\{
g_L,g_A,g_I,g_V,g_W,
A_L,A_A,A_I,
P_L,P_A,P_I
\}.
\]

\item
Copy the online-encoder parameters to the momentum encoder:

\[
\bar\theta\gets\theta.
\]

\item
Freeze the DINOv2 target-construction pipeline, the momentum encoder,
and all online-encoder blocks below the selected top-\(K\) set.

\item
Define the trainable parameter set

\[
\Theta_{\mathrm{train}}
=
\Theta_{\mathrm{fact}}
\cup
\Theta_{\mathrm{top}\text{-}K}.
\]

\item
Initialize the optimizer, learning-rate scheduler, validation record,
and best-checkpoint state.

% =====================================================================
\AlgPhase{For each optimization step \(s=1,\ldots,S\):
minibatch and structural evidence}

\item
Sample a source-grouped minibatch

\[
\{\widetilde x_n\}_{n=1}^{N}
\sim
\mathcal D_{\mathrm{train}}.
\]

\item
For every clip \(n\), retrieve the cached teacher packet

\[
\mathcal Z_n
=
\left\{
\mathcal P_n,\,
\mathcal A(\mathcal P_n),\,
\mathcal E_n,\,
(T_{n,k},\omega_{n,k})_{k\in\{L,A,V,I\}}
\right\}.
\]

Here, \(\mathcal P_n\) contains region-level pseudo-annotations,
\(\mathcal A(\mathcal P_n)\) contains associated tracklets, and
\(\mathcal E_n\) contains retained interaction candidates.

\item
Construct the valid-target indicator

\[
r_{n,k}
=
\mathbf 1
\left[
\omega_{n,k}\ge\tau_k
\right],
\qquad
k\in\{L,A,V,I\}.
\]

Targets below the factor-specific threshold are excluded from that
factor loss rather than converted into negative supervision.

\item
Sample context and target masks from the shared masking policy:

\[
(M_c,M_t)\sim\mathcal M.
\]

\item
Construct masked context and target inputs:

\[
\widetilde x_n^{\,c}
=
M_c\odot\widetilde x_n,
\qquad
\widetilde x_n^{\,t}
=
M_t\odot\widetilde x_n.
\]

% =====================================================================
\AlgPhase{Online context path and stop-gradient target path}

\item
Encode the visible context using the online encoder:

\[
h_n
=
f_\theta
\left(
\widetilde x_n^{\,c}
\right).
\]

\item
Compute the future target representation using the momentum encoder:

\[
Y_n^\star
=
\operatorname{sg}
\left[
\overline f_{\bar\theta}
\left(
\widetilde x_n^{\,t}
\right)
\right].
\]

No gradient is propagated through \(Y_n^\star\).

\item
For each target token \(p\), form the target-conditioned query

\[
q_{n,p}
=
Q(h_n,\pi_p,M_t),
\]

where \(\pi_p\) is the target-token positional encoding.

% =====================================================================
\AlgPhase{Layout-factor prediction}

\item
Predict the layout coordinate for every target token:

\[
c_{n,L,p}
=
g_L(q_{n,p})
\in\mathbb R^{r_L}.
\]

\item
Predict the corresponding layout target:

\[
\widehat T_{n,L}
=
P_L(c_{n,L}).
\]

% =====================================================================
\AlgPhase{Visibility-gated agent prediction}

\item
Construct each object-centric state \(o_n^{(i)}\) from the cached
region descriptor, visible mask, bounding box, confidence, and
tracklet state.

\item
For every retained agent \(i\), predict its factor coordinate:

\[
s_{n,p}^{(i)}
=
g_A
\left(
q_{n,p},o_n^{(i)}
\right)
\in\mathbb R^{r_A}.
\]

\item
Predict the corresponding soft visibility gate:

\[
v_{n,p}^{(i)}
=
\sigma
\left(
g_V
\left(
q_{n,p},o_n^{(i)}
\right)
\right)
\in(0,1).
\]

\item
Aggregate the agent coordinates under soft visibility:

\[
c_{n,A,p}
=
\frac{
\sum_i
v_{n,p}^{(i)}s_{n,p}^{(i)}
}{
\epsilon+
\sum_i v_{n,p}^{(i)}
}.
\]

\item
Predict the agent target:

\[
\widehat T_{n,A}
=
P_A(c_{n,A}).
\]

% =====================================================================
\AlgPhase{Sparse interaction prediction}

\item
For every candidate pair
\((i,j)\in\mathcal E_n\), compute the normalized pair state

\[
\overline e_{n,p}^{(ij)}
=
\frac{
g_I
\left(
q_{n,p},
o_n^{(i)},
o_n^{(j)}
\right)
}{
\epsilon+
\left\|
g_I
\left(
q_{n,p},
o_n^{(i)},
o_n^{(j)}
\right)
\right\|_2
}.
\]

\item
Predict the soft interaction strength

\[
w_{n,p}^{(ij)}
=
\sigma
\left(
g_W
\left(
q_{n,p},
o_n^{(i)},
o_n^{(j)}
\right)
\right)
\in(0,1).
\]

\item
Set

\[
M_n=\max\{1,|\mathcal E_n|\}
\]

and aggregate the interaction coordinate:

\[
c_{n,I,p}
=
\frac{1}{M_n}
\sum_{(i,j)\in\mathcal E_n}
w_{n,p}^{(ij)}
\overline e_{n,p}^{(ij)}.
\]

If \(\mathcal E_n=\varnothing\), set
\(c_{n,I,p}=\mathbf 0\).

\item
Predict the interaction target:

\[
\widehat T_{n,I}
=
P_I(c_{n,I}).
\]

% =====================================================================
\AlgPhase{Factor synthesis}

\item
Stack the layout, agent, and interaction coordinates over clips and
target tokens to obtain \(C_L,C_A,C_I\).

\item
Synthesize the predicted future-token matrix:

\[
\widehat Y
=
C_LA_L^\top
+
C_AA_A^\top
+
C_IA_I^\top.
\]

% =====================================================================
\AlgPhase{Reliability-weighted objectives}

\item
Compute the future-latent prediction loss:

\[
\mathcal L_{\mathrm{JEPA}}
=
\frac{1}{Nm}
\left\|
\widehat Y-Y^\star
\right\|_F^2.
\]

\item
Compute reliability-normalized factor supervision:

\[
\mathcal L_{\mathrm{factor}}
=
\sum_{k\in\{L,A,I\}}
\frac{
\sum_n
r_{n,k}\omega_{n,k}
\ell_k
\left(
\widehat T_{n,k},T_{n,k}
\right)
}{
\epsilon+
\sum_n
r_{n,k}\omega_{n,k}
}.
\]

\item
Compute reliability-weighted visibility calibration:

\[
\mathcal L_V
=
\operatorname{WBCE}
\left(
\{v_n^{(i)}\},
\{T_{n,V}^{(i)}\};
\{\omega_{n,V}^{(i)}\}
\right).
\]

\item
Compute interaction sparsity:

\[
\mathcal L_{\mathrm{sparse}}
=
\frac{1}{N}
\sum_{n=1}^{N}
\frac{1}{M_n}
\sum_{(i,j)\in\mathcal E_n}
w_n^{(ij)}.
\]

\item
Compute the off-block covariance penalty
\(\mathcal L_{\mathrm{cov}}\).

\item
Compute the normalized RBF-kernel dependence penalty
\(\mathcal L_{\mathrm{nlin}}\), using stop-gradient median pairwise
distances as kernel bandwidths.

\item
Combine the linear and nonlinear separation terms:

\[
\mathcal L_{\mathrm{sep}}
=
\mathcal L_{\mathrm{cov}}
+
\beta_{\mathrm{nlin}}
\mathcal L_{\mathrm{nlin}}.
\]

\item
Assemble the complete objective:

\[
\mathcal L
=
\mathcal L_{\mathrm{JEPA}}
+
\lambda_{\mathrm{sep}}\mathcal L_{\mathrm{sep}}
+
\lambda_{\mathrm{sparse}}\mathcal L_{\mathrm{sparse}}
+
\lambda_V\mathcal L_V
+
\lambda_{\mathrm{sup}}\mathcal L_{\mathrm{factor}}.
\]

% =====================================================================
\AlgPhase{Restricted optimization and momentum update}

\item
Clear gradients on \(\Theta_{\mathrm{train}}\).

\item
Backpropagate

\[
\nabla_{\Theta_{\mathrm{train}}}\mathcal L.
\]

No gradient enters the cached targets, DINOv2, or the momentum encoder.

\item
Clip the trainable gradient norm at \(\gamma\).

\item
Apply the optimizer update only to

\[
\Theta_{\mathrm{fact}}
\cup
\Theta_{\mathrm{top}\text{-}K}.
\]

\item
Advance the learning-rate and loss-weight schedules.

\item
Update the momentum encoder:

\[
\bar\theta
=
\mu_s\bar\theta
+
(1-\mu_s)\theta.
\]

\item
Record the component losses, effective target weights, retained target
coverage, mean visibility gate, interaction sparsity, learning rate,
and gradient norm.

% =====================================================================
\AlgPhase{Validation-only checkpoint selection}

\item
Whenever \(s\bmod S_{\mathrm{val}}=0\), compute the prespecified
validation score \(J_{\mathrm{val}}(\theta,\phi)\).

\item
If \(J_{\mathrm{val}}\) improves the stored validation record, save

\[
(\theta^\star,\phi^\star)
\gets
(\theta,\phi),
\]

together with the optimizer state, executed configuration, threshold
set, fold identifier, and random-state record.

\item
Continue optimization until \(s=S\). The test partition is never
consulted during gradient updates, hyperparameter selection, threshold
selection, or checkpoint selection.

% =====================================================================
\AlgPhase{Return}

\item
Return the validation-selected checkpoint

\[
(\theta^\star,\phi^\star).
\]

\end{algsteps}
\end{tcolorbox}

Optimization hyperparameters, loss coefficients, parameter precision,
gradient clipping, and momentum schedule are fixed before final test
evaluation. Checkpoints are selected using the prespecified validation
criterion; test metrics never influence the number of steps,
hyperparameters, thresholds, or checkpoint selection.

% =====================================================================
\subsection{Baseline Matching and Attribution Controls}
\label{app:baseline_configuration}

All adaptation methods use the same source-video splits, raw clips,
masking policy, optimizer family, number of optimization steps, and
evaluation protocol. Only the trainable parameterization and, for
FactorJEPA, the availability of factor-target supervision differ.

The comparison implements the attribution chain

\begin{equation*}
\resizebox{0.98\columnwidth}{!}{$\displaystyle
\underbrace{
\{\text{LoRA},\text{DoRA},\text{Auto-RGN}\}
}_{\text{generic adaptation}}
\;\longrightarrow\;
\underbrace{
\text{FactorJEPA-RAW}
}_{\text{factorized architecture}}
\;\longrightarrow\;
\underbrace{
\text{FactorJEPA}
}_{\text{architecture + factor targets}}
$}
\end{equation*}

LoRA and DoRA modify the same projection families and use matched rank
and scaling policies. Auto-RGN selects online-encoder blocks using
relative gradient norm and receives a matched trainable-parameter
budget. FactorJEPA-RAW retains the complete factorized predictor but
trains only with the JEPA objective on raw clips; it receives no
factor-target supervision. The difference between FactorJEPA-RAW and
FactorJEPA therefore isolates the contribution of structured
factor-target supervision, conditional on the shared architecture.

Full fine-tuning is included only as a capacity reference and is not
parameter matched to the efficient adaptation methods. Frozen V-JEPA
provides the no-adaptation reference. Exact trainable modules,
parameter counts, ranks, selected blocks, and compute measurements are
consolidated in Table~\ref{tab:reproducibility_configuration}.

% Restrained, print-safe palette
\definecolor{CfgNavy}{HTML}{18364B}
\definecolor{CfgBlue}{HTML}{E4EEF7}
\definecolor{CfgCyan}{HTML}{E4F3F1}
\definecolor{CfgGold}{HTML}{FFF1CC}
\definecolor{CfgRose}{HTML}{F9E4E7}
\definecolor{CfgGreen}{HTML}{E1F1E6}
\definecolor{CfgGray}{HTML}{F3F5F6}
\definecolor{CfgInk}{HTML}{263A47}

% =====================================================================
% longtable cannot be placed inside table/table*
% =====================================================================
% page break removed for single-column continuous flow
\onecolumn

\begingroup
\scriptsize
\setlength{\LTleft}{0pt}
\setlength{\LTright}{0pt}
\setlength{\LTpre}{4pt}
\setlength{\LTpost}{4pt}
\setlength{\tabcolsep}{5pt}
\renewcommand{\arraystretch}{1.16}
\arrayrulecolor{CfgNavy!45}

\begin{longtable}{
    >{\RaggedRight\arraybackslash}p{0.18\textwidth}
    >{\RaggedRight\arraybackslash}p{0.245\textwidth}
    >{\RaggedRight\arraybackslash}p{0.245\textwidth}
    >{\RaggedRight\arraybackslash}p{0.27\textwidth}
}

\caption{
\textbf{Canonical FactorJEPA architecture, optimization, baseline, and
resource configuration.}
The table specifies a complete reproducible reference configuration for
the V-JEPA~2.1 ViT-G and ViT-g experiments. Backbone identifiers and
architectural scales follow the official V-JEPA~2.1 release. All
project-specific optimization, factorization, adaptation, and profiling
settings are fixed below. Resource entries are prespecified execution
ceilings; they must not be described as measured consumption unless
verified against profiling logs.
}
\label{tab:reproducibility_configuration}
\\

% ---------------------------------------------------------------------
% First-page header
% ---------------------------------------------------------------------
\toprule
\rowcolor{CfgNavy}
\color{white}\textbf{Configuration item}
&
\color{white}\textbf{V-JEPA~2.1 ViT-G}
&
\color{white}\textbf{V-JEPA~2.1 ViT-g}
&
\color{white}\textbf{Control, interpretation, or measurement rule}
\\
\midrule
\endfirsthead

% ---------------------------------------------------------------------
% Repeated header
% ---------------------------------------------------------------------
\multicolumn{4}{c}{
\textbf{Table~\thetable\ continued:
Canonical FactorJEPA configuration}
}
\\[3pt]
\toprule
\rowcolor{CfgNavy}
\color{white}\textbf{Configuration item}
&
\color{white}\textbf{V-JEPA~2.1 ViT-G}
&
\color{white}\textbf{V-JEPA~2.1 ViT-g}
&
\color{white}\textbf{Control, interpretation, or measurement rule}
\\
\midrule
\endhead

% ---------------------------------------------------------------------
% Continued-page footer
% ---------------------------------------------------------------------
\midrule
\multicolumn{4}{r}{
\textit{Continued on the next page}
}
\\
\endfoot

% ---------------------------------------------------------------------
% Final footer
% ---------------------------------------------------------------------
\bottomrule
\endlastfoot

% =====================================================================
% PANEL A
% =====================================================================
\rowcolor{CfgBlue}
\multicolumn{4}{l}{
\textcolor{CfgInk}{
\textbf{Panel A: Backbone, teacher, and video preprocessing}
}}
\\

\textbf{Backbone identifier}
&
\texttt{vjepa2\_1\_vit\_gigantic\_384}
&
\texttt{vjepa2\_1\_vit\_giant\_384}
&
Official V-JEPA~2.1 PyTorch-Hub model identifiers. Both checkpoints
remain the unique initialization source for all compared methods.
\\

\rowcolor{CfgGray}
\textbf{Backbone scale}
&
Approximately \(2.0\)B encoder parameters.
&
Approximately \(1.0\)B encoder parameters.
&
Parameter counts exclude the momentum copy when reporting trainable
parameters because the momentum encoder receives no gradient.
\\

\textbf{Encoder architecture}
&
Embedding dimension \(d=1664\); \(48\) transformer blocks;
\(16\) attention heads; MLP ratio \(4\).
&
Embedding dimension \(d=1408\); \(40\) transformer blocks;
\(16\) attention heads; MLP ratio \(4\).
&
Both models use pre-norm transformer blocks, patch size \(16\), and
rotary positional encoding.
\\

\rowcolor{CfgGray}
\textbf{Video tokenizer}
&
Patch size \(16\times16\); tubelet size \(2\).
&
Patch size \(16\times16\); tubelet size \(2\).
&
No backbone-specific spatial or temporal interpolation is introduced
during the controlled comparison.
\\

\textbf{Model input}
&
\(384\times384\) RGB; \(16\) sampled frames.
&
\(384\times384\) RGB; \(16\) sampled frames.
&
All clips undergo the same resize, center-preserving crop, privacy
filter, temporal sampling, and normalization.
\\

\rowcolor{CfgGray}
\textbf{Temporal sampling}
&
\(4\) fps; \(4.0\)-s sampled window.
&
\(4\) fps; \(4.0\)-s sampled window.
&
A window is sampled entirely inside one shot. No sample crosses a shot
or source-video boundary.
\\

\textbf{Context and target}
&
Frames \(1{:}12\) form context; frames \(13{:}16\) form the future
target; horizon \(1.0\) s.
&
Frames \(1{:}12\) form context; frames \(13{:}16\) form the future
target; horizon \(1.0\) s.
&
Context duration is \(3.0\) s. Temporal indices are identical across
all adaptation methods within each replication.
\\

\rowcolor{CfgGray}
\textbf{Pixel normalization}
&
ImageNet mean
\((0.485,0.456,0.406)\) and standard deviation
\((0.229,0.224,0.225)\).
&
Same.
&
Normalization is applied after privacy filtering and spatial
resampling.
\\

\textbf{DINOv2 checkpoint}
&
\multicolumn{2}{
>{\RaggedRight\arraybackslash}p{0.50\textwidth}
}{
\texttt{dinov2\_vitg14\_reg}; patch size \(14\); descriptor dimension
\(1536\); final normalized patch-token representation.
}
&
The teacher is frozen and evaluated in BF16 without stochastic
augmentation.
\\

\rowcolor{CfgGray}
\textbf{Teacher input}
&
\multicolumn{2}{
>{\RaggedRight\arraybackslash}p{0.50\textwidth}
}{
\(518\times518\) RGB; ImageNet normalization; the same \(16\) temporal
indices used by the JEPA branch.
}
&
Teacher preprocessing is deterministic and partition independent.
\\

\textbf{Target-cache precision}
&
\multicolumn{2}{
>{\RaggedRight\arraybackslash}p{0.50\textwidth}
}{
DINOv2 descriptors and factor targets stored in FP16; reliabilities
stored in FP32.
}
&
Targets are computed once before training. No gradient enters DINOv2,
the structural operators, cached targets, or reliability weights.
\\

\rowcolor{CfgGray}
\textbf{Region retention}
&
\multicolumn{2}{
>{\RaggedRight\arraybackslash}p{0.50\textwidth}
}{
Minimum confidence \(0.55\); minimum area \(0.1\%\) of valid image
support; duplicate suppression at mask IoU \(0.70\).
}
&
Thresholds are fixed on training/validation data and reused without
modification on test data.
\\

\textbf{Temporal association}
&
\multicolumn{2}{
>{\RaggedRight\arraybackslash}p{0.50\textwidth}
}{
Hungarian matching with
\(0.45\) box-IoU cost,
\(0.35\) mask-IoU cost, and
\(0.20\) descriptor-cosine cost;
maximum gap \(4\) frames;
minimum track length \(3\) frames.
}
&
Association is restricted to a single source video and shot.
Unmatched regions are retained only after satisfying the track
initiation criterion.
\\

\rowcolor{CfgGray}
\textbf{Reliability range}
&
\multicolumn{2}{
>{\RaggedRight\arraybackslash}p{0.50\textwidth}
}{
\(\omega_{n,k}\in[0.10,1.00]\), computed from region confidence,
temporal support, track continuity, and factor-specific validity.
}
&
Reliability weights are stop-gradient quantities and are normalized
within factor before minibatch aggregation.
\\

% =====================================================================
% PANEL B
% =====================================================================
\rowcolor{CfgCyan}
\multicolumn{4}{l}{
\textcolor{CfgInk}{
\textbf{Panel B: Masking and predictor architecture}
}}
\\

\textbf{Small-block masks}
&
\multicolumn{2}{
>{\RaggedRight\arraybackslash}p{0.50\textwidth}
}{
\(8\) blocks per clip; spatial scale \(0.15\); aspect-ratio range
\([0.75,1.50]\); temporal scale \(1.0\).
}
&
The same mask realization is reused across methods for a given
replication and sampled clip.
\\

\rowcolor{CfgGray}
\textbf{Large-block masks}
&
\multicolumn{2}{
>{\RaggedRight\arraybackslash}p{0.50\textwidth}
}{
\(2\) blocks per clip; spatial scale \(0.70\); aspect-ratio range
\([0.75,1.50]\); temporal scale \(1.0\).
}
&
Small- and large-block policies are jointly applied. Complement masks
are not forced.
\\

\textbf{Predictor backbone}
&
\multicolumn{2}{
>{\RaggedRight\arraybackslash}p{0.50\textwidth}
}{
\(24\) transformer layers; width \(384\); \(12\) heads;
MLP ratio \(4\); RoPE; learned mask tokens; no predictor registers.
}
&
The monolithic and factorized variants use the same predictor depth,
width, attention count, positional encoding, and target-query support.
\\

\rowcolor{CfgGray}
\textbf{Factor target dimensions}
&
\multicolumn{2}{
>{\RaggedRight\arraybackslash}p{0.50\textwidth}
}{
\(p_L=256\), \(p_A=256\), \(p_I=256\);
one scalar visibility target per retained agent.
}
&
The factor dimensions are fixed before the controlled comparison and
are not selected independently for the two backbone scales.
\\

\textbf{Factor ranks}
&
\multicolumn{2}{
>{\RaggedRight\arraybackslash}p{0.50\textwidth}
}{
Layout rank \(r_L=64\);
agent rank \(r_A=96\);
interaction rank \(r_I=64\).
}
&
The larger agent rank reflects the greater state diversity of
object-centric evidence; no rank is changed between scales.
\\

\rowcolor{CfgGray}
\textbf{Visibility head}
&
Two-layer MLP:
\(384\rightarrow256\rightarrow1\);
GELU; sigmoid output.
&
Same.
&
Visibility targets and predictions lie in \([0,1]\). Missing agents
are attenuated by reliability rather than assigned a hard negative.
\\

\textbf{Interaction head}
&
Two-layer pair MLP:
\(768\rightarrow256\rightarrow1\);
GELU; sigmoid strength.
&
Same.
&
Candidate pairs are formed within normalized image-plane radius
\(0.25\). At most \(12\) nearest valid neighbors are retained per
agent.
\\

\rowcolor{CfgGray}
\textbf{Trainable encoder depth}
&
Top \(K_G=2\) online-encoder blocks.
&
Top \(K_g=1\) online-encoder block.
&
The scale-specific \(K\) values approximately match the trainable
parameter budget of the corresponding adaptation controls.
\\

\textbf{Frozen components}
&
Bottom \(46\) online blocks; complete momentum encoder; DINOv2
teacher; structural operators; cached targets.
&
Bottom \(39\) online blocks; complete momentum encoder; DINOv2
teacher; structural operators; cached targets.
&
The momentum encoder is updated only by EMA.
\\

% =====================================================================
% PANEL C
% =====================================================================
\rowcolor{CfgGold}
\multicolumn{4}{l}{
\textcolor{CfgInk}{
\textbf{Panel C: Optimization, regularization, and model selection}
}}
\\

\textbf{Optimizer}
&
AdamW,
\(\beta_1=0.9\),
\(\beta_2=0.95\),
\(\epsilon=10^{-8}\).
&
Same.
&
Optimizer state is retained only for trainable parameters.
\\

\rowcolor{CfgGray}
\textbf{Learning rates}
&
Factorized predictor and heads:
\(1.0\times10^{-4}\);
top encoder blocks:
\(1.0\times10^{-5}\).
&
Same.
&
Encoder learning rate is \(0.1\times\) the predictor learning rate.
No layer-wise decay is applied within the selected top-\(K\) blocks.
\\

\textbf{Weight decay}
&
\(0.04\).
&
\(0.04\).
&
Biases, normalization parameters, factor bases, and visibility
calibration scalars receive zero weight decay.
\\

\rowcolor{CfgGray}
\textbf{Schedule}
&
Linear warm-up for \(1{,}000\) steps, followed by cosine decay to
\(1.0\times10^{-6}\).
&
Same.
&
Schedule is indexed by optimizer updates rather than processed clips.
\\

\textbf{Training duration}
&
\(20{,}000\) optimizer updates.
&
\(20{,}000\) optimizer updates.
&
Every method receives the same number of updates and sampled raw
clips.
\\

\rowcolor{CfgGray}
\textbf{Per-device batch}
&
\(2\) clips/GPU with gradient accumulation \(8\).
&
\(4\) clips/GPU with gradient accumulation \(4\).
&
Both scales use \(8\) GPUs, producing the same global batch of
\(128\) clips.
\\

\textbf{Numerical precision}
&
BF16 parameters and activations; FP32 optimizer states and loss
accumulation.
&
Same.
&
Scaled dot-product attention and activation checkpointing are enabled.
\\

\rowcolor{CfgGray}
\textbf{Gradient clipping}
&
Global norm clipped at \(1.0\).
&
Same.
&
Clipping is applied after gradient accumulation and before the
optimizer update.
\\

\textbf{Dropout}
&
Attention dropout \(0\); MLP dropout \(0\); stochastic depth \(0\).
&
Same.
&
Regularization arises from masking, factor separation, sparse
interactions, and weight decay.
\\

\rowcolor{CfgGray}
\textbf{EMA momentum}
&
Constant \(\mu=0.99925\).
&
Same.
&
After each optimizer update,
\(\bar\theta\leftarrow
\mu\bar\theta+(1-\mu)\theta\).
\\

\textbf{JEPA coefficient}
&
\(\lambda_{\mathrm{JEPA}}=1.00\).
&
Same.
&
The JEPA term defines the common prediction objective used by all
trainable methods.
\\

\rowcolor{CfgGray}
\textbf{Factor coefficient}
&
\(\lambda_{\mathrm{sup}}=1.00\).
&
Same.
&
Set to zero for FactorJEPA-RAW and all generic adaptation baselines.
\\

\textbf{Visibility coefficient}
&
\(\lambda_V=0.25\).
&
Same.
&
Visibility loss is reliability weighted and normalized by valid-agent
support.
\\

\rowcolor{CfgGray}
\textbf{Separation coefficient}
&
\(\lambda_{\mathrm{sep}}=0.05\).
&
Same.
&
Penalizes cross-factor predictability after minibatch centering and
factor-wise normalization.
\\

\textbf{Sparsity coefficient}
&
\(\lambda_{\mathrm{sparse}}=0.01\).
&
Same.
&
Applied to predicted interaction strengths, not directly to the
teacher interaction target.
\\

\rowcolor{CfgGray}
\textbf{Nonlinearity coefficient}
&
\(\beta_{\mathrm{nlin}}=0.10\).
&
Same.
&
Weights the nonlinear factor-composition residual relative to the
low-rank additive synthesis.
\\

\textbf{Validation frequency}
&
Every \(500\) optimizer updates.
&
Same.
&
Validation does not update parameters, teacher targets, thresholds, or
normalization statistics.
\\

\rowcolor{CfgGray}
\textbf{Checkpoint selection}
&
Lowest validation Future-frame L1; ties resolved using validation
Mask-ratio slope.
&
Same.
&
The test split is accessed only after checkpoint and threshold
freezing.
\\

\textbf{Replication}
&
Three independent runs with seeds
\(\{17,29,43\}\).
&
Same.
&
Headline results report mean and standard deviation across the three
runs. Clustered confidence intervals operate above the clip level.
\\

% =====================================================================
% PANEL D
% =====================================================================
\rowcolor{CfgRose}
\multicolumn{4}{l}{
\textcolor{CfgInk}{
\textbf{Panel D: Method-specific adaptation configuration}
}}
\\

\textbf{Frozen V-JEPA}
&
No gradient updates; \(0\) trainable parameters.
&
No gradient updates; \(0\) trainable parameters.
&
Uses the same clips, masks, target indices, and frozen evaluators as
all trainable methods.
\\

\rowcolor{CfgGray}
\textbf{Full fine-tuning}
&
Online encoder and monolithic predictor;
approximately \(2.04\)B trainable parameters.
&
Online encoder and monolithic predictor;
approximately \(1.04\)B trainable parameters.
&
Not parameter matched. Momentum encoder remains stop-gradient and is
updated by EMA.
\\

\textbf{LoRA}
&
Rank \(80\), scale \(\alpha=160\), dropout \(0.05\);
approximately \(114\)M trainable parameters.
&
Rank \(64\), scale \(\alpha=128\), dropout \(0.05\);
approximately \(68\)M trainable parameters.
&
Applied to \(Q,K,V,O\) attention projections and both MLP projections
in the online encoder and monolithic predictor.
\\

\rowcolor{CfgBlue}
\textbf{DoRA}
&
Rank \(80\), \(\alpha=160\), dropout \(0.05\);
approximately \(116\)M trainable parameters.
&
Rank \(64\), \(\alpha=128\), dropout \(0.05\);
approximately \(69\)M trainable parameters.
&
Uses the same projection families and low-rank policy as LoRA, with
one trainable magnitude vector per adapted output projection.
\\

\textbf{Auto-RGN}
&
Monolithic predictor and two selected encoder blocks;
approximately \(111\)M trainable parameters.
&
Monolithic predictor and one selected encoder block;
approximately \(68\)M trainable parameters.
&
Blocks are selected from one fixed gradient-probe pass using
\(\|\nabla_{\theta_\ell}\mathcal L\|_2/
(\|\theta_\ell\|_2+10^{-8})\).
\\

\rowcolor{CfgGold}
\textbf{FactorJEPA-RAW}
&
Factorized predictor and top two encoder blocks;
approximately \(115\)M trainable parameters.
&
Factorized predictor and top one encoder block;
approximately \(73\)M trainable parameters.
&
Uses only \(\mathcal L_{\mathrm{JEPA}}\),
\(\mathcal L_{\mathrm{sep}}\), and
\(\mathcal L_{\mathrm{sparse}}\);
no DINOv2-derived factor targets are consumed.
\\

\rowcolor{CfgGreen}
\textbf{FactorJEPA}
&
Factorized predictor and top two encoder blocks;
approximately \(115\)M trainable parameters.
&
Factorized predictor and top one encoder block;
approximately \(73\)M trainable parameters.
&
Uses the complete objective:
\(\mathcal L_{\mathrm{JEPA}}+
\lambda_{\mathrm{sup}}\mathcal L_{\mathrm{factor}}+
\lambda_V\mathcal L_V+
\lambda_{\mathrm{sep}}\mathcal L_{\mathrm{sep}}+
\lambda_{\mathrm{sparse}}\mathcal L_{\mathrm{sparse}}\).
\\

\textbf{Budget tolerance}
&
LoRA, DoRA, Auto-RGN, and FactorJEPA remain within \(5\%\) of the
target trainable-parameter budget after exact implementation-level
counting.
&
Same.
&
If exact instantiated counts exceed the tolerance, LoRA/DoRA rank or
the Auto-RGN block budget must be adjusted before training.
\\

% =====================================================================
% PANEL E
% =====================================================================
\rowcolor{CfgGreen}
\multicolumn{4}{l}{
\textcolor{CfgInk}{
\textbf{Panel E: Resource envelope and profiling contract}
}}
\\

\textbf{Training hardware}
&
\(8\times\) NVIDIA H100 SXM, \(80\) GiB per GPU.
&
Same.
&
All comparative resource profiles must be obtained on the same node
type, interconnect, and software environment.
\\

\rowcolor{CfgGray}
\textbf{Software environment}
&
PyTorch \(2.6.0\); CUDA \(12.4\); cuDNN \(9.1\);
BF16; SDPA enabled.
&
Same.
&
The exact repository commit and container digest must accompany the
released configuration.
\\

\textbf{Teacher-cache envelope}
&
At most \(42\) GPU-hours; peak allocation \(48\) GiB/GPU.
&
The same cache is reused; no second scale-specific teacher pass.
&
Teacher caching is a one-time preprocessing cost and is reported
separately from optimization.
\\

\rowcolor{CfgGray}
\textbf{Frozen evaluation envelope}
&
No training GPU-hours; inference allocation at most \(38\) GiB/GPU.
&
No training GPU-hours; inference allocation at most \(30\) GiB/GPU.
&
Zero training cost does not imply zero evaluation cost.
\\

\textbf{Full fine-tuning envelope}
&
At most \(160\) GPU-hours;
peak allocation at most \(78\) GiB/GPU.
&
At most \(96\) GPU-hours;
peak allocation at most \(64\) GiB/GPU.
&
Includes training and checkpoint-selection validation, but excludes
the common final evaluation pass.
\\

\rowcolor{CfgBlue}
\textbf{LoRA envelope}
&
At most \(96\) GPU-hours;
peak allocation at most \(54\) GiB/GPU.
&
At most \(56\) GPU-hours;
peak allocation at most \(44\) GiB/GPU.
&
Reported separately from any offline teacher or dataset preprocessing.
\\

\textbf{DoRA envelope}
&
At most \(100\) GPU-hours;
peak allocation at most \(56\) GiB/GPU.
&
At most \(60\) GPU-hours;
peak allocation at most \(46\) GiB/GPU.
&
Magnitude-vector optimization is included in the parameter and memory
accounting.
\\

\rowcolor{CfgBlue}
\textbf{Auto-RGN envelope}
&
At most \(104\) GPU-hours;
peak allocation at most \(60\) GiB/GPU.
&
At most \(60\) GPU-hours;
peak allocation at most \(48\) GiB/GPU.
&
Includes the fixed gradient-probe pass used for block selection.
\\

\textbf{FactorJEPA-RAW envelope}
&
At most \(112\) GPU-hours;
peak allocation at most \(62\) GiB/GPU.
&
At most \(64\) GPU-hours;
peak allocation at most \(50\) GiB/GPU.
&
No offline DINOv2 target-cache cost is attributed to this control.
\\

\rowcolor{CfgGreen}
\textbf{FactorJEPA envelope}
&
At most \(120\) optimization GPU-hours;
peak allocation at most \(64\) GiB/GPU.
&
At most \(70\) optimization GPU-hours;
peak allocation at most \(52\) GiB/GPU.
&
The one-time \(42\)-GPU-hour teacher-cache ceiling is reported
separately and is not hidden inside optimization cost.
\\

\textbf{GPU-hour definition}
&
\multicolumn{2}{
>{\RaggedRight\arraybackslash}p{0.50\textwidth}
}{
\(
H_{\mathrm{GPU}}
=
\sum_r
n_{\mathrm{GPU}}^{(r)}
\Delta t_r
\),
where \(\Delta t_r\) is the wall-clock duration of run \(r\).
}
&
Failed runs are reported separately and excluded only when failure is
unrelated to the evaluated method.
\\

\rowcolor{CfgGray}
\textbf{Peak-memory definition}
&
\multicolumn{2}{
>{\RaggedRight\arraybackslash}p{0.50\textwidth}
}{
Maximum value returned by
\texttt{torch.cuda.max\_memory\_allocated()} after initialization and
five warm-up updates.
}
&
Measured using the reported per-device batch, precision, and gradient
accumulation.
\\

\textbf{Latency protocol}
&
Batch \(1\); \(50\) warm-up iterations; \(500\) synchronized timed
iterations.
&
Same.
&
Report median and \(95\)th-percentile ms/clip. Model-only latency uses
resident tensors; end-to-end latency additionally includes decode,
resize, normalization, and host-to-device transfer.
\\

\rowcolor{CfgGray}
\textbf{Latency ceiling}
&
Model-only median at most \(230\) ms/clip.
&
Model-only median at most \(160\) ms/clip.
&
These are execution ceilings, not measured results. Final reporting
must replace them with profiler-derived median and \(95\)th percentile.
\\

\textbf{Resource aggregation}
&
Mean and standard deviation across the three training replications.
&
Same.
&
GPU-hours are summed per run; peak memory is the maximum per run;
latency is summarized from the synchronized inference distribution.
\\

\end{longtable}

\arrayrulecolor{black}
\endgroup

% clearpage removed for continuous flow
\twocolumn

% Suggested preamble additions:
% \usepackage[table]{xcolor}
% \usepackage{tabularx}
% \usepackage{makecell}
% \usepackage{booktabs}

\definecolor{AttrBlue}{HTML}{EAF2FB}
\definecolor{AttrTeal}{HTML}{E7F5F2}
\definecolor{AttrGold}{HTML}{FFF3D2}
\definecolor{AttrGray}{HTML}{F2F3F5}
\definecolor{AttrRose}{HTML}{FBECEC}
\definecolor{AttrPending}{HTML}{992F2F}

% Replace with exported mean [paired 95% CI].
\newcommand{\Score}{
\textcolor{AttrPending}{\textbf{[mean; 95\% CI]}}
}

% =====================================================================
\section{Attribution and Executed Component Ablations}
\label{app:attribution}

The main results evaluate FactorJEPA as a complete predictive system.
This appendix uses the executed comparison family to distinguish three
sources of performance: \textbf{matched monolithic adaptation}, the
\textbf{factorized structural package}, and
\textbf{DINOv2-derived structured supervision}. We then evaluate the
executed interaction-objective and encoder-adaptation variants.

The available experiments do not form a complete
architecture--supervision factorial. We therefore restrict the
attribution language to contrasts directly supported by the executed
runs. In particular, when FactorJEPA-RAW includes separation and
interaction-sparsity regularization, its comparison with a monolithic
predictor measures the contribution of the
\emph{factorized structural package}, not predictor architecture in
isolation.

% =====================================================================
\subsection{Executed Comparison Contract}
\label{app:attribution_contract}

Table~\ref{tab:executed_contract} defines the training intervention
associated with each primary comparison. All trainable methods use the
same raw clips, context and target masks, target indices, optimizer,
training duration, validation rule, and frozen evaluation protocol.
The momentum target encoder remains stop-gradient and follows the same
EMA update schedule.

We use \textbf{Auto-RGN} as the primary matched monolithic reference
because it updates the monolithic predictor and the same number of
gradient-selected online-encoder blocks while remaining close to
FactorJEPA's trainable-parameter budget. LoRA and DoRA provide
additional parameter-efficient controls. Full fine-tuning is retained
only as a non-matched capacity ceiling.

\begin{table*}[t]
\centering
\caption{
\textbf{Executed attribution contract.}
FactorJEPA-RAW and FactorJEPA share the factorized predictor,
separation objective, interaction-sparsity objective, trainable encoder
scope, and parameter budget. They differ in access to the
DINOv2-derived factor and visibility targets. Auto-RGN is the primary
parameter-matched monolithic reference; full fine-tuning is a
non-matched capacity ceiling.
}
\label{tab:executed_contract}
\scriptsize
\setlength{\tabcolsep}{4pt}
\renewcommand{\arraystretch}{1.15}

\begin{tabularx}{\textwidth}{
    p{0.15\textwidth}
    p{0.12\textwidth}
    p{0.13\textwidth}
    p{0.26\textwidth}
    p{0.15\textwidth}
    X
}
\toprule
\textbf{Method}
&
\textbf{Predictor}
&
\textbf{Factor targets}
&
\textbf{Executed objective}
&
\textbf{Trainable scope}
&
\textbf{Attribution role}
\\
\midrule

\rowcolor{AttrGray}
\textbf{Frozen V-JEPA}
&
Monolithic
&
None
&
No optimization
&
None
&
No-adaptation reference.
\\

\textbf{V-JEPA Full-FT}
&
Monolithic
&
None
&
\(\mathcal L_{\mathrm{JEPA}}\)
&
Complete online encoder and predictor
&
Non-matched capacity ceiling.
\\

\textbf{V-JEPA LoRA}
&
Monolithic
&
None
&
\(\mathcal L_{\mathrm{JEPA}}\)
&
Matched low-rank projection updates
&
Parameter-efficient adaptation control.
\\

\textbf{V-JEPA DoRA}
&
Monolithic
&
None
&
\(\mathcal L_{\mathrm{JEPA}}\)
&
Matched decomposed low-rank updates
&
Parameter-efficient adaptation control.
\\

\rowcolor{AttrBlue}
\textbf{V-JEPA Auto-RGN}
&
Monolithic
&
None
&
\(\mathcal L_{\mathrm{JEPA}}\)
&
Monolithic predictor and selected top-\(K\) blocks
&
Primary matched monolithic reference.
\\

\rowcolor{AttrGold}
\textbf{FactorJEPA-RAW}
&
Factorized
&
None
&
\(\mathcal L_{\mathrm{JEPA}}
+\lambda_{\mathrm{sep}}\mathcal L_{\mathrm{sep}}
+\lambda_{\mathrm{sparse}}\mathcal L_{\mathrm{sparse}}\)
&
Factorized predictor and selected top-\(K\) blocks
&
Factorized structural package without teacher targets.
\\

\rowcolor{AttrTeal}
\textbf{FactorJEPA}
&
Factorized
&
\(T_L,T_A,T_V,T_I\)
&
\(\mathcal L_{\mathrm{JEPA}}
+\lambda_{\mathrm{sup}}\mathcal L_{\mathrm{factor}}
+\lambda_V\mathcal L_V
+\lambda_{\mathrm{sep}}\mathcal L_{\mathrm{sep}}
+\lambda_{\mathrm{sparse}}\mathcal L_{\mathrm{sparse}}\)
&
Factorized predictor and selected top-\(K\) blocks
&
Complete model: structural package plus structured supervision.
\\

\bottomrule
\end{tabularx}

\vspace{2pt}
\parbox{0.97\textwidth}{
\textit{Parameter matching.}
At ViT-G, Auto-RGN and FactorJEPA contain approximately \(111\)M
and \(115\)M trainable parameters, respectively; at ViT-g, they contain
approximately \(68\)M and \(73\)M. Full fine-tuning contains
approximately \(2.04\)B and \(1.04\)B trainable parameters and is
therefore not used for component attribution.
}
\end{table*}

All structured targets are produced automatically by the frozen
teacher pipeline. No manually annotated factor labels enter any
training condition. The headline evaluators remain frozen and do not
reuse DINOv2-derived targets as evaluation labels.

% =====================================================================
\subsection{Factorized Structural Package and Structured Targets}
\label{app:attribution_primary}

The primary attribution chain is

\begin{equation*}
\resizebox{0.97\columnwidth}{!}{$\displaystyle
\underbrace{\text{Auto-RGN}}_{\substack{\text{matched monolithic}\\
\text{adaptation}}}
\;\longrightarrow\;
\underbrace{\text{FactorJEPA-RAW}}_{\substack{\text{factorized predictor}\\
+\ \text{structural regularization}}}
\;\longrightarrow\;
\underbrace{\text{FactorJEPA}}_{\substack{\text{factorized package}\\
+\ \text{structured targets}}}
$}
\end{equation*}

For diagnostic \(m\), let \(Y_M^{(m)}\) denote the score of method
\(M\). We direction-normalize the metrics as

\begin{equation*}
\widetilde Y_M^{(m)}
=
\begin{cases}
-Y_M^{(m)}, & \text{if lower is better},\\
\phantom{-}Y_M^{(m)}, & \text{if higher is better}.
\end{cases}
\end{equation*}

We report three prespecified effects:

\begin{equation*}
\begin{aligned}
\Delta_{\mathrm{pkg}}
&=
\widetilde Y_{\mathrm{FactorJEPA\text{-}RAW}}
-
\widetilde Y_{\mathrm{Auto\text{-}RGN}},
\\[2pt]
\Delta_{\mathrm{sup}\mid\mathrm{pkg}}
&=
\widetilde Y_{\mathrm{FactorJEPA}}
-
\widetilde Y_{\mathrm{FactorJEPA\text{-}RAW}},
\\[2pt]
\Delta_{\mathrm{total}}
&=
\widetilde Y_{\mathrm{FactorJEPA}}
-
\widetilde Y_{\mathrm{Auto\text{-}RGN}}.
\end{aligned}
\end{equation*}

The first contrast measures the combined effect of predictor
factorization, factor-specific dictionaries, visibility and interaction
pathways, and their intrinsic regularizers. It is therefore a
\textbf{package effect}, not a pure architecture effect.

The second contrast is more specific. FactorJEPA and FactorJEPA-RAW
share the predictor, synthesis dictionaries, separation and sparsity
terms, encoder scope, and optimization protocol. Their difference
therefore isolates the incremental contribution of
\(\mathcal L_{\mathrm{factor}}\) and \(\mathcal L_V\), conditional on
the shared factorized structural package.

\begin{figure*}[p]
\centering
\includegraphics[
    width=0.98\textwidth,
    trim=0bp 1520bp 0bp 0bp,
    clip
]{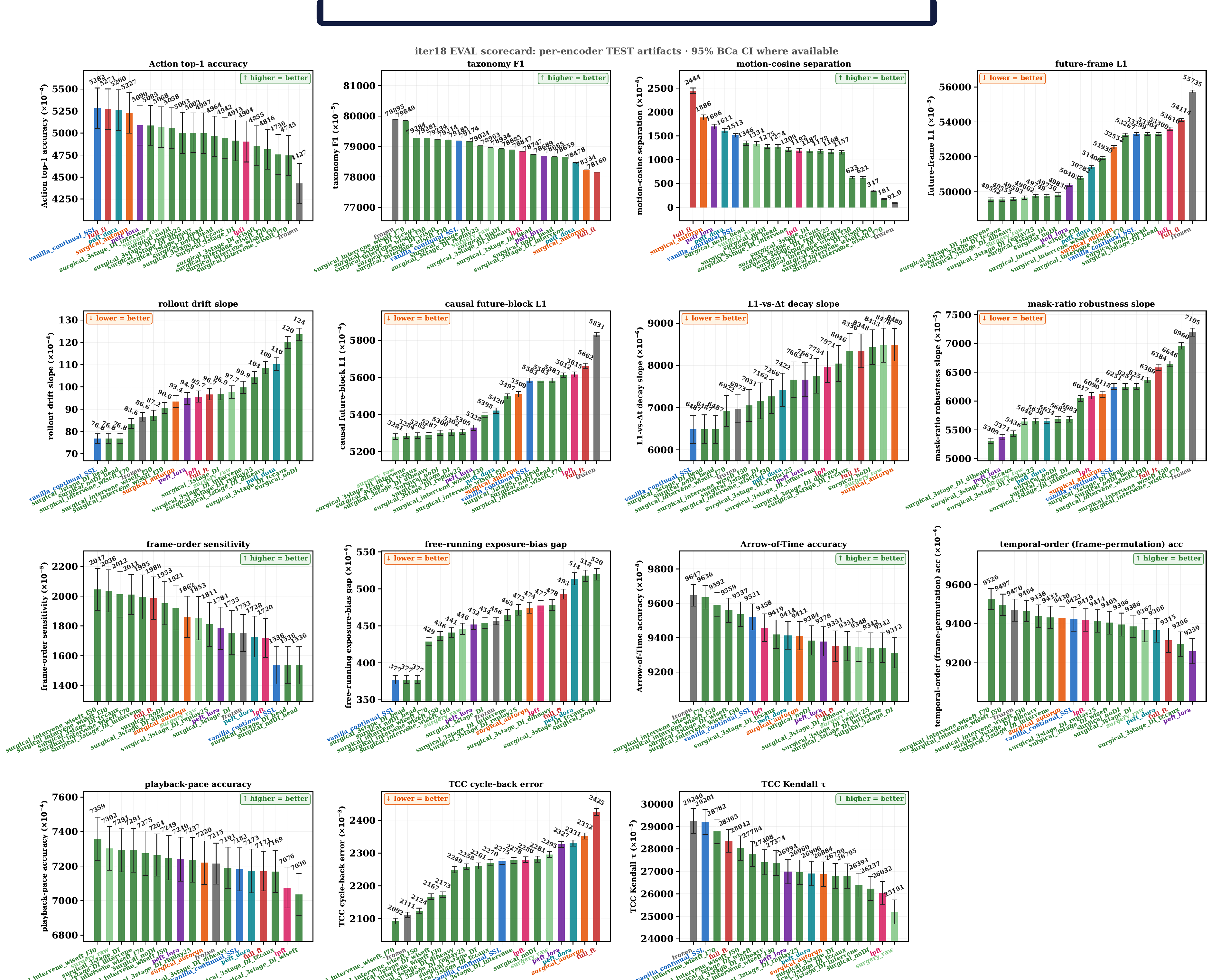}
\caption{
\textbf{Complete ViT-G attribution and ablation scorecard.}
The comparison includes frozen and continually adapted V-JEPA,
full and parameter-efficient fine-tuning, Auto-RGN,
FactorJEPA-RAW, the executed FactorJEPA objective variants, and the
WiseFT encoder-scope variants. Bars report held-out means; error bars
denote \(95\%\) BCa confidence intervals where available. Each panel
states whether higher or lower values are preferred. Internal variants
are mapped to their publication-facing definitions in
Table~\ref{tab:executed_contract}. Attribution claims use the
prespecified contrasts in
Figure~\ref{fig:ablation_scorecard_vitg}, rather than selecting the
best method separately for each diagnostic.
}
\label{fig:ablation_scorecard_vitg}
\end{figure*}

Figure~\ref{fig:ablation_scorecard_vitg} reports the absolute
performance and paired contrasts.

% ViT-G: upper half of the combined PDF

Confidence intervals are estimated using a
\textbf{paired hierarchical bootstrap}. Each replicate applies the same
resampled seeds, cities, source videos, and clips to all methods in a
contrast. Cities are resampled first and source videos are resampled
within city; all clips from a sampled source video remain together.
This pairing removes variation shared by the compared methods and
preserves the geographic and temporal dependence of the evaluation
set.

Causal L1 is treated as an operational measure of consistency under
the frozen evaluator-side intervention protocol, not as evidence of
causal identification. Its intervention generator is fixed before
model comparison, consumes neither DINOv2 factor targets nor manual
factor annotations, and is applied identically to every method.

% =====================================================================
\subsection{Factor-Objective and Intervention Variants}
\label{app:attribution_factor_variants}

The primary-scale ablations contain three configurations of the
factorized model:

\begin{itemize}
    \item \textbf{FactorJEPA-3S} uses the executed three-stage
    layout--agent--interaction curriculum and serves as the base
    factorized configuration.

    \item \textbf{FactorJEPA-INT} uses the same backbone, predictor,
    factor targets, and encoder scope while enabling the executed
    intervention-oriented training configuration.

    \item \textbf{FactorJEPA-DI+} retains the three-stage architecture
    but increases the executed weighting or sampling emphasis assigned
    to the dynamic-interaction component.
\end{itemize}

These runs are treated as \textbf{configuration interventions}, not
one-factor pathway removals. In particular, FactorJEPA-DI+ tests
sensitivity to interaction emphasis; it does not establish that the
interaction pathway is necessary. Similarly, FactorJEPA-INT measures
the effect of the executed intervention-oriented configuration and
should not be interpreted as a general causal-supervision ablation.

Relative to FactorJEPA-3S, define

\begin{equation*}
\begin{aligned}
\Delta_{\mathrm{INT}}
&=
\widetilde Y_{\mathrm{FactorJEPA\text{-}INT}}
-
\widetilde Y_{\mathrm{FactorJEPA\text{-}3S}},
\\
\Delta_{\mathrm{DI+}}
&=
\widetilde Y_{\mathrm{FactorJEPA\text{-}DI+}}
-
\widetilde Y_{\mathrm{FactorJEPA\text{-}3S}}.
\end{aligned}
\end{equation*}

We evaluate these effects jointly across future prediction,
intervention consistency, masking robustness, and motion accessibility.
A favorable result on one diagnostic is not treated as a universal
improvement when accompanied by degradation on another. This is
particularly important for the observed prediction--motion trade-off.

The final manuscript-facing configuration is selected exclusively by
the prespecified validation rule. Test performance is not used to
choose among FactorJEPA-3S, FactorJEPA-INT, and FactorJEPA-DI+.

% =====================================================================
\subsection{Encoder-Adaptation and Resource Sensitivity}
\label{app:attribution_encoder_scope}

The remaining variants test whether the observed gains can be
explained by the amount of encoder adaptation rather than the
factorized predictor. V-JEPA LP-FT and full fine-tuning provide the
minimal and maximal monolithic adaptation endpoints. The three
FactorJEPA-WiseFT variants vary the executed encoder-scope parameter
while retaining the factorized predictor and structured objective.

Let \(b\in\{30,50,70\}\) denote the executed WiseFT scope setting and
let \(P_b\), \(H_b\), and \(M_b\) denote its trainable parameters,
measured GPU-hours, and peak allocated memory. For each setting, we
report

\begin{equation*}
\mathcal R_b
=
\left(
P_b,\,
H_b,\,
M_b,\,
Y_b^{\mathrm{future}},\,
Y_b^{\mathrm{causal}},\,
Y_b^{\mathrm{mask}},\,
Y_b^{\mathrm{motion}}
\right).
\end{equation*}

The reported setting names must be accompanied by their operational
meaning: the exact trainable blocks, frozen blocks, and trainable
parameter count. Internal labels such as \texttt{f30},
\texttt{f50}, and \texttt{f70} are not used without this mapping.

A configuration \(b\) is Pareto dominated when another configuration
\(b'\) satisfies

\begin{equation*}
\resizebox{0.98\columnwidth}{!}{$\displaystyle
P_{b'}\le P_b,
\qquad
H_{b'}\le H_b,
\qquad
\widetilde Y_{b'}^{(m)}
\ge
\widetilde Y_b^{(m)}
\quad
\forall\,m\in\mathcal M_{\mathrm{primary}}
$}
\end{equation*}

with at least one strict inequality. This analysis distinguishes gains
that persist under constrained encoder adaptation from gains obtained
solely through additional trainable capacity.

% =====================================================================
\subsection{Consolidated Attribution}
\label{app:attribution_summary}

Figure~\ref{fig:ablation_scorecard_vitg} consolidates the
prespecified paired contrasts. Each point reports an absolute,
direction-normalized effect; horizontal bars report paired
source-clustered \(95\%\) confidence intervals. Positive values favor
the named intervention.

The evidence is interpreted under the following claim boundaries:

\begin{itemize}
    \item A positive \(\Delta_{\mathrm{pkg}}\) supports the
    \textbf{factorized structural package}. It does not isolate
    predictor architecture from separation and sparsity regularization.

    \item A positive
    \(\Delta_{\mathrm{sup}\mid\mathrm{pkg}}\) supports the incremental
    contribution of the DINOv2-derived factor and visibility targets,
    conditional on the shared factorized architecture.

    \item A positive \(\Delta_{\mathrm{total}}\) establishes improvement
    over the matched monolithic Auto-RGN reference. Full fine-tuning is
    interpreted separately because it is not parameter matched.

    \item FactorJEPA-INT and FactorJEPA-DI+ establish sensitivity to the
    corresponding executed configuration changes. They do not isolate
    unexecuted pathway removals.

    \item WiseFT comparisons establish robustness to encoder-adaptation
    scope only when the performance trend is considered jointly with
    measured parameters, GPU-hours, and memory.

    \item Teacher-relative factor diagnostics may explain a mechanism,
    but they do not replace the frozen future-latent, intervention,
    masking, motion, or RGB evaluators.
\end{itemize}

Accordingly, the executed experiments identify two principal effects:
the advantage of the complete factorized structural package over
matched monolithic adaptation, and the incremental value of structured
teacher targets within that package. Finer architecture--supervision
interaction claims are left open because the corresponding
supervision-matched monolithic control was not executed.

% =====================================================================
\section{Metric Definitions, Clustered Inference, and Robustness}
\label{app:metrics_statistics}

This appendix specifies the evaluation contract underlying every
reported result. We distinguish \textbf{training targets} from
\textbf{evaluation signals}: DINOv2-derived layout, agent, visibility,
and interaction targets supervise FactorJEPA but are not reused as
labels for the headline metrics. Unless stated otherwise, every
evaluator is frozen before test evaluation and is applied identically
to all methods.

Metrics are first computed at their lowest valid evaluation unit and
then aggregated through the city--source-video hierarchy. Clips from
the same recording are not treated as independent observations.
Primary conclusions are restricted to Future-frame MSE, Intervention
L1, Mask-ratio slope, and Motion cosine. All remaining diagnostics are
treated as secondary or exploratory.

% =====================================================================
\subsection{Evaluation Contract and Metric Registry}
\label{app:metric_registry}

Table~\ref{tab:metric_registry} defines the complete metric registry.
The \emph{evaluation unit} identifies the lowest unit at which a score
is formed; statistical uncertainty is subsequently estimated above the
clip level.

\begin{table*}[t]
\centering
\caption{
\textbf{Evaluation metric registry.}
The table records the operational definition, preferred direction,
evaluation unit, frozen signal or evaluator, and inferential status of
every reported diagnostic. ``Intervention L1'' corresponds to the
quantity previously labeled Causal L1; the revised name reflects that
the metric measures intervention-response consistency rather than
causal identification.
}
\label{tab:metric_registry}
\scriptsize
\setlength{\tabcolsep}{3.5pt}
\renewcommand{\arraystretch}{1.13}

\begin{tabularx}{\textwidth}{
    p{0.16\textwidth}
    p{0.33\textwidth}
    p{0.07\textwidth}
    p{0.12\textwidth}
    p{0.19\textwidth}
    X
}
\toprule
\textbf{Metric}
&
\textbf{Operational definition}
&
\textbf{Dir.}
&
\textbf{Unit}
&
\textbf{Frozen signal or evaluator}
&
\textbf{Status}
\\
\midrule

\rowcolor{AttrTeal}
\multicolumn{6}{l}{\textbf{Primary predictive diagnostics}}
\\
\midrule

Future-frame MSE
&
Mean squared error between predicted and target future-token
embeddings.
&
\(\downarrow\)
&
Clip--horizon
&
Momentum target encoder
&
Primary
\\

Intervention L1
&
Normalized L1 discrepancy between predicted and target latent changes
under the same automatically generated intervention.
&
\(\downarrow\)
&
Clip--intervention
&
Intervention generator and target encoder
&
Primary
\\

Mask-ratio slope
&
OLS slope of Future-frame MSE as the visible context is progressively
reduced.
&
\(\downarrow\)
&
Clip--mask curve
&
Target encoder and fixed mask sampler
&
Primary
\\

Motion cosine
&
Cosine similarity between linearly decoded and target motion
descriptors.
&
\(\uparrow\)
&
Clip
&
Motion estimator and fixed-capacity probe
&
Primary
\\

\midrule
\rowcolor{AttrBlue}
\multicolumn{6}{l}{\textbf{Semantic diagnostics}}
\\
\midrule

Action top-1
&
Top-1 accuracy of a fixed-capacity action probe.
&
\(\uparrow\)
&
Clip
&
Frozen action labels and probe protocol
&
Secondary
\\

Taxonomy F1
&
F1 over the shared dynamic-agent taxonomy.
&
\(\uparrow\)
&
Clip--class
&
Frozen taxonomy evaluator
&
Secondary
\\

\midrule
\rowcolor{AttrGold}
\multicolumn{6}{l}{\textbf{Prediction-stability diagnostics}}
\\
\midrule

Rollout-drift slope
&
Increase in prediction error over autoregressive rollout depth.
&
\(\downarrow\)
&
Clip--rollout
&
Target encoder
&
Secondary
\\

L1-vs-\(\Delta t\) decay
&
Increase in latent error over the evaluated future horizons.
&
\(\downarrow\)
&
Clip--horizon
&
Target encoder
&
Secondary
\\

Exposure-bias gap
&
Difference between free-running and teacher-conditioned prediction
error.
&
\(\downarrow\)
&
Clip
&
Target encoder
&
Secondary
\\

\midrule
\rowcolor{AttrGray}
\multicolumn{6}{l}{\textbf{Temporal diagnostics}}
\\
\midrule

Frame-order sensitivity
&
Prediction-error increase after controlled frame-order corruption.
&
\(\uparrow\)
&
Clip
&
Fixed temporal permutation
&
Secondary
\\

Arrow-of-Time
&
Accuracy for distinguishing forward from reversed clips.
&
\(\uparrow\)
&
Clip
&
Fixed temporal classifier
&
Secondary
\\

Temporal-order accuracy
&
Accuracy for detecting frame permutations.
&
\(\uparrow\)
&
Clip
&
Fixed temporal-order classifier
&
Secondary
\\

Playback-pace accuracy
&
Accuracy for identifying the applied temporal-rate transformation.
&
\(\uparrow\)
&
Clip
&
Fixed pace classifier
&
Secondary
\\

TCC cycle-back error
&
Temporal distance between a source frame and its cycle-consistent
match.
&
\(\downarrow\)
&
Frame pair
&
Frozen correspondence features
&
Secondary
\\

TCC Kendall \(\tau\)
&
Rank agreement between predicted and true temporal correspondences.
&
\(\uparrow\)
&
Clip pair
&
Frozen correspondence features
&
Secondary
\\

\bottomrule
\end{tabularx}
\end{table*}

% =====================================================================
\subsection{Primary Predictive Diagnostics}
\label{app:primary_metric_definitions}

\paragraph{Future-frame MSE.}
For clip \(n\), horizon \(h\in\mathcal H\), and target token
\(p\in\{1,\ldots,m_h\}\), let
\(\widehat Y_{n,h,p}\in\mathbb R^d\) denote the predicted future token
and \(Y^\star_{n,h,p}\) the stop-gradient target-encoder token. The
clip-level error is

\begin{equation*}
e_n^{\mathrm{future}}
=
\frac{1}{|\mathcal H|}
\sum_{h\in\mathcal H}
\frac{1}{m_h d}
\sum_{p=1}^{m_h}
\left\|
\widehat Y_{n,h,p}
-
Y^\star_{n,h,p}
\right\|_2^2 .
\end{equation*}

All valid target tokens receive equal weight. Target-token
normalization, target horizons, and masking are fixed across methods.
The executed scorecard reports this MSE quantity; it is not relabeled
as L1. Horizon-specific errors are reported as sensitivity analyses and
do not replace the pooled primary score.

\paragraph{Intervention L1.}
Let \(\mathcal G_{\mathrm{int}}\) be the frozen automatic
evaluator-side intervention generator and let

\begin{equation*}
\mathcal A_n
=
\mathcal G_{\mathrm{int}}(x_n)
\end{equation*}

denote its retained intervention set for clip \(n\). The generator,
intervention types, spatial and temporal support rules, and exclusion
criteria are frozen before model comparison. It consumes neither
DINOv2 factor targets nor manually annotated factor labels.

For intervention \(a\in\mathcal A_n\), apply the same transformation
\(I_a\) to the evaluated clip and define

\begin{equation*}
\begin{aligned}
\Delta\widehat Y_{n,a}
&=
\widehat Y\!\left(I_a(x_n)\right)
-
\widehat Y(x_n),
\\
\Delta Y^\star_{n,a}
&=
\operatorname{sg}
\left[
Y^\star\!\left(I_a(x_n)\right)
-
Y^\star(x_n)
\right].
\end{aligned}
\end{equation*}

The normalized intervention error is

\begin{equation*}
e_{n,a}^{\mathrm{int}}
=
\frac{
\left\|
\Delta\widehat Y_{n,a}
-
\Delta Y^\star_{n,a}
\right\|_1
}{
\epsilon_{\mathrm{int}}
+
\left\|
\Delta Y^\star_{n,a}
\right\|_1
}.
\end{equation*}

The clip score averages over retained interventions:

\begin{equation*}
e_n^{\mathrm{int}}
=
\frac{1}{\max\{1,|\mathcal A_n|\}}
\sum_{a\in\mathcal A_n}
e_{n,a}^{\mathrm{int}} .
\end{equation*}

We report the numerator, target-effect denominator, retained
intervention count, and near-zero-effect rate alongside the normalized
score. The primary specification uses a fixed
\(\epsilon_{\mathrm{int}}\). Alternative constants and minimum-effect
filters are evaluated only as sensitivity checks. This metric measures
consistency with intervention-induced latent changes; it does not
establish causal identification.

\paragraph{Mask-ratio slope.}
Let \(\mathcal R=\{r_1,\ldots,r_J\}\) be the prespecified context-mask
ratios. For each ratio \(r\), we sample \(K\) masks from a fixed sampler
and reuse the same realizations across methods. Define

\begin{equation*}
e_n(r)
=
\frac{1}{K}
\sum_{k=1}^{K}
e_n^{\mathrm{future}}
\left(
M_{r,k}\odot x_n
\right).
\end{equation*}

The primary robustness statistic is the absolute OLS slope

\begin{equation*}
\beta_n^{\mathrm{mask}}
=
\frac{
\sum_{r\in\mathcal R}
(r-\bar r)
\left(
e_n(r)-\bar e_n
\right)
}{
\sum_{r\in\mathcal R}
(r-\bar r)^2
},
\end{equation*}

where \(\bar e_n\) is the mean error over the evaluated ratios. Smaller
values indicate slower degradation as context evidence is removed. We
also report the complete masking curve,

\begin{equation*}
\mathcal C_n^{\mathrm{mask}}
=
\left\{
\bigl(r,e_n(r)\bigr):
r\in\mathcal R
\right\},
\end{equation*}

together with relative slope and masking AUC as secondary robustness
summaries.

\paragraph{Motion cosine.}
Let \(E_{\mathrm{mot}}\) be the independently frozen motion estimator
and

\begin{equation*}
u_n^\star
=
E_{\mathrm{mot}}
\left(
x_{n,t:t+\Delta}
\right)
\in\mathbb R^{d_{\mathrm{mot}}}
\end{equation*}

its target descriptor. For every world model \(M\), a fixed-capacity
linear probe \(W_M\) is fitted on the training split:

\begin{equation*}
W_M
=
\arg\min_W
\sum_{n\in\mathcal D_{\mathrm{train}}}
\left\|
W\,\operatorname{pool}
\left(
\widehat Y_n^{M}
\right)
-
u_n^\star
\right\|_2^2
+
\lambda_{\mathrm{mot}}\|W\|_F^2 .
\end{equation*}

The regularization coefficient is selected on the validation split
under the same grid for every method. The test descriptor is

\begin{equation*}
\widehat u_n^M
=
W_M
\operatorname{pool}
\left(
\widehat Y_n^M
\right),
\end{equation*}

and Motion cosine is

\begin{equation*}
s_n^{\mathrm{motion}}
=
\frac{
\left\langle
\widehat u_n^M,u_n^\star
\right\rangle
}{
\left\|\widehat u_n^M\right\|_2
\left\|u_n^\star\right\|_2
+
\epsilon_{\mathrm{mot}}
}.
\end{equation*}

The motion estimator and probe protocol supply no FactorJEPA training
signal. The metric measures \textbf{linear accessibility of motion
information}, not complete motion reconstruction.

% =====================================================================
\subsection{Secondary Semantic and Temporal Diagnostics}
\label{app:secondary_metric_definitions}

\paragraph{Semantic probes.}
Action top-1 and Taxonomy F1 use fixed-capacity probes trained on the
training split and selected on the validation split. Probe architecture,
optimization, and regularization are identical across world models.
Taxonomy F1 uses the same class support and averaging convention for
every method; unsupported classes are not silently removed on a
method-specific basis.

\paragraph{Rollout and horizon stability.}
Let \(e_n^{(q)}\) denote future-token error after rollout step \(q\).
Rollout-drift slope is the OLS coefficient of \(e_n^{(q)}\) against
\(q\). Similarly, L1-vs-\(\Delta t\) decay is the slope of latent
prediction error over the evaluated future horizons. Both are computed
using a fixed set of rollout depths and horizons.

The free-running exposure-bias gap is

\begin{equation*}
g_n^{\mathrm{exp}}
=
e_{n,\mathrm{free}}^{\mathrm{future}}
-
e_{n,\mathrm{teacher}}^{\mathrm{future}},
\end{equation*}

where both errors use the same target encoder and future horizon.

\paragraph{Order and pace sensitivity.}
Frame-order sensitivity is the increase in prediction error after a
fixed temporal permutation:

\begin{equation*}
s_n^{\mathrm{order}}
=
e_n^{\mathrm{future}}
\left(
I_{\mathrm{perm}}(x_n)
\right)
-
e_n^{\mathrm{future}}(x_n).
\end{equation*}

Arrow-of-Time, temporal-order, and playback-pace accuracy use frozen
evaluation protocols with fixed transformation classes. Arrow-of-Time
distinguishes forward from reversed clips; temporal-order detects frame
permutations; playback-pace identifies the applied temporal-rate
transformation.

\paragraph{Temporal correspondence.}
For each source frame, the temporal-correspondence evaluator identifies
a nearest match in the paired sequence and maps that match back to the
source. TCC cycle-back error measures the resulting temporal distance.
TCC Kendall \(\tau\) measures rank agreement between predicted and
ground-truth temporal ordering. Both use the same frozen features and
matching rule for every method.

% =====================================================================
\subsection{Paired Clustered Inference and Multiplicity}
\label{app:clustered_statistics}

Let \(y_{M,s,c,v,i}\) denote a metric for method \(M\), training seed
\(s\), city \(c\), source video \(v\), and clip \(i\). For metrics with
multiple horizons, masks, or interventions, those observations are
first reduced to one clip-level score using the definitions above.

\paragraph{Primary estimand.}
To prevent high-volume cities or recordings from dominating the
evaluation, we use equal weighting at the city and source-video levels:

\begin{equation*}
\begin{aligned}
\bar y_{M,c,v}
&=
\frac{1}{S}
\sum_{s=1}^{S}
\frac{1}{|\mathcal I_{c,v}|}
\sum_{i\in\mathcal I_{c,v}}
y_{M,s,c,v,i},
\\
\widehat\mu_M
&=
\frac{1}{|\mathcal C|}
\sum_{c\in\mathcal C}
\frac{1}{|\mathcal V_c|}
\sum_{v\in\mathcal V_c}
\bar y_{M,c,v}.
\end{aligned}
\end{equation*}

The primary paired effect between methods \(A\) and \(B\) is

\begin{equation*}
\widehat\delta_{A,B}
=
\widetilde\mu_A
-
\widetilde\mu_B,
\end{equation*}

where the direction-normalized mean is

\begin{equation*}
\widetilde\mu_M
=
\begin{cases}
-\widehat\mu_M, & \text{if lower is better},\\
\phantom{-}\widehat\mu_M, & \text{if higher is better}.
\end{cases}
\end{equation*}

Consequently, \(\widehat\delta_{A,B}>0\) always favors method \(A\).
Clip-weighted estimates are reported as an aggregation sensitivity,
not as the primary estimand.

\paragraph{Paired hierarchical bootstrap.}
For each bootstrap replicate, we:

\begin{enumerate}
    \item resample cities with replacement;
    \item resample source videos within each sampled city;
    \item retain all clips from every sampled source video;
    \item apply the same sampled cities, videos, clips, masks, horizons,
    and interventions to all compared methods; and
    \item recompute the complete metric and aggregation pipeline.
\end{enumerate}

This preserves within-recording dependence and method pairing.
We report \(95\%\) BCa intervals for absolute method means and paired
effects. The source-video group, rather than the clip, is the lowest
independently resampled unit.

\paragraph{Training-seed variation.}
For every method and metric, we report all \(S=3\) executed seed values,
their mean, and

\begin{equation*}
\operatorname{SD}_{\mathrm{seed}}(M)
=
\sqrt{
\frac{1}{S-1}
\sum_{s=1}^{S}
\left(
\widehat\mu_{M,s}
-
\overline\mu_M
\right)^2
}.
\end{equation*}

The primary geographic estimand averages the three seeds before
city-level aggregation. As a sensitivity analysis, a crossed bootstrap
resamples seeds independently of cities while applying the same
resampled seed indices to all methods. Because only three seeds are
available, seed is not treated as a primary random-effect variance
component.

\paragraph{Mixed-effects sensitivity.}
As a model-based check, we fit

\begin{equation*}
\begin{aligned}
y_{M,s,c,v,i}
={}&
\beta_0
+
\beta_M^{\mathrm{method}}
+
\gamma_s^{\mathrm{seed}}
+
u_c^{\mathrm{city}}
\\
&+
u_{v(c)}^{\mathrm{source}}
+
\varepsilon_{M,s,c,v,i},
\end{aligned}
\end{equation*}

where method is the effect of interest, seed is a fixed blocking
factor, and city and source video are nested random intercepts. The
mixed-effects analysis is considered supportive only when its effect
direction agrees with the paired clustered estimate.

\paragraph{Multiple comparisons.}
Within each backbone, the primary confirmatory family contains the
prespecified attribution contrasts crossed with the four primary
diagnostics. Two-sided paired-bootstrap \(p\)-values are adjusted using
Holm's step-down procedure. The secondary scorecard is controlled using
Benjamini--Hochberg FDR and is explicitly labeled exploratory.

We report unadjusted absolute effects and clustered intervals together
with adjusted decisions. An unadjusted \(95\%\) interval excluding zero
is not described as Holm-corrected evidence unless the corresponding
adjusted decision also passes the prespecified level.

% =====================================================================
\subsection{Metric Sensitivity and Complete Numerical Results}
\label{app:metric_robustness}

Robustness is evaluated along four prespecified axes.

\paragraph{Metric-construction sensitivity.}
For future prediction, we report horizon-specific MSE and, where
executed, tokenwise L1 as a secondary alternative. For intervention
consistency, we vary \(\epsilon_{\mathrm{int}}\) and the minimum target
effect while reporting the retained intervention fraction. For
missing-evidence robustness, we compare absolute slope, relative slope,
and masking AUC while preserving the same mask realizations.

A conclusion is considered stable only when its direction is preserved
across the prespecified variants. The primary metric definition is not
changed after inspecting test performance.

\paragraph{Aggregation sensitivity.}
We compare:

\begin{itemize}
    \item equal-city and clip-weighted means;
    \item city--source-video and source-video-only bootstrap intervals;
    \item seed-averaged and crossed seed-resampling estimates; and
    \item complete and leave-one-city-out evaluation.
\end{itemize}

For each leave-one-city-out run, all clips from the omitted city are
removed before recomputing the metric and method contrast.

\paragraph{Cross-scale ranking robustness.}
For diagnostic \(m\), let
\(\mathbf s_G^{(m)}\) and \(\mathbf s_g^{(m)}\) contain the
direction-normalized scores of the common method set at the ViT-G and
ViT-g scales. We report

\begin{equation*}
\begin{aligned}
\rho_m
&=
\operatorname{Spearman}
\left(
\mathbf s_G^{(m)},
\mathbf s_g^{(m)}
\right),
\\
\tau_m
&=
\operatorname{Kendall}
\left(
\mathbf s_G^{(m)},
\mathbf s_g^{(m)}
\right).
\end{aligned}
\end{equation*}

Because the evaluated methods are a fixed comparison set, these
correlations are interpreted descriptively. Robustness is assessed
through leave-one-method-out and leave-one-family-out ranges:

\begin{equation*}
\mathcal R_{\rho,m}
=
\left[
\min_j \rho_m^{(-j)},
\,
\max_j \rho_m^{(-j)}
\right].
\end{equation*}

Method families comprise frozen, conventional fine-tuning,
parameter-efficient adaptation, FactorJEPA variants, and encoder-scope
variants. Cross-scale agreement is described as
\textbf{scale robustness}, not independent replication.

\paragraph{Complete scorecards.}
Figures~\ref{fig:ablation_scorecard_vitg} and
\ref{fig:ablation_scorecard_vitg} show all executed diagnostics for
the ViT-G and ViT-g backbones. Exact primary values and clustered
intervals are reported in
Figure~\ref{fig:ablation_scorecard_vitg}. The complete numerical
scorecard additionally records, for every method and diagnostic:

\begin{equation*}
\resizebox{0.98\columnwidth}{!}{$\displaystyle
\left(
\text{mean},
\;
95\%\ \mathrm{clustered\ CI},
\;
\text{seed SD},
\;
\text{paired reference effect},
\;
\text{adjusted decision}
\right)
$}
\end{equation*}

The graphical scorecards provide the broad performance profile, whereas
the numerical tables and prespecified contrasts determine the
inferential conclusions. Accordingly, no method is declared uniformly
superior on the basis of isolated secondary wins.

% =====================================================================
\section{Latent-to-RGB Decoding and Qualitative Analysis}
\label{app:rgb_decoder}

The latent-to-RGB decoder provides a \textbf{secondary,
interpretability-oriented evaluation} of the predicted future
representation. It does not contribute to FactorJEPA training and is
not used for model selection. Its purpose is to expose what a predicted
future latent implies in pixel space under a fixed rendering model.

The evaluation follows three constraints. First, FactorJEPA and all
comparison models remain frozen during decoder training. Second, one
decoder is trained per backbone scale using only target-encoder latents
from the training partition. Third, the selected decoder is frozen and
applied without method-specific adaptation to every predicted-latent
source. Consequently, RGB differences within a backbone scale cannot
be attributed to different decoder capacities or optimization budgets.

% =====================================================================
\subsection{Decoder Architecture and Latent Transport}
\label{app:rgb_architecture}

Let
\(
Y\in\mathbb{R}^{B\times T_y\times P\times d_J}
\)
denote a sequence of future JEPA tokens, where \(P=h_Jw_J\) is the
spatial token count and \(d_J\in\{1664,1408\}\) for the ViT-G and
ViT-g backbones, respectively. The Cosmos decoder expects a spatially
organized conditioning tensor rather than an unordered JEPA token
sequence. We therefore introduce a trainable transport operator
\(\mathcal T_\omega\) that aligns token dimensionality, spatial
organization, temporal order, and feature statistics.

For future index \(\tau\), the transported representation is

\begin{equation*}
\begin{aligned}
U_\tau
&=
\operatorname{reshape}_{h_J\times w_J}
\left[
W_{\mathrm{in}}\,
\operatorname{LN}(Y_\tau)
\right],                                                     \\
Z_\tau
&=
\mathcal B_\omega
\left(
U_\tau
+
E_{\mathrm{space}}
+
E_{\mathrm{time}}(\tau)
\right),                                                     \\
\mathcal T_\omega(Y)_\tau
&=
W_{\mathrm{out}}\,
\operatorname{LN}(Z_\tau)
\in
\mathbb{R}^{h_C\times w_C\times d_C},
\end{aligned}
\end{equation*}

where \(W_{\mathrm{in}}\) and \(W_{\mathrm{out}}\) are learned
projections, \(\mathcal B_\omega\) is the executed transport stack,
and \(E_{\mathrm{space}}\) and \(E_{\mathrm{time}}\) preserve spatial
and temporal position. Any interpolation required to obtain
\(h_C\times w_C\) is performed inside \(\mathcal T_\omega\) and is
identical for all latent sources at a fixed backbone scale.

The rendered future is

\begin{equation*}
\widehat x
=
\mathcal D_\Omega(Y)
=
\mathcal G_{\psi_C}
\left(
\mathcal T_\omega(Y)
\right),
\qquad
\Omega=\{\omega,\psi_C^{\mathrm{adapt}}\},
\end{equation*}

where \(\mathcal G_{\psi_C}\) is initialized from the reported
Cosmos checkpoint~\cite{nvidia2025cosmos}. The notation
\(\psi_C^{\mathrm{adapt}}\) denotes exactly those Cosmos parameters
updated during decoder adaptation; all remaining Cosmos parameters
stay frozen. After validation-based checkpoint selection, the complete
mapping \(\mathcal D_\Omega\) is frozen.

\begin{table*}[t]
\centering
\caption{
\textbf{Latent-to-RGB decoder contract.}
One scale-specific decoder is trained for each JEPA embedding width.
Within a backbone scale, every method uses exactly the same transport
operator, Cosmos component, preprocessing, and evaluation settings.
Fields marked \emph{from log} must be copied verbatim from the executed
decoder configuration.
}
\label{tab:rgb_decoder_contract}
\scriptsize
\setlength{\tabcolsep}{5pt}
\renewcommand{\arraystretch}{1.15}

\begin{tabularx}{\textwidth}{
    p{0.20\textwidth}
    >{\raggedright\arraybackslash}X
    >{\raggedright\arraybackslash}X
}
\toprule
\textbf{Component}
&
\textbf{Executed configuration}
&
\textbf{Comparison control}
\\
\midrule

\rowcolor{DecoderBlue}
JEPA latent source
&
ViT-G: \(d_J=1664\);
ViT-g: \(d_J=1408\);
future frames \(13{:}16\) from the common \(384\times384\),
16-frame protocol.
&
Target and predicted latents use the same temporal indices,
normalization, token ordering, and spatial support.
\\

Transport operator
&
Layer normalization, input projection, spatial reshaping,
spatiotemporal transport stack, output projection, and positional
encoding. Exact depth, width, heads, and \(d_C\):
\emph{[from log]}.
&
One transport operator per backbone scale; no method-specific
projection or normalization.
\\

Cosmos initialization
&
Checkpoint identifier and revision:
\emph{[from log]}. Checkpoint hash:
\emph{[from log]}.
&
The same initialization is used for all methods evaluated at a fixed
backbone scale.
\\

Cosmos update scope
&
Trainable modules:
\emph{[from log]}. Frozen modules:
\emph{[from log]}.
&
The update scope is fixed before decoder training and is not selected
separately for individual prediction methods.
\\

RGB output
&
Frames, resolution, color range, and frame rate:
\emph{[from log]}.
&
All outputs undergo the same clipping, resizing, and inverse
normalization before evaluation.
\\

Frozen upstream models
&
Online encoder, momentum target encoder, monolithic or factorized
predictor, DINOv2 teacher, and factor-target operators.
&
RGB reconstruction gradients never enter a world model or
pseudo-target generator.
\\

Frozen RGB evaluators
&
LPIPS network, optical-flow estimator, and agent detector.
Exact checkpoints and thresholds: \emph{[from log]}.
&
No RGB evaluator supplies FactorJEPA supervision or participates in
decoder optimization.
\\

\bottomrule
\end{tabularx}
\end{table*}

% =====================================================================
\subsection{Decoder Training and Comparison Neutrality}
\label{app:rgb_training}

For a training example \(n\), let \(x_n^+\) denote the observed future
RGB target and let

\[
Y_n^\star
=
\operatorname{sg}
\left[
\overline f_{\bar\theta}(x_n^+)
\right]
\]

be the corresponding frozen target-encoder latent. The decoder is
trained \textbf{only} on pairs
\(\{(Y_n^\star,x_n^+)\}\) from the training partition:

\begin{equation*}
\Omega^\star
=
\arg\min_{\Omega}
\frac{1}{|\mathcal D_{\mathrm{train}}|}
\sum_{n\in\mathcal D_{\mathrm{train}}}
\mathcal L_{\mathrm{decode}}
\left(
\mathcal D_\Omega(Y_n^\star),
x_n^+
\right).
\end{equation*}

Predicted latents from FactorJEPA, FactorJEPA-RAW, or any adaptation
baseline are excluded from decoder optimization and checkpoint
selection. This prevents the decoder from adapting preferentially to
the latent distribution of one prediction method.

The reconstruction objective is

\begin{equation*}
\begin{aligned}
\mathcal L_{\mathrm{decode}}
={}&
\lambda_{\mathrm{pix}}\mathcal L_{\mathrm{char}}
+
\lambda_{\mathrm{perc}}\mathcal L_{\mathrm{LPIPS}}
+
\lambda_{\mathrm{str}}\mathcal L_{\mathrm{SSIM}}
\\
&+
\lambda_{\mathrm{mot}}\mathcal L_{\mathrm{motion}}
+
\lambda_{\mathrm{reg}}\mathcal R(\Omega),
\end{aligned}
\end{equation*}

where the robust pixel term is

\begin{equation*}
\mathcal L_{\mathrm{char}}
=
\frac{1}{|\mathcal P|}
\sum_{p\in\mathcal P}
\sqrt{
\left(
\widetilde x(p)-x^+(p)
\right)^2+\epsilon_{\mathrm{char}}^2
}.
\end{equation*}

Here,
\(\mathcal L_{\mathrm{LPIPS}}\) measures perceptual discrepancy,
\(\mathcal L_{\mathrm{SSIM}}=1-\operatorname{SSIM}\) penalizes
structural distortion, and \(\mathcal L_{\mathrm{motion}}\), when
enabled in the executed configuration, compares frozen-estimator flow
between the final context frame and the reconstructed or observed
future. Terms assigned zero weight are disabled rather than silently
omitted.

The checkpoint is selected exclusively by the prespecified validation
criterion. No test frame, test latent, test metric, or prediction from
a compared method influences decoder training, early stopping, or
hyperparameter selection.

At evaluation, the selected parameters \(\Omega^\star\) are frozen and

\begin{equation*}
\widehat x_{n,M}
=
\mathcal D_{\Omega^\star}
\left(
\widehat Y_{n,M}
\right)
\end{equation*}

is computed for every method \(M\) using identical preprocessing,
sampling parameters, precision, and random seeds. When decoding is
stochastic, all methods use the same prespecified seed set and results
are averaged over that set. Best-of-\(K\) sample selection is not used
for the primary comparison.

% =====================================================================
\subsection{Oracle Reconstruction and Forecast-Gap Decomposition}
\label{app:rgb_oracle}

A decoded prediction can fail because the predicted latent is
incorrect, because the decoder cannot invert even a correct target
latent, or because both effects interact. We expose these sources
through an \textbf{oracle target-latent control}.

For each held-out example, define

\begin{equation*}
\widetilde x_n^\star
=
\mathcal D_{\Omega^\star}(Y_n^\star),
\qquad
\widehat x_{n,M}
=
\mathcal D_{\Omega^\star}(\widehat Y_{n,M}).
\end{equation*}

For a lower-is-better distortion \(d\), the three reported quantities
are

\begin{equation*}
\begin{aligned}
E_{\mathrm{oracle}}
&=
d(\widetilde x_n^\star,x_n^+),                                   \\
E_{\mathrm{e2e}}(M)
&=
d(\widehat x_{n,M},x_n^+),                                       \\
G_{\mathrm{visible}}(M)
&=
d(\widehat x_{n,M},\widetilde x_n^\star).
\end{aligned}
\end{equation*}

The oracle term measures the decoder's reconstruction limitation, the
end-to-end term measures the complete prediction-and-rendering system,
and \(G_{\mathrm{visible}}\) measures how strongly replacing the
target latent with the predicted latent changes the rendered future.
For metric distances such as pixel \(L_1\),

\begin{equation*}
E_{\mathrm{e2e}}(M)
\le
G_{\mathrm{visible}}(M)
+
E_{\mathrm{oracle}}
\end{equation*}

by the triangle inequality. This bound is not asserted for similarity
scores or learned perceptual measures that are not guaranteed metrics.

We additionally report the oracle-relative degradation. For a
lower-is-better metric \(Q^\downarrow\),

\begin{equation*}
\Delta_Q^\downarrow(M)
=
Q^\downarrow(\widehat x_{n,M},x_n^+)
-
Q^\downarrow(\widetilde x_n^\star,x_n^+),
\end{equation*}

whereas for a higher-is-better metric \(Q^\uparrow\),

\begin{equation*}
\Delta_Q^\uparrow(M)
=
Q^\uparrow(\widetilde x_n^\star,x_n^+)
-
Q^\uparrow(\widehat x_{n,M},x_n^+).
\end{equation*}

\begin{table*}[ht!]
\centering
\caption{
\textbf{Latent-to-RGB evaluation under a common frozen decoder.}
Entries should be reported as held-out mean
\([95\%\ \mathrm{clustered\ CI}]\). Values in parentheses denote the
direction-normalized deficit relative to target-latent oracle decoding;
positive deficits indicate worse performance than the oracle.
No predicted-latent source is used to train or select the decoder.
}
\label{tab:rgb_quantitative}
\scriptsize
\setlength{\tabcolsep}{4pt}
\renewcommand{\arraystretch}{1.15}

\begin{tabularx}{\textwidth}{
    p{0.22\textwidth}
    >{\centering\arraybackslash}X
    >{\centering\arraybackslash}X
    >{\centering\arraybackslash}X
    >{\centering\arraybackslash}X
    >{\centering\arraybackslash}X
}
\toprule
\textbf{Latent source}
&
\textbf{PSNR} \(\uparrow\)
&
\textbf{SSIM} \(\uparrow\)
&
\textbf{LPIPS} \(\downarrow\)
&
\textbf{Flow EPE} \(\downarrow\)
&
\textbf{Agent F1} \(\uparrow\)
\\
\midrule

\rowcolor{DecoderBlue}
Target latent \emph{(oracle)}
&
\multicolumn{1}{c}{\textit{[result]}}
&
\multicolumn{1}{c}{\textit{[result]}}
&
\multicolumn{1}{c}{\textit{[result]}}
&
\multicolumn{1}{c}{\textit{[result]}}
&
\multicolumn{1}{c}{\textit{[result]}}
\\

V-JEPA Auto-RGN
& \textit{[result]} & \textit{[result]} & \textit{[result]}
& \textit{[result]} & \textit{[result]} \\

\rowcolor{DecoderGold}
FactorJEPA-RAW
& \textit{[result]} & \textit{[result]} & \textit{[result]}
& \textit{[result]} & \textit{[result]} \\

\rowcolor{DecoderTeal}
\textbf{FactorJEPA}
& \textit{[result]} & \textit{[result]} & \textit{[result]}
& \textit{[result]} & \textit{[result]} \\

\bottomrule
\end{tabularx}
\end{table*}

\begin{table*}[ht!]
\centering
\caption{
\textbf{Operational taxonomy and attribution rules for decoded-future
failures.}
Let \(\mathsf{X}=x^{+}\) denote the observed future,
\(\mathsf{O}=\widetilde x^{\star}\) the target-latent oracle
reconstruction, and \(\mathsf{P}=\widehat x_M\) a decoded prediction.
Panel A uses the matched triplet
\((\mathsf{X},\mathsf{O},\mathsf{P})\) to localize error; Panel B
defines observable failure modes and corroborating diagnostics.
Attribution is diagnostic rather than causal because an off-manifold
predicted latent may expose nonlinear decoder sensitivities.
}
\label{tab:rgb_failure_taxonomy}
\scriptsize
\setlength{\tabcolsep}{4.5pt}
\renewcommand{\arraystretch}{1.15}

\begin{tabularx}{\textwidth}{
    p{0.16\textwidth}
    p{0.30\textwidth}
    p{0.27\textwidth}
    >{\raggedright\arraybackslash}X
}
\toprule
\textbf{Class}
&
\textbf{Operational signature}
&
\textbf{Oracle comparison}
&
\textbf{Interpretation or diagnostic}
\\
\midrule

\rowcolor{DecoderBlue}
\multicolumn{4}{l}{
\textbf{Panel A: Oracle-based attribution rules}
}
\\
\midrule

Oracle-present
&
The same artifact is visible in both
\(\mathsf{O}\) and \(\mathsf{P}\) relative to \(\mathsf{X}\).
&
\(\mathsf{O}\) already fails to reconstruct the relevant content.
&
Primarily a decoder-capacity or target-latent-invertibility limitation;
it must not be attributed solely to forecasting.
\\

Prediction-conditioned
&
\(\mathsf{O}\) preserves the relevant content, whereas
\(\mathsf{P}\) does not.
&
The artifact appears only after replacing the target latent with
\(\widehat Y_M\).
&
Localized to the prediction-conditioned pathway: inaccurate predicted
latent, distribution shift at the decoder input, or their interaction.
\\

Mixed or amplified
&
The artifact is present in \(\mathsf{O}\) but becomes materially
stronger in \(\mathsf{P}\).
&
Both oracle reconstruction error and oracle-to-prediction deviation
are non-negligible.
&
Decoder and forecasting effects coexist; report oracle error,
end-to-end error, and visible forecast gap separately.
\\

\midrule
\rowcolor{DecoderGold}
\multicolumn{4}{l}{
\textbf{Panel B: Observable decoded-future failure modes}
}
\\
\midrule

Layout drift
&
Static boundaries, façades, road geometry, or background support move
despite being stable in \(\mathsf{X}\).
&
Check whether the same displacement occurs in \(\mathsf{O}\).
&
Static-region error, boundary displacement, and spurious background
flow distinguish geometric drift from local texture variation.
\\

Agent omission
&
An agent visible in \(\mathsf{X}\) is absent, severely attenuated, or
merged into the background in \(\mathsf{P}\).
&
Determine whether the corresponding agent is recoverable in
\(\mathsf{O}\).
&
Agent recall, Agent F1, and dynamic-region error quantify the failure;
small or heavily occluded agents should be reported separately.
\\

Agent duplication
&
One reference agent produces multiple overlapping instances or
spatially inconsistent fragments in \(\mathsf{P}\).
&
If duplication is also present in \(\mathsf{O}\), classify it as
oracle-present.
&
Agent precision, duplicate-detection rate, and connected-component
fragmentation provide corroborating evidence.
\\

Interaction inconsistency
&
Nearby agents exhibit incorrect relative ordering, separation,
direction, or collision geometry.
&
Compare pairwise geometry in
\(\mathsf{X}\), \(\mathsf{O}\), and \(\mathsf{P}\).
&
Pairwise displacement, relative-motion error, and interaction-edge
agreement distinguish interaction failure from independent agent
misplacement.
\\

Motion under-dispersion
&
Moving agents become blurred, nearly stationary, or displaced toward
an average future.
&
An accurate \(\mathsf{O}\) with reduced motion only in
\(\mathsf{P}\) indicates prediction-conditioned averaging.
&
Flow EPE, predicted-to-reference flow-magnitude ratio, and
dynamic-region sharpness quantify the effect. It may reflect
deterministic prediction under multimodal futures.
\\

Appearance substitution
&
Coarse location and occupancy remain plausible, but local appearance
or category-specific structure changes.
&
Determine whether the substitution is already visible in
\(\mathsf{O}\).
&
Agent-crop LPIPS and frozen-descriptor similarity separate appearance
loss from geometric or occupancy failure.
\\

Unsupported content
&
\(\mathsf{P}\) introduces an agent, boundary, texture, or motion pattern
unsupported by \(\mathsf{X}\).
&
Oracle presence suggests decoder hallucination; prediction-only
presence indicates an off-manifold latent or prediction--decoder
interaction.
&
False-positive agent rate, static-region residuals, perceptual error,
and oracle comparison jointly support classification.
\\

\bottomrule
\end{tabularx}
\end{table*}

Positive values indicate degradation relative to oracle decoding.
These quantities are called \textbf{decoder-conditioned forecast
deficits}; they are not interpreted as an exact additive decomposition
because \(\mathcal D_{\Omega^\star}\) is nonlinear.

% =====================================================================
\subsection{Quantitative Decoded-Output Evaluation}
\label{app:rgb_quantitative}

We evaluate every decoded future using the same held-out examples and
five complementary diagnostics:

\begin{itemize}
    \item \textbf{PSNR \(\uparrow\)} measures pixel-level fidelity
    relative to the observed future;

    \item \textbf{SSIM \(\uparrow\)} measures preservation of local
    luminance, contrast, and structural organization;

    \item \textbf{LPIPS \(\downarrow\)} measures perceptual discrepancy
    using one fixed pretrained feature network;

    \item \textbf{Flow EPE \(\downarrow\)} measures motion disagreement
    between the observed and decoded futures using one frozen optical
    flow estimator; and

    \item \textbf{Agent F1 \(\uparrow\)} measures preservation of
    detectable dynamic agents using one frozen detector and a fixed
    matching threshold.
\end{itemize}

For motion evaluation, let \(x_n^c\) be the final context frame. Using
the frozen estimator \(\Phi_{\mathrm{flow}}\),

\begin{equation*}
F_n^\star
=
\Phi_{\mathrm{flow}}(x_n^c,x_n^+),
\qquad
\widehat F_{n,M}
=
\Phi_{\mathrm{flow}}(x_n^c,\widehat x_{n,M}),
\end{equation*}

and

\begin{equation*}
\operatorname{EPE}_{n,M}
=
\frac{1}{|\mathcal P|}
\sum_{p\in\mathcal P}
\left\|
\widehat F_{n,M}(p)-F_n^\star(p)
\right\|_2.
\end{equation*}

Agent F1 uses detections produced independently on the observed and
decoded future frames. Predictions are matched by the prespecified
class-compatible overlap rule; the detector, confidence threshold,
matching threshold, and taxonomy are held fixed across methods. This
metric assesses \emph{detector-visible agent preservation}, not
photorealism or manually verified object correctness.

We further report results over prespecified density and visibility
strata. Strata are computed from the fixed automatic measurement
pipeline before RGB evaluation and remain unchanged across methods.
Boundaries are derived only from the training distribution. Clips
within a source video retain the same cluster identity when estimating
uncertainty. These analyses test whether decoded-output quality
degrades disproportionately in crowded or heavily occluded scenes;
they are secondary to the complete held-out comparison.

% =====================================================================
\subsection{Qualitative Comparisons and Failure Analysis}
\label{app:rgb_qualitative}

Qualitative examples use a fixed column order:

\begin{equation*}
\begin{aligned}
&\text{context}
\;\big|\;
\text{observed future}
\;\big|\;
\text{oracle decode}
\\[-1pt]
&\qquad\big|\;
\text{V-JEPA}
\;\big|\;
\text{FactorJEPA-RAW}
\;\big|\;
\text{FactorJEPA}.
\end{aligned}
\end{equation*}

The oracle column is retained in every row because it distinguishes an
artifact already present when decoding the correct target latent from
an error introduced by future prediction. All methods use the same
display range, output resolution, temporal index, decoder checkpoint,
and sampling seed.

\begin{figure*}[ht!]
\centering
\includegraphics[
    width=0.99\textwidth,
    keepaspectratio
]{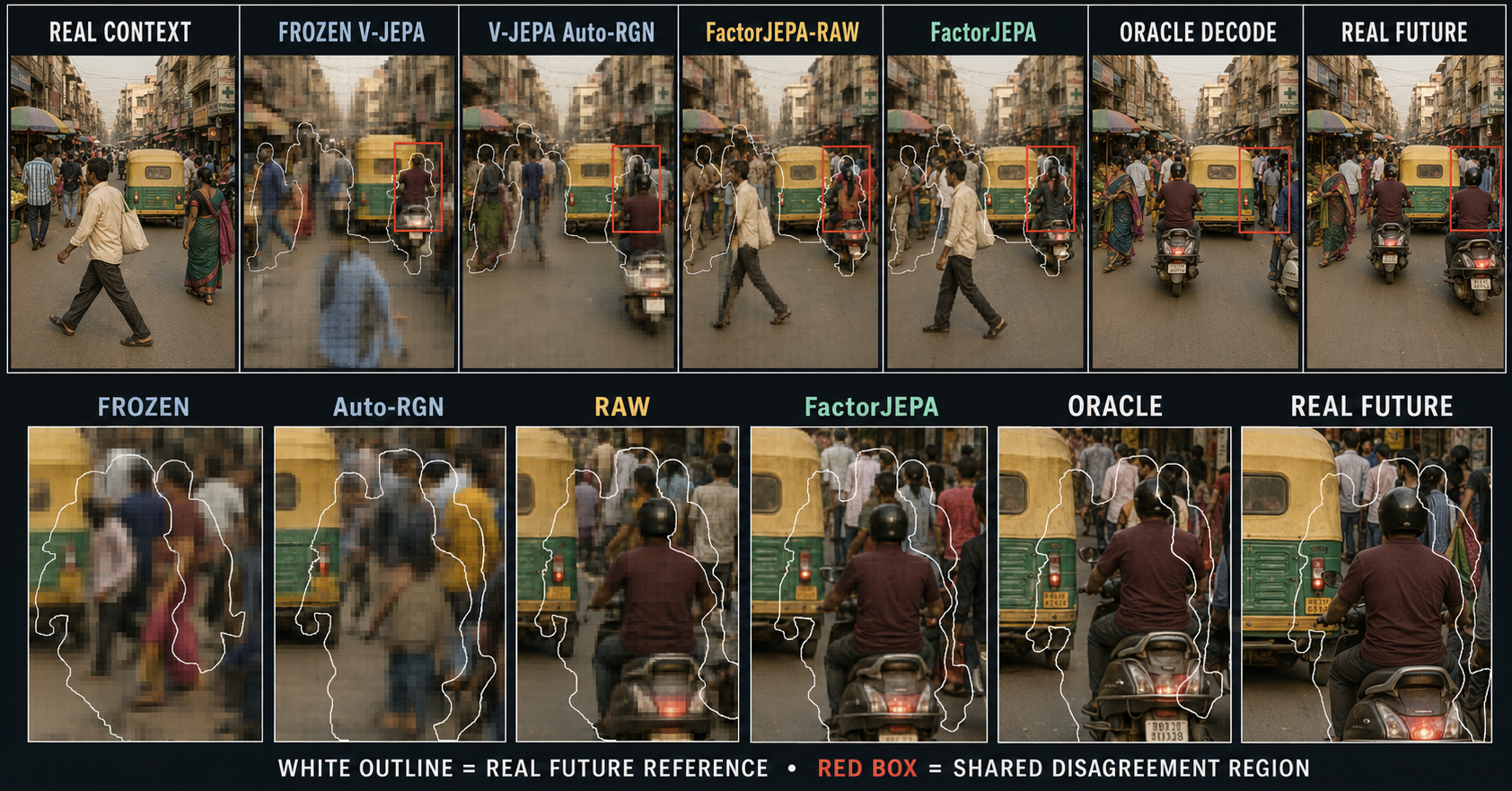}
\caption{
\textbf{Illustrative layout for future-consistency inspection.}
The top row compares the observed context with decoded futures from
frozen V-JEPA, V-JEPA Auto-RGN, FactorJEPA-RAW, and FactorJEPA,
followed by the target-latent oracle reconstruction and observed
future. The identical white contour, derived from the observed future,
is overlaid on every predicted panel to expose disagreement in agent
occupancy, position, visibility, and interaction geometry. Red boxes
identify a common high-disagreement region, enlarged in the bottom
row. The synthesized panels demonstrate the intended visualization
protocol and are not experimental model outputs.
}
\label{fig:rgb_prediction_gallery_layout}
\end{figure*}

For an additional functional diagnostic, we remove one factor
contribution before decoding:

\begin{equation*}
\widehat x_n^{(-k)}
=
\mathcal D_{\Omega^\star}
\left(
\widehat Y_n-\widehat Y_{n,k}
\right),
\qquad
k\in\{L,A,I\},
\end{equation*}

and visualize the normalized influence map

\begin{equation*}
\mathcal I_{n,k}(p)
=
\frac{
\left|
\widehat x_n(p)-\widehat x_n^{(-k)}(p)
\right|
}{
\epsilon+
\sum_{q\in\mathcal P}
\left|
\widehat x_n(q)-\widehat x_n^{(-k)}(q)
\right|
}.
\end{equation*}

These maps show where the rendered output is sensitive to removal of a
predictive pathway. They are \textbf{functional perturbation
diagnostics}, not causal pixel explanations and not evidence of
identifiable latent factors.

Failure cases are categorized using observable symptoms and the oracle
control:

\paragraph{Interpretation.}
RGB decoding offers an interpretable view of future-latent predictions,
but it remains a decoder-conditioned diagnostic. The latent-space
metrics provide the primary comparison because they evaluate predicted
representations directly. RGB metrics and qualitative results provide
complementary evidence about whether those representations preserve
visually recoverable layout, agent occupancy, interaction structure,
and motion under one fixed rendering interface.       % Appendix A--E

\end{document}